%% file: main.tex
\documentclass{article}
\usepackage[preprint]{icml2026}

\usepackage{microtype}
\usepackage{graphicx}
\usepackage{subcaption}
\usepackage{booktabs}
\usepackage{hyperref}
\usepackage{algorithm}
\usepackage{algorithmic}
\usepackage{amsmath}
\usepackage{amssymb}
\usepackage{mathtools}
\usepackage{amsthm}
\usepackage{multirow}
\usepackage{xcolor}
\usepackage[capitalize,noabbrev]{cleveref}

\theoremstyle{plain}
\newtheorem{theorem}{Theorem}
\newtheorem{proposition}{Proposition}
\newtheorem{lemma}{Lemma}
\newtheorem{corollary}{Corollary}
\theoremstyle{definition}
\newtheorem{definition}{Definition}

\theoremstyle{remark}

\newcommand{\cin}{c^{+}}
\newcommand{\cdel}{c^{-}}
\newcommand{\AUC}{\mathrm{AUC}}
\newcommand{\Btrap}{B_{\mathrm{trap}}}
\newcommand{\Bab}{B_{\mathrm{ab}}}
\newcommand{\cid}{c_{\mathrm{id}}}
\newcommand{\R}{\mathbb{R}}
\newcommand{\cars}{CARS}
\makeatletter\newcommand{\tabinput}[1]{\IfFileExists{#1}{\input{#1} }{(pending)\\}}\makeatother

\input{tables/numbers.tex}

\input{tables/values.tex}

\input{tables/numbers2.tex}

\IfFileExists{tables/numbers3.tex}{

\input{tables/numbers3.tex}}{}

\icmltitlerunning{When Is a Relation Worth Trusting?}

\begin{document}

\twocolumn[
  \icmltitle{When Is a Relation Worth Trusting? Manipulation Costs, Exposure,
    and the Trust Trap in Multi-Relational Graph Learning}
  \icmlsetsymbol{equal}{*}
  
\begin{icmlauthorlist}
    \icmlauthor{Vincess Dongmo}{equal,aff1}
    \icmlauthor{Astride Melvin Fokam Ninyim}{equal,aff1,aff2}
  \end{icmlauthorlist}

  \icmlaffiliation{aff1}{SYNTHI-AI, Cameroon}
  \icmlaffiliation{aff2}{Université du Québec à Chicoutimi (UQAC), Canada}

  \icmlcorrespondingauthor{Vincess Dongmo}{vincess.dongmo@synthi-ai.com}
  \icmlcorrespondingauthor{Astride Melvin Fokam Ninyim}{amfninyim@etu.uqac.ca}

  \icmlkeywords{graph neural networks, adversarial robustness, fraud detection, heterogeneous graphs, strategic manipulation}
  \vskip 0.3in
]

\printAffiliationsAndNotice{\icmlEqualContribution}

\begin{abstract}
Multi-relational graph detectors learn from clean data how much to trust each relation, but an adversary pays a different price to manipulate each relation and each action; taking such costs as given, we ask when using a relation is worse than ignoring it. Our first main result proves that degree gating strictly beats every selection or reweighting of a relation when part of the population is isolated on it and the attacker cannot afford to open the gate, and that this window is sharp; on real graphs, validated gating gains exactly on the relations that meet these conditions and nowhere else. Our second shows that a per-node exposure bound, which extends to message passing, tells a platform which relation to harden: exposure per unit cost, computed without any attack, picks the relation whose hardening raises the exact trap threshold most in \HDexpHit{} of \HDn{} cases, against at most \HDbestOther{} for informativeness, cheapness, or an analysis with equal edit costs. Supporting tools give exact worst cases for linear relational models, certified brackets for R-GCN and tree ensembles, and a finite-sample guarantee for adversarial validation, under which clean-trained graph detectors fall below their graph-free counterparts in \NTrapBase{} of \NCellsBase{} settings on six graphs. A selection protocol built on these results is never significantly worse than ignoring the graph; with at most four relations it mostly abstains, as validation alone would, and its cost-aware proposals help when relations are many, so far shown on synthetic graphs. Without cost information, the safe choice is to abstain.
\end{abstract}

\section{Introduction}
\label{sec:intro}

Fraud, collusion, and spam detectors increasingly learn from multi-relational graphs: reviews that share a user or a product, firms that share an owner or a supplier, accounts that share a device. Such models learn from clean data how much to trust each relation, through relation-specific weights \citep{schlichtkrull2018rgcn}, semantic attention \citep{wang2019han}, neighbor selection \citep{dou2020caregnn,liu2021pcgnn}, or spectral filters \citep{tang2022bwgnn,wu2023splitgnn}. That such detectors can be manipulated is well known: fraudsters camouflage their relations \citep{dou2020caregnn,zhang2024digin}, structural attacks degrade graph models \citep{zugner2018nettack,geisler2021prbcd}, gangs coordinate injections \citep{choi2025monti}, and cost-aware attacks price the nodes of homogeneous graphs \citep{liao2024riskeeper}.

What is missing is a theory of how the \emph{economics} of manipulation interact with the \emph{information} each relation carries. An adversary does not pay the same price for every relation or every action: reviewing a popular product creates an edge almost for free, reusing an aged account is costly, and concealing a tie to an accomplice may be cheap or expensive depending on identity verification. A relation can therefore be statistically valuable and economically unsafe. Neighbor filters ask which neighbors to keep within a relation, cost-aware defenses ask which nodes are at risk, and robust optimization asks how to train under attack. We ask a prior question: \emph{should a relation be used at all, given how informative and how manipulable it is?}

\paragraph{Contributions.} The paper has two main results; the rest are tools and supporting results (\cref{tab:delta} compares each with the workshop version and the closest prior work).
\textbf{(i) Degree gating strictly beats selection and reweighting} of a relation on which part of the population is isolated, as long as the attacker cannot open the gate, and the window is sharp (\cref{thm:gatedom}); on real graphs, validated gating gains on exactly the relations that meet these conditions (\cref{tab:gatereal}).
\textbf{(ii) Which relations are dangerous, and which to harden.} A per-node exposure bound that extends to message passing (\cref{prop:exposure,prop:gnn}) and the budget beyond which no detector can use the graph (\cref{thm:abandon}) say when a relation's signal is outweighed by its manipulation cost. Combined with exact thresholds, they answer a question that robustness analyses with equal edit costs cannot pose: which relation a platform should make harder to manipulate. The first-order exposure per unit cost, computed without any attack, picks the relation whose hardening raises the exact trap threshold most in \HDexpHit{} of \HDn{} cases, against at most \HDbestOther{} for informativeness, cheapness, weight size, or an exact analysis under equal costs (\cref{tab:hardening}).
\textbf{Tools: exact, certified, and validated evaluation.} Exact worst cases and thresholds for linear relational models under heterogeneous costs (\cref{prop:exact}), certified brackets for R-GCN and tree ensembles that we label tight, informative, or loose (\cref{prop:cert}), and a finite-sample guarantee for adversarial validation whose radius grows with the number of candidates checked (\cref{prop:val}).
\emph{Supporting results}: the threshold's structure (\cref{prop:structure}), robust planning, maximin choice, and hardening (\cref{cor:robust}). \emph{Protocol}: \cars{}-D lets the theory propose three candidates and adversarial validation pick one.
\paragraph{What the evidence supports.} (1) Clean-trained relational detectors are fragile under cheap manipulations: exactly for linear models, through evaluated upper bounds for nonlinear ones. (2) Gating helps exactly where the theorem says, and nowhere else. (3) On real graphs with one to four relations, \cars{}-D is never significantly worse than ignoring the graph and mostly abstains, as validation alone would; its cost-aware proposals help with many relations, so far on synthetic graphs (\cref{tab:decomp}). All of this is conditional on specified costs: the paper studies selection under given cost scenarios, not real manipulation costs (\cref{apd:threat}).
A workshop paper \citep{anon2026workshop} introduced the threat model and studied the phenomenon on the CSBM, YelpChi, and Amazon; the exposure bound (\cref{prop:exposure}), the linear-response part of \cref{thm:abandon}, and the cost-monotonicity part of \cref{cor:robust} appeared there. \Cref{apd:threat} (\cref{tab:delta}) lists what is new.

\section{Problem Setup}
\label{sec:setup}

Let $G=(V,\{E_r\}_{r=1}^R,X)$ be a graph with $R$ relations, node features $X$, and labels $y_v\in\{0,1\}$ (1 for fraud); $\mathcal N_r(v)$ denotes the neighbors of $v$ under $r$. A detector $f$ is trained on the clean graph. At test time, each fraudulent node $v$ edits its own edges (evasion): on relation $r$ it may inject $b_{v,r}$ edges at unit cost $\cin_r$ or delete $k_{v,r}$ edges to accomplices at unit cost $\cdel_r$. Writing $\theta_v=(b_v,k_v)\in\R^{2R}_{\ge0}$ and $c=(\cin,\cdel)\in\R^{2R}_{>0}$, the per-node feasible set is
\begin{equation}
\label{eq:budget}
\Theta_B(c)=\big\{\theta\ge0:\ c^\top\theta\le B,\ k_r\le K_r\big\},
\end{equation}
with $K_r$ the number of accomplices on $r$. A \emph{gang} instead shares a total budget $\bar B|T|$ over its members $T$. When relations are projections of entities (two reviews share a user), we also consider \emph{realizable} actions that reassign an entity attribute and rebuild all induced edges at once (\cref{sec:exp}). Costs encode relative manipulation difficulty rather than money; since $\Theta_{\alpha B}(\alpha c)=\Theta_B(c)$, only cost-to-budget ratios matter.

\begin{definition}[Trust trap]
\label{def:trap}
Let $f_0$ be the graph-free counterpart of $f$, the same learner on node features only. With $\Delta_{\mathrm{clean}}$ the clean performance gap of $f$ over $f_0$ and $\Delta_{\mathrm{rob}}(B,c)$ the gap under the worst attack in $\Theta_B(c)$, $f$ is in a \emph{trust trap} at $(B,c)$ if $\Delta_{\mathrm{clean}}>0\ge\Delta_{\mathrm{rob}}(B,c)$. Its \emph{trap threshold} is $\Btrap(f;c)=\inf\{B:\Delta_{\mathrm{rob}}(B,c)\le0\}$.
\end{definition}

\section{Theory}
\label{sec:theory}

\paragraph{Levels of guarantee.} We separate three kinds of statements. \emph{Exact} results hold without approximation inside a stated model: \cref{prop:structure} for any detector, \cref{thm:abandon}(a) in the CSBM, and \cref{prop:exact} for linear relational models. \emph{Bounds} hold for broader classes but may be loose: \cref{prop:exposure,prop:gnn,prop:gate}, the certified AUC of \cref{prop:cert}, and the statistical guarantee of \cref{prop:val}. \emph{Evaluated} worst cases, reported for nonlinear detectors in \cref{sec:exp}, come from the strongest attacks we know and upper-bound the true worst case; \cref{prop:cert} brackets it from below.

\subsection{The structure of the trap threshold}
The first result is elementary and needs no model of the data; we state it because the experiments and the robust-planning results rely on its exact form, including where it does \emph{not} give concavity. An attack policy $\pi$ assigns each fraudulent node $v$ an edit vector $\theta_v(\pi)\in\mathbb Z^{2R}_{\ge0}$ (counts of injections and deletions per relation, with $k\le K$); performance is measured after all edits are applied. Let $\mathcal T(f)$ be the set of \emph{trap policies}, under which $f$ performs at most as well as $f_0$. A policy is affordable under per-node budgets if $\max_v c^\top\theta_v(\pi)\le B$, and under a gang budget if $\frac1{|T|}\sum_vc^\top\theta_v(\pi)\le\bar B$.

\begin{lemma}[Structure]
\label{prop:structure}
For every detector $f$ and performance measure, (a) $\Btrap(f;c)=\min_{\pi\in\mathcal T(f)}\max_vc^\top\theta_v(\pi)$ and the gang threshold is $\bar B_{\mathrm{trap}}(f;c)=\min_{\pi\in\mathcal T(f)}\frac1{|T|}\sum_vc^\top\theta_v(\pi)\le\Btrap(f;c)$, both minima being attained. (b) Both thresholds are nondecreasing and positively homogeneous of degree one in $c$. (c) $\bar B_{\mathrm{trap}}$ is concave and piecewise linear in $c$, and so is $\Btrap$ whenever performance depends on a policy only through a common edit vector, as in the CSBM; in general $\Btrap$ need not be concave.
\end{lemma}

The gang threshold and the CSBM threshold are lower envelopes of lines, one per attack strategy: they grow linearly in a relation's cost while the cheapest trap attack uses that relation and saturate once another strategy becomes cheaper. Under heterogeneous per-node budgets, a max over nodes enters, and the threshold can be convex (two fraud nodes that can only be trapped through different relations give $\Btrap=\max(c_1,c_2)$). Monotonicity and homogeneity hold in every case, and they suffice for robust planning.

\begin{corollary}[Robust planning, maximin choice, and hardening]
\label{cor:robust}
(a) If $c'\ge c$ componentwise, then $\Theta_B(c')\subseteq\Theta_B(c)$, so every detector's worst-case performance under $c'$ is at least its worst-case performance under $c$. Since $\Theta_{\rho B}(c)=\Theta_B(c/\rho)$, a detector whose worst case is $A$ at planning costs $c/\rho$ and budget $B$ keeps worst case at least $A$ for every true cost vector $c'\ge c/\rho$ and, at costs $c$, for every budget up to $\rho B$. For a gang, the smallest threshold over any polytope of cost vectors is attained at one of its vertices. (b) If the true costs are only known to lie in a set $U$ with a componentwise-minimal element $\underline c$ (a box, for instance), then $\min_{c\in U}A_B(f;c)=A_B(f;\underline c)$ for every detector $f$, so the maximin choice over $U$ among any set of detectors is the one with the best worst case at $\underline c$. (c) If some cheapest trap policy at $c$ uses no action on relation $r$, raising the costs of $r$ leaves $\Btrap(f;c)$ unchanged: hardening a relation raises the threshold only if every cheapest trap policy uses it.
\end{corollary}

Overestimating costs can only overestimate robustness; planning at the lower corner of a cost box certifies the whole box, and the price of robustness is the performance lost at the nominal costs.

\paragraph{A multi-relational CSBM.}
To make the envelope explicit, we extend the contextual stochastic block model \citep{deshpande2018csbm,baranwal2021graph,ma2022homophily}. A node's projected feature is $x_v\sim\mathcal N(\pm m,1)$ ($+$ for fraud). Under relation $r$, a fraudulent node has $d_r$ neighbors, a fraction $p_r$ of them fraudulent, and a benign node has $d^0_r$ neighbors, a fraction $q_r\le p_r$ fraudulent, independently across relations. A linear relational detector scores $z_v=w_0x_v+\sum_rw_ra_{v,r}$, with $a_{v,r}$ the mean of $x_u$ over $\mathcal N_r(v)$; injected neighbors have mean $-(m+\delta)$, $\delta\ge0$. Under an attack $\theta$, the class moments are explicit (\cref{apd:proofs}), and with $\Delta(\theta)=\mu_1(\theta)-\mu_0$,
\begin{equation}
\label{eq:auc}
\AUC(w;\theta)=\Phi(\zeta(w;\theta)),\quad \zeta=\frac{w^\top\Delta(\theta)}{\sqrt{w^\top(\Sigma_1(\theta)+\Sigma_0)w}}.
\end{equation}
The clean-optimal (Fisher) weights are $w^\star\propto(\Sigma_1(0)+\Sigma_0)^{-1}\Delta(0)$, and the graph-free detector $e_0$ has $\zeta_0=\sqrt2m$ under every attack.

In the base configuration of \cref{sec:exp} (an identity, a behavior, and a dense relation), the cheapest trap attack against $w^\star$ deletes all $p_{\mathrm{id}}d_{\mathrm{id}}=\TBTslope$ identity accomplices and injects $\TBTbeh$ behavior edges, so that $\Btrap=\TBTslope\,\cid+\TBTint$ until a camouflage of the behavior and dense relations becomes cheaper, at $\TBTsat$ (\cref{fig:main}a). The trap is bought by severing ties: when the absence of ties is the signal, deletion is the cheapest manipulation.

\subsection{Exposure: a lower bound}
\begin{proposition}[Exposure]
\label{prop:exposure}
Let $L_j(B)=\sup_{\theta\in\Theta_B(c)}(-\partial\zeta/\partial\theta_j)_+$. Then $\Btrap(w;c)\ge\sup\{B:\ B\max_jL_j(B)/c_j<\zeta(w;0)-\zeta_0\}$. At $\theta=0$, the injection exposure in relation $r$ is $w_r(2p_rm+\delta)/(s(w)\,\cin_rd_r)+O(w_r^2d_r^{-2})$, with $s(w)^2=w^\top(\Sigma_1(0)+\Sigma_0)w$.
\end{proposition}

To first order, the adversary spends its whole budget on the action of largest exposure, reliance $w_r$ over effective cost $\cin_rd_r$; a cheap relation is safe only if it is dense, and a dense relation is useful only if it is informative, so danger peaks at intermediate degrees. The bound extends beyond linear detectors.

\begin{proposition}[Exposure of message passing]
\label{prop:gnn}
Let $h_v=\sigma(W_0x_v+\sum_rW_r\bar x_{v,r})$ with $\bar x_{v,r}$ the mean of $x_u$ over $\mathcal N_r(v)$, and let the score $s_v=g(h_v)$ be $\Lambda$-Lipschitz in the pre-activation. An attack with $b_r$ injections and $k_r$ deletions changes $s_v$ by at most $\Lambda\sum_r\|W_r\|_2\Delta_{v,r}$, where $\Delta_{v,r}=(b_r+k_r)D_{v,r}/(d_r(v)-k_r+b_r)$ if $r$ keeps at least one neighbor of $v$, $D_{v,r}$ bounding $\|x_u-\bar x_{v,r}\|$ over the edited endpoints (with $\bar x_{v,r}=0$ for an isolated node), and $\Delta_{v,r}=\|\bar x_{v,r}\|$ if the attack empties $r$ (the empty aggregate is $0$). For one unit of budget, the exposure of relation $r$ is at most $\Lambda\|W_r\|_2D_{v,r}/(\cin_r(d_r(v)+1))$ for injections, and $\Lambda\|W_r\|_2D_{v,r}/(\cdel_r(d_r(v)-1))$ for deletions if $d_r(v)\ge2$, or $\Lambda\|W_r\|_2\|\bar x_{v,r}\|/\cdel_r$ if $d_r(v)=1$.
\end{proposition}

Reliance becomes the relation's operator norm and degree enters exactly as in the linear model; stacking layers multiplies the Lipschitz constants. The bound is per node, and low-degree nodes are the weak spot: a node without neighbors on $r$ has exposure $\Lambda\|W_r\|_2D_{v,r}/c_r$, since one injection replaces its empty aggregate. Gating a relation by degree removes this weak spot.

A degree gate lets relation $r$ enter $h_v$ only if $d_r(v)\ge\tau_r$; it replaces $d_r(v)$ by at least $\tau_r$ in the bound of \cref{prop:gnn} for every open node, and an attacker must pay $\cdel_r(d_r(v)-\tau_r+1)$ to close a gate and $\cin_r(\tau_r-d_r(v))$ to open one (\cref{prop:gate} in \cref{apd:proofs}). Gating trades coverage for exposure; the next result shows that, when part of the population is isolated on a relation, which is the regime of FDCompCN and YelpChi, gating is strictly better than anything relation selection or reweighting can do.

\begin{theorem}[Gating strictly beats selection and reweighting]
\label{thm:gatedom}
In the CSBM with one relation $r$, let a fraction $\pi_0\in(0,1)$ of the nodes of each class be isolated on $r$ and the others have degree $d$, with $p>q$. Let the attacker afford $b=\lfloor B/\cin\rfloor$ injections per node, with $1\le b<\bar b_d:=d\,(\mu_f-\mu_b)/(\mu_b+m+\delta)$, where $\mu_f=(2p-1)m$ and $\mu_b=(2q-1)m$, and no gate-closing deletion ($B<\cdel(d-\tau+1)$). Consider linear detectors $s_v=w_0x_v+w\,\bar x_{v,r}\mathbb 1[d_r(v)\ge\tau]$ with $w\ge0$. For every gate $\tau$ with $b<\tau\le d$: (a) for every $w>0$, the worst-case AUC of the gated detector is strictly larger than that of the ungated one ($\tau=0$) with the same weights; (b) the derivative of the gated worst-case AUC in $w$ at $w=0$ is proportional to $(1-\pi_0)(\mu'_d-\mu_b)>0$, with $\mu'_d=(d\mu_f-b(m+\delta))/(d+b)$, so some $w>0$ makes the gated detector strictly better than dropping the relation ($w=0$). Hence $\sup_w A_{\mathrm{gated}}>\max\{\sup_{w\ge0}A_{\mathrm{ungated}},A_{\mathrm{drop}}\}$. (c) The window is sharp: if $b\ge\tau$, the gated and ungated detectors with the same weights have the same worst-case AUC, and if $\tau>d$, the gated detector is never better than dropping the relation.
\end{theorem}

Selection keeps or drops a relation for every node, and reweighting scales it for every node; neither can separate isolated nodes, on which one injection replaces an empty aggregate, from connected nodes, on which it is diluted by $d$ neighbors. A gate that the attacker cannot open within budget does. The closed-form check in \cref{tab:gatetheory} finds the best ungated weight to be $w=0$ (dropping the relation) for every isolated fraction $\pi_0\ge0.5$ in our grid, while gating keeps a strict gain; exact attacks on CSBM graphs confirm it for RLR, and only where $b<\tau$ (\cref{tab:gategraph}); by part (c), the window cannot be widened. On real graphs, validated gating beats validated reweighting and dropping on exactly the relations where the conditions hold, informative relations on which part of the fraud is isolated, and only while the gate cannot be opened: YelpChi's R-U-R (\GRisoYelp\% of fraud isolated; up to \GRmaxYelp{} AUC) and S-FFSD's T-S-T (\GRisoSffsd\%; up to \GRmaxSffsd{}), \GRwinRels{} of the \GRrelsTotal{} relations tested; none of its \GRwinClosed{} wins occurs once the gate can be opened, and validation errors cost up to \GRlossMax{} elsewhere (\cref{tab:gatereal}). The theorem explains a mechanism; it does not promise a gain where no relation is both informative and partly isolated. \cref{sec:cars} chooses $\tau_r$ from data.

\subsection{Abandoning the graph and robust weights}
Let $\bar b_r=\frac{2m(p_r-q_r)d_r}{2q_rm+\delta}$ and $\bar k_r=\frac{(p_r-q_r)d_r}{1-q_r}$ be the numbers of injections and deletions that neutralize relation $r$ ($\Delta_r=0$), $\kappa_r=\min(\cin_r\bar b_r,\cdel_r\bar k_r)$ its \emph{neutralization cost}, $\zeta_r=\Delta_r(0)/\sigma_r$ its clean signal-to-noise ratio, and $\zeta_\star^2=\zeta_0^2+\sum_r\zeta_r^2$.

\begin{theorem}[Abandonment and robust weights]
\label{thm:abandon}
(a) In the CSBM, no detector beats the graph-free one under attack once $B\ge\sum_r\kappa_r$: $\Bab:=\sup_w\Btrap(w;c)\le\sum_r\kappa_r$. (b) If each action lowers $\Delta_r$ at the constant rate that neutralizes $r$ and leaves variances unchanged (linear response), then $\Bab=\sum_r\kappa_r$, $\Btrap(w^\star)=\zeta_\star(\zeta_\star-\zeta_0)\min_{r:p_r>q_r}\kappa_r/\zeta_r^2$, and the minimax detector clips the Fisher weights at caps proportional to neutralization cost: $w_r=\min(\lambda\Delta_r/\sigma_r^2,\,t\kappa_r/\Delta_r)$ for some $\lambda,t\ge0$.
\end{theorem}

Part (a) is exact: the neutralization set of each relation is a half-space in $(b_r,k_r)$, so its cheapest point is a vertex. The gap between the trap of the clean-optimal detector, governed by its weakest link, and abandonment, which requires neutralizing every relation, is the price of naive trust: in between, structure still has value, but only for a detector that discounts each relation by its exposure. Numerically, the abandonment threshold lies within \TBABgap{} of $\sum_r\kappa_r$ at every cost (\cref{fig:main}a). Gangs that share a budget drive the AUC down to the lower convex envelope of the per-node attack curve (\cref{prop:gang} in \cref{apd:proofs}).

\subsection{Exact worst case of linear relational models}
\begin{proposition}[Exact attack]
\label{prop:exact}
For relational logistic regression $s_v=w_0^\top x_v+\sum_rw_r^\top\bar x_{v,r}$ with costs on a grid of step $\epsilon$, the target-side worst case under \eqref{eq:budget} is computed exactly in $O(R\,(B/\epsilon)^2+\sum_r b^{\max}_rk^{\max}_r)$ per target: for fixed counts, the optimal edits inject the candidates with the smallest projections $w_r^\top x_u$ and delete the accomplices with the largest, and a dynamic program allocates the budget across relations.
\end{proposition}
Because the dynamic program returns the worst score of every target for every budget on the grid at once, the separable worst-case AUC curve $B\mapsto A(B)$ of a linear relational model, gated or not, and hence its trap threshold, are computed exactly in one pass per target; relation subsets give the per-relation and joint neutralization budgets on real graphs.

\subsection{Certified evaluation and validated selection}
For nonlinear detectors, an attack only shows that the true worst case is \emph{at most} what it achieves. A lower bound needs per-node certificates, and a guarantee for any selection rule needs to account for the validation sample.

\begin{proposition}[Certified AUC]
\label{prop:cert}
Under the target-side model (each fraudulent node edits its own ties; benign scores fixed), let $\ell_v\le\min_{\theta\in\Theta_B(c)}s_v(\theta)$ for every fraudulent test node. Then the true worst-case AUC satisfies $A_{\mathrm{cert}}\le A_{\mathrm{true}}(B)\le A_{\mathrm{eval}}$, where $A_{\mathrm{cert}}=\frac1{n_+n_-}\sum_{v,w}\mathbb 1[\ell_v>s_w]$ and $A_{\mathrm{eval}}$ is the AUC under any evaluated attack. In particular, if $|s_v(\theta)-s_v(0)|\le\Delta_v$ on $\Theta_B(c)$ (\cref{prop:gnn}), the AUC drops by at most the fraction of fraud--benign pairs whose clean margin $s_v-s_w$ is at most $\Delta_v$. For mean aggregation, each coordinate of $W_r\bar x'_{v,r}$ after $b$ injections and $k$ deletions lies in $[(S_j-\mathrm{top}_k(\mathrm{acc})_j+\mathrm{bot}_b(\mathcal C)_j)/(d-k+b),\ (S_j-\mathrm{bot}_k(\mathrm{acc})_j+\mathrm{top}_b(\mathcal C)_j)/(d-k+b)]$, with $S=\sum_{u\in\mathcal N_r(v)}W_rx_u$ and the top and bottom sums taken per coordinate over accomplices and candidates; if the attack empties the relation ($d-k+b=0$), the aggregate is exactly $0$; these intervals propagate through R-GCN layers and through tree ensembles.
\end{proposition}

For linear relational models $A_{\mathrm{cert}}=A_{\mathrm{true}}$ by \cref{prop:exact}; for R-GCN and gradient-boosted trees the bracket can be wide, and its width measures how much of a robustness claim rests on the attacks.

\begin{proposition}[Validated selection]
\label{prop:val}
Let a rule pick, among $K$ fixed candidate detectors, the one with the highest evaluated worst-case AUC on $n_+$ fraudulent and $n_-$ benign validation nodes drawn i.i.d.\ from the evaluated population, under a target-side evaluation in which each node's attacked score depends only on that node. With $\varepsilon=\sqrt{(1/n_++1/n_-)\log(2K/\delta)/2}$, with probability at least $1-\delta$ the picked detector's population evaluated worst-case AUC is at least that of every candidate minus $2\varepsilon$; in particular, at least that of the graph-free candidate minus $2\varepsilon$.
\end{proposition}

The guarantee is relative to the evaluated attacks, or to the true worst case if certified scores replace evaluated ones, and it degrades with $\log K$: a selection rule that checks few, well-chosen candidates pays less for validation than one that searches many configurations. This is the role of the plug-in in \cref{sec:cars}.

\section{Cost-Aware Relation Selection}
\label{sec:cars}

\cref{thm:abandon}(b) says what a robust linear detector does with a relation: keep it at its clean weight if its exposure is low, clip it otherwise, and drop it when it is cheap to neutralize relative to its signal; \cref{prop:gate} adds that exposure is a property of each node's degree. \cars{} turns these results into a training-free defense (\cref{alg:cars} in \cref{apd:setup}).

\textbf{(1) A moment plug-in of the threat.} From training labels, \cars{} fits a class-balanced relational logistic regression and summarizes every node by its ego projection and, for each relation $r$, by the mean of its neighbors' projections on the relation's own weight direction. It clusters training fraud and benign nodes by their log-degree vectors and estimates, per group, the class-conditional means and variances of these summaries, the mean projection $\alpha_r$ of accomplices, the projection $z_r$ of the most benign-looking candidates, and the fraud-neighbor rate $p_r$, which caps deletions. An attack with $b$ injections and $k$ deletions moves a fraud node's mean on $r$ to $(d\mu-k\alpha_r+bz_r)/(d-k+b)$, with edit counts capped by $\lfloor B/c\rfloor$, so a relation on which not one edit is affordable is untouchable. On the multi-relational CSBM these moments reduce to the closed forms of \cref{sec:theory}. Because attacks are per node, the worst-case AUC is the mixture over fraud groups of per-group worst cases, each a small constrained program.

\textbf{(2) Selection and gates.} \cars{} evaluates every subset of informative relations with its clean-optimal linear detector and keeps the subset of highest plug-in worst-case AUC. \cars{}-D also chooses a degree threshold $\tau_r$ per relation from a grid, by coordinate ascent from the \cars{} solution; in the plug-in, the attacker may edit a relation while keeping its gate open, close it by deleting accomplices, or open it by injecting candidates. Plug-in models can misjudge real graphs, so \cars{}-D finally keeps, among the graph-free counterpart itself (LR, the MLP, or HGB-X), the \cars{} configuration, and the gated configuration, the one with the highest evaluated worst-case AUC on validation nodes under the full attack suite, exhaustive-greedy attack included. The plug-in's role is to propose few candidates: by \cref{prop:val}, the validation guarantee degrades with the number of candidates checked, and each check costs a full attack evaluation; since the counterpart is a candidate, the picked detector is, with high probability, at most $2\varepsilon$ below it.

\textbf{(3) Robust planning and adversarial training.} With uncertain costs, \cars{}-R plans at costs $c/\rho$, which by \cref{cor:robust} covers every true cost vector above $c/\rho$. \cars{}+AT trains a cost-aware adversarially trained model and keeps it instead of \cars{}-D only if its validation worst case is higher. \cars{} and \cars{}-D apply to any detector, trees included, take seconds to minutes, and generate no attack on training nodes.

\section{Experiments}
\label{sec:exp}

\input{exp.tex}

\section{Related Work}
\label{sec:related}
\input{related.tex}

\section{Conclusion and Limitations}
\label{sec:conclusion}
\input{conclusion.tex}

\section*{Impact Statement}
This paper studies how adversaries can manipulate multi-relational fraud detectors and how defenders can decide which relations to trust or harden. The thresholds, exposure measures, and exact attacks could in principle help an adversary find the cheapest manipulation; we judge the defensive value to dominate, since the attacks are standard white-box evaluations of the kind defenders already run, the exact attack applies only to linear models, and the main practical outputs are decisions to abstain from a relation, to gate it, or to make it harder to manipulate. Our cost profiles are scenarios rather than measurements, and our results should not be read as estimates of the vulnerability of any deployed system. The datasets are public benchmarks; S-FFSD is simulated, and no personal data was collected.

\bibliography{refs}
\bibliographystyle{icml2026}

\newpage
\appendix
\onecolumn
\input{appendix_proofs.tex}
\input{appendix_details.tex}
\input{appendix_threat.tex}
\input{appendix_results.tex}

\input{appendix_earlier.tex}

\end{document}

%% file: tables/numbers.tex
\newcommand{\GapDall}{.002}
\newcommand{\CellsDall}{45}
\newcommand{\OkDall}{45}

\newcommand{\NTrapBase}{58}
\newcommand{\NCellsBase}{90}
\newcommand{\MaxLoss}{.906}

\newcommand{\TBTslope}{3.2}
\newcommand{\TBTint}{1.45}
\newcommand{\TBTsat}{30.4}
\newcommand{\TBTbeh}{0.72}
\newcommand{\TBABgap}{9\%}
\newcommand{\TLBratioMin}{0.29}
\newcommand{\TLBratioMax}{0.69}
\newcommand{\TBTtwoLineGap}{0.1\%}
\newcommand{\TGlobalGap}{3\times10^{-12}}

\newcommand{\EXPviol}{0}
\newcommand{\EXPn}{15590}

\newcommand{\EXPrhoMax}{0.95}

\newcommand{\POISONcols}{lcccccc}
\newcommand{\SYBILcols}{lccccccccc}

%% file: tables/values.tex
\newcommand{\V}[3]{\csname tt@#1@#2@#3\endcsname}
\expandafter\def\csname tt@Amazon@AT-RLR@P1 B=2\endcsname{.935}
\expandafter\def\csname tt@Amazon@AT-RLR@clean@P1 B=2\endcsname{.967}
\expandafter\def\csname tt@Amazon@AT-RLR@P1 B=20\endcsname{.887}
\expandafter\def\csname tt@Amazon@AT-RLR@clean@P1 B=20\endcsname{.959}
\expandafter\def\csname tt@Amazon@AT-RLR@P2 B=20\endcsname{.901}
\expandafter\def\csname tt@Amazon@AT-RLR@clean@P2 B=20\endcsname{.965}
\expandafter\def\csname tt@Amazon@BWGNN@P1 B=2\endcsname{.967}
\expandafter\def\csname tt@Amazon@BWGNN@clean@P1 B=2\endcsname{.978}
\expandafter\def\csname tt@Amazon@BWGNN@P1 B=20\endcsname{.947}
\expandafter\def\csname tt@Amazon@BWGNN@clean@P1 B=20\endcsname{.978}
\expandafter\def\csname tt@Amazon@BWGNN@P2 B=20\endcsname{.946}
\expandafter\def\csname tt@Amazon@BWGNN@clean@P2 B=20\endcsname{.978}
\expandafter\def\csname tt@Amazon@CA-AT@P1 B=2\endcsname{.892}
\expandafter\def\csname tt@Amazon@CA-AT@clean@P1 B=2\endcsname{.976}
\expandafter\def\csname tt@Amazon@CA-AT@P1 B=20\endcsname{.877}
\expandafter\def\csname tt@Amazon@CA-AT@clean@P1 B=20\endcsname{.976}
\expandafter\def\csname tt@Amazon@CA-AT@P2 B=20\endcsname{.881}
\expandafter\def\csname tt@Amazon@CA-AT@clean@P2 B=20\endcsname{.977}
\expandafter\def\csname tt@Amazon@CA-HGB@P1 B=2\endcsname{.979}
\expandafter\def\csname tt@Amazon@CA-HGB@clean@P1 B=2\endcsname{.987}
\expandafter\def\csname tt@Amazon@CA-HGB@P1 B=20\endcsname{.960}
\expandafter\def\csname tt@Amazon@CA-HGB@clean@P1 B=20\endcsname{.987}
\expandafter\def\csname tt@Amazon@CA-HGB@P2 B=20\endcsname{.964}
\expandafter\def\csname tt@Amazon@CA-HGB@clean@P2 B=20\endcsname{.986}
\expandafter\def\csname tt@Amazon@CARE-GNN@P1 B=2\endcsname{.967}
\expandafter\def\csname tt@Amazon@CARE-GNN@clean@P1 B=2\endcsname{.974}
\expandafter\def\csname tt@Amazon@CARE-GNN@P1 B=20\endcsname{.961}
\expandafter\def\csname tt@Amazon@CARE-GNN@clean@P1 B=20\endcsname{.974}
\expandafter\def\csname tt@Amazon@CARE-GNN@P2 B=20\endcsname{.963}
\expandafter\def\csname tt@Amazon@CARE-GNN@clean@P2 B=20\endcsname{.974}
\expandafter\def\csname tt@Amazon@CARS+AT@P1 B=2\endcsname{.973}
\expandafter\def\csname tt@Amazon@CARS+AT@clean@P1 B=2\endcsname{.973}
\expandafter\def\csname tt@Amazon@CARS+AT@P1 B=20\endcsname{.973}
\expandafter\def\csname tt@Amazon@CARS+AT@clean@P1 B=20\endcsname{.973}
\expandafter\def\csname tt@Amazon@CARS+AT@P2 B=20\endcsname{.973}
\expandafter\def\csname tt@Amazon@CARS+AT@clean@P2 B=20\endcsname{.973}
\expandafter\def\csname tt@Amazon@CARS+AT-HGB@P1 B=2\endcsname{.976}
\expandafter\def\csname tt@Amazon@CARS+AT-HGB@clean@P1 B=2\endcsname{.979}
\expandafter\def\csname tt@Amazon@CARS+AT-HGB@P1 B=20\endcsname{.977}
\expandafter\def\csname tt@Amazon@CARS+AT-HGB@clean@P1 B=20\endcsname{.977}
\expandafter\def\csname tt@Amazon@CARS+AT-HGB@P2 B=20\endcsname{.977}
\expandafter\def\csname tt@Amazon@CARS+AT-HGB@clean@P2 B=20\endcsname{.977}
\expandafter\def\csname tt@Amazon@GNNGuard@P1 B=2\endcsname{.912}
\expandafter\def\csname tt@Amazon@GNNGuard@clean@P1 B=2\endcsname{.976}
\expandafter\def\csname tt@Amazon@GNNGuard@P1 B=20\endcsname{.867}
\expandafter\def\csname tt@Amazon@GNNGuard@clean@P1 B=20\endcsname{.976}
\expandafter\def\csname tt@Amazon@GNNGuard@P2 B=20\endcsname{.880}
\expandafter\def\csname tt@Amazon@GNNGuard@clean@P2 B=20\endcsname{.976}
\expandafter\def\csname tt@Amazon@HGB-CARS@P1 B=2\endcsname{.976}
\expandafter\def\csname tt@Amazon@HGB-CARS@clean@P1 B=2\endcsname{.982}
\expandafter\def\csname tt@Amazon@HGB-CARS@P1 B=20\endcsname{.977}
\expandafter\def\csname tt@Amazon@HGB-CARS@clean@P1 B=20\endcsname{.977}
\expandafter\def\csname tt@Amazon@HGB-CARS@P2 B=20\endcsname{.977}
\expandafter\def\csname tt@Amazon@HGB-CARS@clean@P2 B=20\endcsname{.977}
\expandafter\def\csname tt@Amazon@HGB-CARS-D@P1 B=2\endcsname{.975}
\expandafter\def\csname tt@Amazon@HGB-CARS-D@clean@P1 B=2\endcsname{.979}
\expandafter\def\csname tt@Amazon@HGB-CARS-D@P1 B=20\endcsname{.977}
\expandafter\def\csname tt@Amazon@HGB-CARS-D@clean@P1 B=20\endcsname{.977}
\expandafter\def\csname tt@Amazon@HGB-CARS-D@P2 B=20\endcsname{.977}
\expandafter\def\csname tt@Amazon@HGB-CARS-D@clean@P2 B=20\endcsname{.977}
\expandafter\def\csname tt@Amazon@HGB-CARS-Dp@P1 B=2\endcsname{.974}
\expandafter\def\csname tt@Amazon@HGB-CARS-Dp@clean@P1 B=2\endcsname{.984}
\expandafter\def\csname tt@Amazon@HGB-CARS-Dp@P1 B=20\endcsname{.951}
\expandafter\def\csname tt@Amazon@HGB-CARS-Dp@clean@P1 B=20\endcsname{.979}
\expandafter\def\csname tt@Amazon@HGB-CARS-Dp@P2 B=20\endcsname{.961}
\expandafter\def\csname tt@Amazon@HGB-CARS-Dp@clean@P2 B=20\endcsname{.980}
\expandafter\def\csname tt@Amazon@HGB-Graph@P1 B=2\endcsname{.971}
\expandafter\def\csname tt@Amazon@HGB-Graph@clean@P1 B=2\endcsname{.986}
\expandafter\def\csname tt@Amazon@HGB-Graph@P1 B=20\endcsname{.945}
\expandafter\def\csname tt@Amazon@HGB-Graph@clean@P1 B=20\endcsname{.986}
\expandafter\def\csname tt@Amazon@HGB-Graph@P2 B=20\endcsname{.951}
\expandafter\def\csname tt@Amazon@HGB-Graph@clean@P2 B=20\endcsname{.986}
\expandafter\def\csname tt@Amazon@HGB-X@P1 B=2\endcsname{.977}
\expandafter\def\csname tt@Amazon@HGB-X@clean@P1 B=2\endcsname{.977}
\expandafter\def\csname tt@Amazon@HGB-X@P1 B=20\endcsname{.977}
\expandafter\def\csname tt@Amazon@HGB-X@clean@P1 B=20\endcsname{.977}
\expandafter\def\csname tt@Amazon@HGB-X@P2 B=20\endcsname{.977}
\expandafter\def\csname tt@Amazon@HGB-X@clean@P2 B=20\endcsname{.977}
\expandafter\def\csname tt@Amazon@HGB-free@P1 B=2\endcsname{.977}
\expandafter\def\csname tt@Amazon@HGB-free@clean@P1 B=2\endcsname{.977}
\expandafter\def\csname tt@Amazon@HGB-free@P1 B=20\endcsname{.977}
\expandafter\def\csname tt@Amazon@HGB-free@clean@P1 B=20\endcsname{.977}
\expandafter\def\csname tt@Amazon@HGB-free@P2 B=20\endcsname{.977}
\expandafter\def\csname tt@Amazon@HGB-free@clean@P2 B=20\endcsname{.977}
\expandafter\def\csname tt@Amazon@LR@P1 B=2\endcsname{.969}
\expandafter\def\csname tt@Amazon@LR@clean@P1 B=2\endcsname{.969}
\expandafter\def\csname tt@Amazon@LR@P1 B=20\endcsname{.969}
\expandafter\def\csname tt@Amazon@LR@clean@P1 B=20\endcsname{.969}
\expandafter\def\csname tt@Amazon@LR@P2 B=20\endcsname{.969}
\expandafter\def\csname tt@Amazon@LR@clean@P2 B=20\endcsname{.969}
\expandafter\def\csname tt@Amazon@MLP@P1 B=2\endcsname{.975}
\expandafter\def\csname tt@Amazon@MLP@clean@P1 B=2\endcsname{.975}
\expandafter\def\csname tt@Amazon@MLP@P1 B=20\endcsname{.975}
\expandafter\def\csname tt@Amazon@MLP@clean@P1 B=20\endcsname{.975}
\expandafter\def\csname tt@Amazon@MLP@P2 B=20\endcsname{.975}
\expandafter\def\csname tt@Amazon@MLP@clean@P2 B=20\endcsname{.975}
\expandafter\def\csname tt@Amazon@R-GCN@P1 B=2\endcsname{.739}
\expandafter\def\csname tt@Amazon@R-GCN@clean@P1 B=2\endcsname{.976}
\expandafter\def\csname tt@Amazon@R-GCN@P1 B=20\endcsname{.588}
\expandafter\def\csname tt@Amazon@R-GCN@clean@P1 B=20\endcsname{.976}
\expandafter\def\csname tt@Amazon@R-GCN@P2 B=20\endcsname{.594}
\expandafter\def\csname tt@Amazon@R-GCN@clean@P2 B=20\endcsname{.976}
\expandafter\def\csname tt@Amazon@R-GCN-CARS@P1 B=2\endcsname{.943}
\expandafter\def\csname tt@Amazon@R-GCN-CARS@clean@P1 B=2\endcsname{.976}
\expandafter\def\csname tt@Amazon@R-GCN-CARS@P1 B=20\endcsname{.975}
\expandafter\def\csname tt@Amazon@R-GCN-CARS@clean@P1 B=20\endcsname{.975}
\expandafter\def\csname tt@Amazon@R-GCN-CARS@P2 B=20\endcsname{.975}
\expandafter\def\csname tt@Amazon@R-GCN-CARS@clean@P2 B=20\endcsname{.975}
\expandafter\def\csname tt@Amazon@R-GCN-CARS-D@P1 B=2\endcsname{.975}
\expandafter\def\csname tt@Amazon@R-GCN-CARS-D@clean@P1 B=2\endcsname{.975}
\expandafter\def\csname tt@Amazon@R-GCN-CARS-D@P1 B=20\endcsname{.975}
\expandafter\def\csname tt@Amazon@R-GCN-CARS-D@clean@P1 B=20\endcsname{.975}
\expandafter\def\csname tt@Amazon@R-GCN-CARS-D@P2 B=20\endcsname{.975}
\expandafter\def\csname tt@Amazon@R-GCN-CARS-D@clean@P2 B=20\endcsname{.975}
\expandafter\def\csname tt@Amazon@R-GCN-CARS-Dp@P1 B=2\endcsname{.967}
\expandafter\def\csname tt@Amazon@R-GCN-CARS-Dp@clean@P1 B=2\endcsname{.976}
\expandafter\def\csname tt@Amazon@R-GCN-CARS-Dp@P1 B=20\endcsname{.874}
\expandafter\def\csname tt@Amazon@R-GCN-CARS-Dp@clean@P1 B=20\endcsname{.973}
\expandafter\def\csname tt@Amazon@R-GCN-CARS-Dp@P2 B=20\endcsname{.909}
\expandafter\def\csname tt@Amazon@R-GCN-CARS-Dp@clean@P2 B=20\endcsname{.971}
\expandafter\def\csname tt@Amazon@R-GCN-free@P1 B=2\endcsname{.975}
\expandafter\def\csname tt@Amazon@R-GCN-free@clean@P1 B=2\endcsname{.975}
\expandafter\def\csname tt@Amazon@R-GCN-free@P1 B=20\endcsname{.975}
\expandafter\def\csname tt@Amazon@R-GCN-free@clean@P1 B=20\endcsname{.975}
\expandafter\def\csname tt@Amazon@R-GCN-free@P2 B=20\endcsname{.975}
\expandafter\def\csname tt@Amazon@R-GCN-free@clean@P2 B=20\endcsname{.975}
\expandafter\def\csname tt@Amazon@RLR@P1 B=2\endcsname{.865}
\expandafter\def\csname tt@Amazon@RLR@clean@P1 B=2\endcsname{.971}
\expandafter\def\csname tt@Amazon@RLR@P1 B=20\endcsname{.620}
\expandafter\def\csname tt@Amazon@RLR@clean@P1 B=20\endcsname{.971}
\expandafter\def\csname tt@Amazon@RLR@P2 B=20\endcsname{.674}
\expandafter\def\csname tt@Amazon@RLR@clean@P2 B=20\endcsname{.971}
\expandafter\def\csname tt@Amazon@RLR-CARS@P1 B=2\endcsname{.951}
\expandafter\def\csname tt@Amazon@RLR-CARS@clean@P1 B=2\endcsname{.972}
\expandafter\def\csname tt@Amazon@RLR-CARS@P1 B=20\endcsname{.969}
\expandafter\def\csname tt@Amazon@RLR-CARS@clean@P1 B=20\endcsname{.969}
\expandafter\def\csname tt@Amazon@RLR-CARS@P2 B=20\endcsname{.969}
\expandafter\def\csname tt@Amazon@RLR-CARS@clean@P2 B=20\endcsname{.969}
\expandafter\def\csname tt@Amazon@RLR-CARS-D@P1 B=2\endcsname{.969}
\expandafter\def\csname tt@Amazon@RLR-CARS-D@clean@P1 B=2\endcsname{.969}
\expandafter\def\csname tt@Amazon@RLR-CARS-D@P1 B=20\endcsname{.969}
\expandafter\def\csname tt@Amazon@RLR-CARS-D@clean@P1 B=20\endcsname{.969}
\expandafter\def\csname tt@Amazon@RLR-CARS-D@P2 B=20\endcsname{.969}
\expandafter\def\csname tt@Amazon@RLR-CARS-D@clean@P2 B=20\endcsname{.969}
\expandafter\def\csname tt@Amazon@RLR-CARS-Dp@P1 B=2\endcsname{.961}
\expandafter\def\csname tt@Amazon@RLR-CARS-Dp@clean@P1 B=2\endcsname{.971}
\expandafter\def\csname tt@Amazon@RLR-CARS-Dp@P1 B=20\endcsname{.601}
\expandafter\def\csname tt@Amazon@RLR-CARS-Dp@clean@P1 B=20\endcsname{.967}
\expandafter\def\csname tt@Amazon@RLR-CARS-Dp@P2 B=20\endcsname{.728}
\expandafter\def\csname tt@Amazon@RLR-CARS-Dp@clean@P2 B=20\endcsname{.967}
\expandafter\def\csname tt@Amazon@RLR-free@P1 B=2\endcsname{.969}
\expandafter\def\csname tt@Amazon@RLR-free@clean@P1 B=2\endcsname{.969}
\expandafter\def\csname tt@Amazon@RLR-free@P1 B=20\endcsname{.969}
\expandafter\def\csname tt@Amazon@RLR-free@clean@P1 B=20\endcsname{.969}
\expandafter\def\csname tt@Amazon@RLR-free@P2 B=20\endcsname{.969}
\expandafter\def\csname tt@Amazon@RLR-free@clean@P2 B=20\endcsname{.969}
\expandafter\def\csname tt@FDCompCN@AT-RLR@P1 B=2\endcsname{.014}
\expandafter\def\csname tt@FDCompCN@AT-RLR@clean@P1 B=2\endcsname{.619}
\expandafter\def\csname tt@FDCompCN@AT-RLR@P1 B=20\endcsname{.012}
\expandafter\def\csname tt@FDCompCN@AT-RLR@clean@P1 B=20\endcsname{.597}
\expandafter\def\csname tt@FDCompCN@AT-RLR@P2 B=20\endcsname{.012}
\expandafter\def\csname tt@FDCompCN@AT-RLR@clean@P2 B=20\endcsname{.605}
\expandafter\def\csname tt@FDCompCN@BWGNN@P1 B=2\endcsname{.093}
\expandafter\def\csname tt@FDCompCN@BWGNN@clean@P1 B=2\endcsname{.688}
\expandafter\def\csname tt@FDCompCN@BWGNN@P1 B=20\endcsname{.082}
\expandafter\def\csname tt@FDCompCN@BWGNN@clean@P1 B=20\endcsname{.688}
\expandafter\def\csname tt@FDCompCN@BWGNN@P2 B=20\endcsname{.082}
\expandafter\def\csname tt@FDCompCN@BWGNN@clean@P2 B=20\endcsname{.688}
\expandafter\def\csname tt@FDCompCN@CA-AT@P1 B=2\endcsname{.229}
\expandafter\def\csname tt@FDCompCN@CA-AT@clean@P1 B=2\endcsname{.644}
\expandafter\def\csname tt@FDCompCN@CA-AT@P1 B=20\endcsname{.166}
\expandafter\def\csname tt@FDCompCN@CA-AT@clean@P1 B=20\endcsname{.610}
\expandafter\def\csname tt@FDCompCN@CA-AT@P2 B=20\endcsname{.120}
\expandafter\def\csname tt@FDCompCN@CA-AT@clean@P2 B=20\endcsname{.581}
\expandafter\def\csname tt@FDCompCN@CA-HGB@P1 B=2\endcsname{.748}
\expandafter\def\csname tt@FDCompCN@CA-HGB@clean@P1 B=2\endcsname{.784}
\expandafter\def\csname tt@FDCompCN@CA-HGB@P1 B=20\endcsname{.521}
\expandafter\def\csname tt@FDCompCN@CA-HGB@clean@P1 B=20\endcsname{.783}
\expandafter\def\csname tt@FDCompCN@CA-HGB@P2 B=20\endcsname{.540}
\expandafter\def\csname tt@FDCompCN@CA-HGB@clean@P2 B=20\endcsname{.781}
\expandafter\def\csname tt@FDCompCN@CARE-GNN@P1 B=2\endcsname{.026}
\expandafter\def\csname tt@FDCompCN@CARE-GNN@clean@P1 B=2\endcsname{.697}
\expandafter\def\csname tt@FDCompCN@CARE-GNN@P1 B=20\endcsname{.015}
\expandafter\def\csname tt@FDCompCN@CARE-GNN@clean@P1 B=20\endcsname{.697}
\expandafter\def\csname tt@FDCompCN@CARE-GNN@P2 B=20\endcsname{.014}
\expandafter\def\csname tt@FDCompCN@CARE-GNN@clean@P2 B=20\endcsname{.697}
\expandafter\def\csname tt@FDCompCN@CARS+AT@P1 B=2\endcsname{.715}
\expandafter\def\csname tt@FDCompCN@CARS+AT@clean@P1 B=2\endcsname{.715}
\expandafter\def\csname tt@FDCompCN@CARS+AT@P1 B=20\endcsname{.715}
\expandafter\def\csname tt@FDCompCN@CARS+AT@clean@P1 B=20\endcsname{.715}
\expandafter\def\csname tt@FDCompCN@CARS+AT@P2 B=20\endcsname{.715}
\expandafter\def\csname tt@FDCompCN@CARS+AT@clean@P2 B=20\endcsname{.715}
\expandafter\def\csname tt@FDCompCN@CARS+AT-HGB@P1 B=2\endcsname{.789}
\expandafter\def\csname tt@FDCompCN@CARS+AT-HGB@clean@P1 B=2\endcsname{.790}
\expandafter\def\csname tt@FDCompCN@CARS+AT-HGB@P1 B=20\endcsname{.789}
\expandafter\def\csname tt@FDCompCN@CARS+AT-HGB@clean@P1 B=20\endcsname{.789}
\expandafter\def\csname tt@FDCompCN@CARS+AT-HGB@P2 B=20\endcsname{.789}
\expandafter\def\csname tt@FDCompCN@CARS+AT-HGB@clean@P2 B=20\endcsname{.789}
\expandafter\def\csname tt@FDCompCN@GNNGuard@P1 B=2\endcsname{.014}
\expandafter\def\csname tt@FDCompCN@GNNGuard@clean@P1 B=2\endcsname{.652}
\expandafter\def\csname tt@FDCompCN@GNNGuard@P1 B=20\endcsname{.011}
\expandafter\def\csname tt@FDCompCN@GNNGuard@clean@P1 B=20\endcsname{.652}
\expandafter\def\csname tt@FDCompCN@GNNGuard@P2 B=20\endcsname{.011}
\expandafter\def\csname tt@FDCompCN@GNNGuard@clean@P2 B=20\endcsname{.652}
\expandafter\def\csname tt@FDCompCN@HGB-CARS@P1 B=2\endcsname{.787}
\expandafter\def\csname tt@FDCompCN@HGB-CARS@clean@P1 B=2\endcsname{.788}
\expandafter\def\csname tt@FDCompCN@HGB-CARS@P1 B=20\endcsname{.789}
\expandafter\def\csname tt@FDCompCN@HGB-CARS@clean@P1 B=20\endcsname{.789}
\expandafter\def\csname tt@FDCompCN@HGB-CARS@P2 B=20\endcsname{.789}
\expandafter\def\csname tt@FDCompCN@HGB-CARS@clean@P2 B=20\endcsname{.789}
\expandafter\def\csname tt@FDCompCN@HGB-CARS-D@P1 B=2\endcsname{.789}
\expandafter\def\csname tt@FDCompCN@HGB-CARS-D@clean@P1 B=2\endcsname{.790}
\expandafter\def\csname tt@FDCompCN@HGB-CARS-D@P1 B=20\endcsname{.789}
\expandafter\def\csname tt@FDCompCN@HGB-CARS-D@clean@P1 B=20\endcsname{.789}
\expandafter\def\csname tt@FDCompCN@HGB-CARS-D@P2 B=20\endcsname{.789}
\expandafter\def\csname tt@FDCompCN@HGB-CARS-D@clean@P2 B=20\endcsname{.789}
\expandafter\def\csname tt@FDCompCN@HGB-CARS-Dp@P1 B=2\endcsname{.789}
\expandafter\def\csname tt@FDCompCN@HGB-CARS-Dp@clean@P1 B=2\endcsname{.790}
\expandafter\def\csname tt@FDCompCN@HGB-CARS-Dp@P1 B=20\endcsname{.773}
\expandafter\def\csname tt@FDCompCN@HGB-CARS-Dp@clean@P1 B=20\endcsname{.788}
\expandafter\def\csname tt@FDCompCN@HGB-CARS-Dp@P2 B=20\endcsname{.788}
\expandafter\def\csname tt@FDCompCN@HGB-CARS-Dp@clean@P2 B=20\endcsname{.788}
\expandafter\def\csname tt@FDCompCN@HGB-Graph@P1 B=2\endcsname{.473}
\expandafter\def\csname tt@FDCompCN@HGB-Graph@clean@P1 B=2\endcsname{.785}
\expandafter\def\csname tt@FDCompCN@HGB-Graph@P1 B=20\endcsname{.245}
\expandafter\def\csname tt@FDCompCN@HGB-Graph@clean@P1 B=20\endcsname{.785}
\expandafter\def\csname tt@FDCompCN@HGB-Graph@P2 B=20\endcsname{.250}
\expandafter\def\csname tt@FDCompCN@HGB-Graph@clean@P2 B=20\endcsname{.785}
\expandafter\def\csname tt@FDCompCN@HGB-X@P1 B=2\endcsname{.789}
\expandafter\def\csname tt@FDCompCN@HGB-X@clean@P1 B=2\endcsname{.789}
\expandafter\def\csname tt@FDCompCN@HGB-X@P1 B=20\endcsname{.789}
\expandafter\def\csname tt@FDCompCN@HGB-X@clean@P1 B=20\endcsname{.789}
\expandafter\def\csname tt@FDCompCN@HGB-X@P2 B=20\endcsname{.789}
\expandafter\def\csname tt@FDCompCN@HGB-X@clean@P2 B=20\endcsname{.789}
\expandafter\def\csname tt@FDCompCN@HGB-free@P1 B=2\endcsname{.789}
\expandafter\def\csname tt@FDCompCN@HGB-free@clean@P1 B=2\endcsname{.789}
\expandafter\def\csname tt@FDCompCN@HGB-free@P1 B=20\endcsname{.789}
\expandafter\def\csname tt@FDCompCN@HGB-free@clean@P1 B=20\endcsname{.789}
\expandafter\def\csname tt@FDCompCN@HGB-free@P2 B=20\endcsname{.789}
\expandafter\def\csname tt@FDCompCN@HGB-free@clean@P2 B=20\endcsname{.789}
\expandafter\def\csname tt@FDCompCN@LR@P1 B=2\endcsname{.680}
\expandafter\def\csname tt@FDCompCN@LR@clean@P1 B=2\endcsname{.680}
\expandafter\def\csname tt@FDCompCN@LR@P1 B=20\endcsname{.680}
\expandafter\def\csname tt@FDCompCN@LR@clean@P1 B=20\endcsname{.680}
\expandafter\def\csname tt@FDCompCN@LR@P2 B=20\endcsname{.680}
\expandafter\def\csname tt@FDCompCN@LR@clean@P2 B=20\endcsname{.680}
\expandafter\def\csname tt@FDCompCN@MLP@P1 B=2\endcsname{.715}
\expandafter\def\csname tt@FDCompCN@MLP@clean@P1 B=2\endcsname{.715}
\expandafter\def\csname tt@FDCompCN@MLP@P1 B=20\endcsname{.715}
\expandafter\def\csname tt@FDCompCN@MLP@clean@P1 B=20\endcsname{.715}
\expandafter\def\csname tt@FDCompCN@MLP@P2 B=20\endcsname{.715}
\expandafter\def\csname tt@FDCompCN@MLP@clean@P2 B=20\endcsname{.715}
\expandafter\def\csname tt@FDCompCN@R-GCN@P1 B=2\endcsname{.003}
\expandafter\def\csname tt@FDCompCN@R-GCN@clean@P1 B=2\endcsname{.631}
\expandafter\def\csname tt@FDCompCN@R-GCN@P1 B=20\endcsname{.002}
\expandafter\def\csname tt@FDCompCN@R-GCN@clean@P1 B=20\endcsname{.631}
\expandafter\def\csname tt@FDCompCN@R-GCN@P2 B=20\endcsname{.002}
\expandafter\def\csname tt@FDCompCN@R-GCN@clean@P2 B=20\endcsname{.631}
\expandafter\def\csname tt@FDCompCN@R-GCN-CARS@P1 B=2\endcsname{.669}
\expandafter\def\csname tt@FDCompCN@R-GCN-CARS@clean@P1 B=2\endcsname{.671}
\expandafter\def\csname tt@FDCompCN@R-GCN-CARS@P1 B=20\endcsname{.715}
\expandafter\def\csname tt@FDCompCN@R-GCN-CARS@clean@P1 B=20\endcsname{.715}
\expandafter\def\csname tt@FDCompCN@R-GCN-CARS@P2 B=20\endcsname{.715}
\expandafter\def\csname tt@FDCompCN@R-GCN-CARS@clean@P2 B=20\endcsname{.715}
\expandafter\def\csname tt@FDCompCN@R-GCN-CARS-D@P1 B=2\endcsname{.715}
\expandafter\def\csname tt@FDCompCN@R-GCN-CARS-D@clean@P1 B=2\endcsname{.715}
\expandafter\def\csname tt@FDCompCN@R-GCN-CARS-D@P1 B=20\endcsname{.715}
\expandafter\def\csname tt@FDCompCN@R-GCN-CARS-D@clean@P1 B=20\endcsname{.715}
\expandafter\def\csname tt@FDCompCN@R-GCN-CARS-D@P2 B=20\endcsname{.715}
\expandafter\def\csname tt@FDCompCN@R-GCN-CARS-D@clean@P2 B=20\endcsname{.715}
\expandafter\def\csname tt@FDCompCN@R-GCN-CARS-Dp@P1 B=2\endcsname{.666}
\expandafter\def\csname tt@FDCompCN@R-GCN-CARS-Dp@clean@P1 B=2\endcsname{.669}
\expandafter\def\csname tt@FDCompCN@R-GCN-CARS-Dp@P1 B=20\endcsname{.579}
\expandafter\def\csname tt@FDCompCN@R-GCN-CARS-Dp@clean@P1 B=20\endcsname{.710}
\expandafter\def\csname tt@FDCompCN@R-GCN-CARS-Dp@P2 B=20\endcsname{.707}
\expandafter\def\csname tt@FDCompCN@R-GCN-CARS-Dp@clean@P2 B=20\endcsname{.710}
\expandafter\def\csname tt@FDCompCN@R-GCN-free@P1 B=2\endcsname{.715}
\expandafter\def\csname tt@FDCompCN@R-GCN-free@clean@P1 B=2\endcsname{.715}
\expandafter\def\csname tt@FDCompCN@R-GCN-free@P1 B=20\endcsname{.715}
\expandafter\def\csname tt@FDCompCN@R-GCN-free@clean@P1 B=20\endcsname{.715}
\expandafter\def\csname tt@FDCompCN@R-GCN-free@P2 B=20\endcsname{.715}
\expandafter\def\csname tt@FDCompCN@R-GCN-free@clean@P2 B=20\endcsname{.715}
\expandafter\def\csname tt@FDCompCN@RLR@P1 B=2\endcsname{.001}
\expandafter\def\csname tt@FDCompCN@RLR@clean@P1 B=2\endcsname{.656}
\expandafter\def\csname tt@FDCompCN@RLR@P1 B=20\endcsname{.001}
\expandafter\def\csname tt@FDCompCN@RLR@clean@P1 B=20\endcsname{.656}
\expandafter\def\csname tt@FDCompCN@RLR@P2 B=20\endcsname{.001}
\expandafter\def\csname tt@FDCompCN@RLR@clean@P2 B=20\endcsname{.656}
\expandafter\def\csname tt@FDCompCN@RLR-CARS@P1 B=2\endcsname{.674}
\expandafter\def\csname tt@FDCompCN@RLR-CARS@clean@P1 B=2\endcsname{.680}
\expandafter\def\csname tt@FDCompCN@RLR-CARS@P1 B=20\endcsname{.680}
\expandafter\def\csname tt@FDCompCN@RLR-CARS@clean@P1 B=20\endcsname{.680}
\expandafter\def\csname tt@FDCompCN@RLR-CARS@P2 B=20\endcsname{.680}
\expandafter\def\csname tt@FDCompCN@RLR-CARS@clean@P2 B=20\endcsname{.680}
\expandafter\def\csname tt@FDCompCN@RLR-CARS-D@P1 B=2\endcsname{.678}
\expandafter\def\csname tt@FDCompCN@RLR-CARS-D@clean@P1 B=2\endcsname{.682}
\expandafter\def\csname tt@FDCompCN@RLR-CARS-D@P1 B=20\endcsname{.680}
\expandafter\def\csname tt@FDCompCN@RLR-CARS-D@clean@P1 B=20\endcsname{.680}
\expandafter\def\csname tt@FDCompCN@RLR-CARS-D@P2 B=20\endcsname{.680}
\expandafter\def\csname tt@FDCompCN@RLR-CARS-D@clean@P2 B=20\endcsname{.680}
\expandafter\def\csname tt@FDCompCN@RLR-CARS-Dp@P1 B=2\endcsname{.673}
\expandafter\def\csname tt@FDCompCN@RLR-CARS-Dp@clean@P1 B=2\endcsname{.679}
\expandafter\def\csname tt@FDCompCN@RLR-CARS-Dp@P1 B=20\endcsname{.543}
\expandafter\def\csname tt@FDCompCN@RLR-CARS-Dp@clean@P1 B=20\endcsname{.680}
\expandafter\def\csname tt@FDCompCN@RLR-CARS-Dp@P2 B=20\endcsname{.677}
\expandafter\def\csname tt@FDCompCN@RLR-CARS-Dp@clean@P2 B=20\endcsname{.680}
\expandafter\def\csname tt@FDCompCN@RLR-free@P1 B=2\endcsname{.680}
\expandafter\def\csname tt@FDCompCN@RLR-free@clean@P1 B=2\endcsname{.680}
\expandafter\def\csname tt@FDCompCN@RLR-free@P1 B=20\endcsname{.680}
\expandafter\def\csname tt@FDCompCN@RLR-free@clean@P1 B=20\endcsname{.680}
\expandafter\def\csname tt@FDCompCN@RLR-free@P2 B=20\endcsname{.680}
\expandafter\def\csname tt@FDCompCN@RLR-free@clean@P2 B=20\endcsname{.680}
\expandafter\def\csname tt@S-FFSD@AT-RLR@P1 B=2\endcsname{.562}
\expandafter\def\csname tt@S-FFSD@AT-RLR@clean@P1 B=2\endcsname{.645}
\expandafter\def\csname tt@S-FFSD@AT-RLR@P1 B=20\endcsname{.476}
\expandafter\def\csname tt@S-FFSD@AT-RLR@clean@P1 B=20\endcsname{.642}
\expandafter\def\csname tt@S-FFSD@AT-RLR@P2 B=20\endcsname{.529}
\expandafter\def\csname tt@S-FFSD@AT-RLR@clean@P2 B=20\endcsname{.664}
\expandafter\def\csname tt@S-FFSD@BWGNN@P1 B=2\endcsname{.715}
\expandafter\def\csname tt@S-FFSD@BWGNN@clean@P1 B=2\endcsname{.867}
\expandafter\def\csname tt@S-FFSD@BWGNN@P1 B=20\endcsname{.402}
\expandafter\def\csname tt@S-FFSD@BWGNN@clean@P1 B=20\endcsname{.867}
\expandafter\def\csname tt@S-FFSD@BWGNN@P2 B=20\endcsname{.654}
\expandafter\def\csname tt@S-FFSD@BWGNN@clean@P2 B=20\endcsname{.867}
\expandafter\def\csname tt@S-FFSD@CA-AT@P1 B=2\endcsname{.686}
\expandafter\def\csname tt@S-FFSD@CA-AT@clean@P1 B=2\endcsname{.853}
\expandafter\def\csname tt@S-FFSD@CA-AT@P1 B=20\endcsname{.400}
\expandafter\def\csname tt@S-FFSD@CA-AT@clean@P1 B=20\endcsname{.758}
\expandafter\def\csname tt@S-FFSD@CA-AT@P2 B=20\endcsname{.628}
\expandafter\def\csname tt@S-FFSD@CA-AT@clean@P2 B=20\endcsname{.823}
\expandafter\def\csname tt@S-FFSD@CA-HGB@P1 B=2\endcsname{.702}
\expandafter\def\csname tt@S-FFSD@CA-HGB@clean@P1 B=2\endcsname{.864}
\expandafter\def\csname tt@S-FFSD@CA-HGB@P1 B=20\endcsname{.532}
\expandafter\def\csname tt@S-FFSD@CA-HGB@clean@P1 B=20\endcsname{.858}
\expandafter\def\csname tt@S-FFSD@CA-HGB@P2 B=20\endcsname{.639}
\expandafter\def\csname tt@S-FFSD@CA-HGB@clean@P2 B=20\endcsname{.866}
\expandafter\def\csname tt@S-FFSD@CARE-GNN@P1 B=2\endcsname{.654}
\expandafter\def\csname tt@S-FFSD@CARE-GNN@clean@P1 B=2\endcsname{.701}
\expandafter\def\csname tt@S-FFSD@CARE-GNN@P1 B=20\endcsname{.593}
\expandafter\def\csname tt@S-FFSD@CARE-GNN@clean@P1 B=20\endcsname{.701}
\expandafter\def\csname tt@S-FFSD@CARE-GNN@P2 B=20\endcsname{.624}
\expandafter\def\csname tt@S-FFSD@CARE-GNN@clean@P2 B=20\endcsname{.701}
\expandafter\def\csname tt@S-FFSD@CARS+AT@P1 B=2\endcsname{.782}
\expandafter\def\csname tt@S-FFSD@CARS+AT@clean@P1 B=2\endcsname{.782}
\expandafter\def\csname tt@S-FFSD@CARS+AT@P1 B=20\endcsname{.782}
\expandafter\def\csname tt@S-FFSD@CARS+AT@clean@P1 B=20\endcsname{.782}
\expandafter\def\csname tt@S-FFSD@CARS+AT@P2 B=20\endcsname{.856}
\expandafter\def\csname tt@S-FFSD@CARS+AT@clean@P2 B=20\endcsname{.856}
\expandafter\def\csname tt@S-FFSD@CARS+AT-HGB@P1 B=2\endcsname{.845}
\expandafter\def\csname tt@S-FFSD@CARS+AT-HGB@clean@P1 B=2\endcsname{.845}
\expandafter\def\csname tt@S-FFSD@CARS+AT-HGB@P1 B=20\endcsname{.845}
\expandafter\def\csname tt@S-FFSD@CARS+AT-HGB@clean@P1 B=20\endcsname{.845}
\expandafter\def\csname tt@S-FFSD@CARS+AT-HGB@P2 B=20\endcsname{.863}
\expandafter\def\csname tt@S-FFSD@CARS+AT-HGB@clean@P2 B=20\endcsname{.863}
\expandafter\def\csname tt@S-FFSD@GNNGuard@P1 B=2\endcsname{.567}
\expandafter\def\csname tt@S-FFSD@GNNGuard@clean@P1 B=2\endcsname{.868}
\expandafter\def\csname tt@S-FFSD@GNNGuard@P1 B=20\endcsname{.294}
\expandafter\def\csname tt@S-FFSD@GNNGuard@clean@P1 B=20\endcsname{.868}
\expandafter\def\csname tt@S-FFSD@GNNGuard@P2 B=20\endcsname{.529}
\expandafter\def\csname tt@S-FFSD@GNNGuard@clean@P2 B=20\endcsname{.868}
\expandafter\def\csname tt@S-FFSD@HGB-CARS@P1 B=2\endcsname{.449}
\expandafter\def\csname tt@S-FFSD@HGB-CARS@clean@P1 B=2\endcsname{.852}
\expandafter\def\csname tt@S-FFSD@HGB-CARS@P1 B=20\endcsname{.845}
\expandafter\def\csname tt@S-FFSD@HGB-CARS@clean@P1 B=20\endcsname{.845}
\expandafter\def\csname tt@S-FFSD@HGB-CARS@P2 B=20\endcsname{.863}
\expandafter\def\csname tt@S-FFSD@HGB-CARS@clean@P2 B=20\endcsname{.863}
\expandafter\def\csname tt@S-FFSD@HGB-CARS-D@P1 B=2\endcsname{.845}
\expandafter\def\csname tt@S-FFSD@HGB-CARS-D@clean@P1 B=2\endcsname{.845}
\expandafter\def\csname tt@S-FFSD@HGB-CARS-D@P1 B=20\endcsname{.845}
\expandafter\def\csname tt@S-FFSD@HGB-CARS-D@clean@P1 B=20\endcsname{.845}
\expandafter\def\csname tt@S-FFSD@HGB-CARS-D@P2 B=20\endcsname{.863}
\expandafter\def\csname tt@S-FFSD@HGB-CARS-D@clean@P2 B=20\endcsname{.863}
\expandafter\def\csname tt@S-FFSD@HGB-CARS-Dp@P1 B=2\endcsname{.555}
\expandafter\def\csname tt@S-FFSD@HGB-CARS-Dp@clean@P1 B=2\endcsname{.872}
\expandafter\def\csname tt@S-FFSD@HGB-CARS-Dp@P1 B=20\endcsname{.845}
\expandafter\def\csname tt@S-FFSD@HGB-CARS-Dp@clean@P1 B=20\endcsname{.845}
\expandafter\def\csname tt@S-FFSD@HGB-CARS-Dp@P2 B=20\endcsname{.863}
\expandafter\def\csname tt@S-FFSD@HGB-CARS-Dp@clean@P2 B=20\endcsname{.863}
\expandafter\def\csname tt@S-FFSD@HGB-Graph@P1 B=2\endcsname{.551}
\expandafter\def\csname tt@S-FFSD@HGB-Graph@clean@P1 B=2\endcsname{.869}
\expandafter\def\csname tt@S-FFSD@HGB-Graph@P1 B=20\endcsname{.258}
\expandafter\def\csname tt@S-FFSD@HGB-Graph@clean@P1 B=20\endcsname{.869}
\expandafter\def\csname tt@S-FFSD@HGB-Graph@P2 B=20\endcsname{.385}
\expandafter\def\csname tt@S-FFSD@HGB-Graph@clean@P2 B=20\endcsname{.869}
\expandafter\def\csname tt@S-FFSD@HGB-X@P1 B=2\endcsname{.845}
\expandafter\def\csname tt@S-FFSD@HGB-X@clean@P1 B=2\endcsname{.845}
\expandafter\def\csname tt@S-FFSD@HGB-X@P1 B=20\endcsname{.845}
\expandafter\def\csname tt@S-FFSD@HGB-X@clean@P1 B=20\endcsname{.845}
\expandafter\def\csname tt@S-FFSD@HGB-X@P2 B=20\endcsname{.845}
\expandafter\def\csname tt@S-FFSD@HGB-X@clean@P2 B=20\endcsname{.845}
\expandafter\def\csname tt@S-FFSD@HGB-free@P1 B=2\endcsname{.845}
\expandafter\def\csname tt@S-FFSD@HGB-free@clean@P1 B=2\endcsname{.845}
\expandafter\def\csname tt@S-FFSD@HGB-free@P1 B=20\endcsname{.845}
\expandafter\def\csname tt@S-FFSD@HGB-free@clean@P1 B=20\endcsname{.845}
\expandafter\def\csname tt@S-FFSD@HGB-free@P2 B=20\endcsname{.845}
\expandafter\def\csname tt@S-FFSD@HGB-free@clean@P2 B=20\endcsname{.845}
\expandafter\def\csname tt@S-FFSD@LR@P1 B=2\endcsname{.644}
\expandafter\def\csname tt@S-FFSD@LR@clean@P1 B=2\endcsname{.644}
\expandafter\def\csname tt@S-FFSD@LR@P1 B=20\endcsname{.644}
\expandafter\def\csname tt@S-FFSD@LR@clean@P1 B=20\endcsname{.644}
\expandafter\def\csname tt@S-FFSD@LR@P2 B=20\endcsname{.644}
\expandafter\def\csname tt@S-FFSD@LR@clean@P2 B=20\endcsname{.644}
\expandafter\def\csname tt@S-FFSD@MLP@P1 B=2\endcsname{.788}
\expandafter\def\csname tt@S-FFSD@MLP@clean@P1 B=2\endcsname{.788}
\expandafter\def\csname tt@S-FFSD@MLP@P1 B=20\endcsname{.788}
\expandafter\def\csname tt@S-FFSD@MLP@clean@P1 B=20\endcsname{.788}
\expandafter\def\csname tt@S-FFSD@MLP@P2 B=20\endcsname{.788}
\expandafter\def\csname tt@S-FFSD@MLP@clean@P2 B=20\endcsname{.788}
\expandafter\def\csname tt@S-FFSD@R-GCN@P1 B=2\endcsname{.325}
\expandafter\def\csname tt@S-FFSD@R-GCN@clean@P1 B=2\endcsname{.878}
\expandafter\def\csname tt@S-FFSD@R-GCN@P1 B=20\endcsname{.043}
\expandafter\def\csname tt@S-FFSD@R-GCN@clean@P1 B=20\endcsname{.878}
\expandafter\def\csname tt@S-FFSD@R-GCN@P2 B=20\endcsname{.143}
\expandafter\def\csname tt@S-FFSD@R-GCN@clean@P2 B=20\endcsname{.878}
\expandafter\def\csname tt@S-FFSD@R-GCN-CARS@P1 B=2\endcsname{.328}
\expandafter\def\csname tt@S-FFSD@R-GCN-CARS@clean@P1 B=2\endcsname{.843}
\expandafter\def\csname tt@S-FFSD@R-GCN-CARS@P1 B=20\endcsname{.788}
\expandafter\def\csname tt@S-FFSD@R-GCN-CARS@clean@P1 B=20\endcsname{.788}
\expandafter\def\csname tt@S-FFSD@R-GCN-CARS@P2 B=20\endcsname{.858}
\expandafter\def\csname tt@S-FFSD@R-GCN-CARS@clean@P2 B=20\endcsname{.858}
\expandafter\def\csname tt@S-FFSD@R-GCN-CARS-D@P1 B=2\endcsname{.788}
\expandafter\def\csname tt@S-FFSD@R-GCN-CARS-D@clean@P1 B=2\endcsname{.788}
\expandafter\def\csname tt@S-FFSD@R-GCN-CARS-D@P1 B=20\endcsname{.788}
\expandafter\def\csname tt@S-FFSD@R-GCN-CARS-D@clean@P1 B=20\endcsname{.788}
\expandafter\def\csname tt@S-FFSD@R-GCN-CARS-D@P2 B=20\endcsname{.858}
\expandafter\def\csname tt@S-FFSD@R-GCN-CARS-D@clean@P2 B=20\endcsname{.858}
\expandafter\def\csname tt@S-FFSD@R-GCN-CARS-Dp@P1 B=2\endcsname{.512}
\expandafter\def\csname tt@S-FFSD@R-GCN-CARS-Dp@clean@P1 B=2\endcsname{.879}
\expandafter\def\csname tt@S-FFSD@R-GCN-CARS-Dp@P1 B=20\endcsname{.788}
\expandafter\def\csname tt@S-FFSD@R-GCN-CARS-Dp@clean@P1 B=20\endcsname{.788}
\expandafter\def\csname tt@S-FFSD@R-GCN-CARS-Dp@P2 B=20\endcsname{.858}
\expandafter\def\csname tt@S-FFSD@R-GCN-CARS-Dp@clean@P2 B=20\endcsname{.858}
\expandafter\def\csname tt@S-FFSD@R-GCN-free@P1 B=2\endcsname{.788}
\expandafter\def\csname tt@S-FFSD@R-GCN-free@clean@P1 B=2\endcsname{.788}
\expandafter\def\csname tt@S-FFSD@R-GCN-free@P1 B=20\endcsname{.788}
\expandafter\def\csname tt@S-FFSD@R-GCN-free@clean@P1 B=20\endcsname{.788}
\expandafter\def\csname tt@S-FFSD@R-GCN-free@P2 B=20\endcsname{.788}
\expandafter\def\csname tt@S-FFSD@R-GCN-free@clean@P2 B=20\endcsname{.788}
\expandafter\def\csname tt@S-FFSD@RLR@P1 B=2\endcsname{.355}
\expandafter\def\csname tt@S-FFSD@RLR@clean@P1 B=2\endcsname{.660}
\expandafter\def\csname tt@S-FFSD@RLR@P1 B=20\endcsname{.005}
\expandafter\def\csname tt@S-FFSD@RLR@clean@P1 B=20\endcsname{.660}
\expandafter\def\csname tt@S-FFSD@RLR@P2 B=20\endcsname{.061}
\expandafter\def\csname tt@S-FFSD@RLR@clean@P2 B=20\endcsname{.660}
\expandafter\def\csname tt@S-FFSD@RLR-CARS@P1 B=2\endcsname{.504}
\expandafter\def\csname tt@S-FFSD@RLR-CARS@clean@P1 B=2\endcsname{.644}
\expandafter\def\csname tt@S-FFSD@RLR-CARS@P1 B=20\endcsname{.644}
\expandafter\def\csname tt@S-FFSD@RLR-CARS@clean@P1 B=20\endcsname{.644}
\expandafter\def\csname tt@S-FFSD@RLR-CARS@P2 B=20\endcsname{.664}
\expandafter\def\csname tt@S-FFSD@RLR-CARS@clean@P2 B=20\endcsname{.664}
\expandafter\def\csname tt@S-FFSD@RLR-CARS-D@P1 B=2\endcsname{.644}
\expandafter\def\csname tt@S-FFSD@RLR-CARS-D@clean@P1 B=2\endcsname{.644}
\expandafter\def\csname tt@S-FFSD@RLR-CARS-D@P1 B=20\endcsname{.644}
\expandafter\def\csname tt@S-FFSD@RLR-CARS-D@clean@P1 B=20\endcsname{.644}
\expandafter\def\csname tt@S-FFSD@RLR-CARS-D@P2 B=20\endcsname{.664}
\expandafter\def\csname tt@S-FFSD@RLR-CARS-D@clean@P2 B=20\endcsname{.664}
\expandafter\def\csname tt@S-FFSD@RLR-CARS-Dp@P1 B=2\endcsname{.401}
\expandafter\def\csname tt@S-FFSD@RLR-CARS-Dp@clean@P1 B=2\endcsname{.695}
\expandafter\def\csname tt@S-FFSD@RLR-CARS-Dp@P1 B=20\endcsname{.644}
\expandafter\def\csname tt@S-FFSD@RLR-CARS-Dp@clean@P1 B=20\endcsname{.644}
\expandafter\def\csname tt@S-FFSD@RLR-CARS-Dp@P2 B=20\endcsname{.664}
\expandafter\def\csname tt@S-FFSD@RLR-CARS-Dp@clean@P2 B=20\endcsname{.664}
\expandafter\def\csname tt@S-FFSD@RLR-free@P1 B=2\endcsname{.644}
\expandafter\def\csname tt@S-FFSD@RLR-free@clean@P1 B=2\endcsname{.644}
\expandafter\def\csname tt@S-FFSD@RLR-free@P1 B=20\endcsname{.644}
\expandafter\def\csname tt@S-FFSD@RLR-free@clean@P1 B=20\endcsname{.644}
\expandafter\def\csname tt@S-FFSD@RLR-free@P2 B=20\endcsname{.644}
\expandafter\def\csname tt@S-FFSD@RLR-free@clean@P2 B=20\endcsname{.644}
\expandafter\def\csname tt@Tolokers@AT-RLR@P1 B=2\endcsname{.698}
\expandafter\def\csname tt@Tolokers@AT-RLR@clean@P1 B=2\endcsname{.737}
\expandafter\def\csname tt@Tolokers@AT-RLR@P1 B=20\endcsname{.614}
\expandafter\def\csname tt@Tolokers@AT-RLR@clean@P1 B=20\endcsname{.732}
\expandafter\def\csname tt@Tolokers@AT-RLR@P2 B=20\endcsname{.697}
\expandafter\def\csname tt@Tolokers@AT-RLR@clean@P2 B=20\endcsname{.737}
\expandafter\def\csname tt@Tolokers@BWGNN@P1 B=2\endcsname{.729}
\expandafter\def\csname tt@Tolokers@BWGNN@clean@P1 B=2\endcsname{.831}
\expandafter\def\csname tt@Tolokers@BWGNN@P1 B=20\endcsname{.661}
\expandafter\def\csname tt@Tolokers@BWGNN@clean@P1 B=20\endcsname{.831}
\expandafter\def\csname tt@Tolokers@BWGNN@P2 B=20\endcsname{.726}
\expandafter\def\csname tt@Tolokers@BWGNN@clean@P2 B=20\endcsname{.831}
\expandafter\def\csname tt@Tolokers@CA-AT@P1 B=2\endcsname{.734}
\expandafter\def\csname tt@Tolokers@CA-AT@clean@P1 B=2\endcsname{.819}
\expandafter\def\csname tt@Tolokers@CA-AT@P1 B=20\endcsname{.603}
\expandafter\def\csname tt@Tolokers@CA-AT@clean@P1 B=20\endcsname{.813}
\expandafter\def\csname tt@Tolokers@CA-AT@P2 B=20\endcsname{.734}
\expandafter\def\csname tt@Tolokers@CA-AT@clean@P2 B=20\endcsname{.819}
\expandafter\def\csname tt@Tolokers@CA-HGB@P1 B=2\endcsname{.680}
\expandafter\def\csname tt@Tolokers@CA-HGB@clean@P1 B=2\endcsname{.818}
\expandafter\def\csname tt@Tolokers@CA-HGB@P1 B=20\endcsname{.469}
\expandafter\def\csname tt@Tolokers@CA-HGB@clean@P1 B=20\endcsname{.816}
\expandafter\def\csname tt@Tolokers@CA-HGB@P2 B=20\endcsname{.680}
\expandafter\def\csname tt@Tolokers@CA-HGB@clean@P2 B=20\endcsname{.818}
\expandafter\def\csname tt@Tolokers@CARE-GNN@P1 B=2\endcsname{.705}
\expandafter\def\csname tt@Tolokers@CARE-GNN@clean@P1 B=2\endcsname{.747}
\expandafter\def\csname tt@Tolokers@CARE-GNN@P1 B=20\endcsname{.622}
\expandafter\def\csname tt@Tolokers@CARE-GNN@clean@P1 B=20\endcsname{.747}
\expandafter\def\csname tt@Tolokers@CARE-GNN@P2 B=20\endcsname{.705}
\expandafter\def\csname tt@Tolokers@CARE-GNN@clean@P2 B=20\endcsname{.747}
\expandafter\def\csname tt@Tolokers@CARS+AT@P1 B=2\endcsname{.741}
\expandafter\def\csname tt@Tolokers@CARS+AT@clean@P1 B=2\endcsname{.746}
\expandafter\def\csname tt@Tolokers@CARS+AT@P1 B=20\endcsname{.740}
\expandafter\def\csname tt@Tolokers@CARS+AT@clean@P1 B=20\endcsname{.740}
\expandafter\def\csname tt@Tolokers@CARS+AT@P2 B=20\endcsname{.741}
\expandafter\def\csname tt@Tolokers@CARS+AT@clean@P2 B=20\endcsname{.746}
\expandafter\def\csname tt@Tolokers@CARS+AT-HGB@P1 B=2\endcsname{.734}
\expandafter\def\csname tt@Tolokers@CARS+AT-HGB@clean@P1 B=2\endcsname{.734}
\expandafter\def\csname tt@Tolokers@CARS+AT-HGB@P1 B=20\endcsname{.734}
\expandafter\def\csname tt@Tolokers@CARS+AT-HGB@clean@P1 B=20\endcsname{.734}
\expandafter\def\csname tt@Tolokers@CARS+AT-HGB@P2 B=20\endcsname{.734}
\expandafter\def\csname tt@Tolokers@CARS+AT-HGB@clean@P2 B=20\endcsname{.734}
\expandafter\def\csname tt@Tolokers@GNNGuard@P1 B=2\endcsname{.637}
\expandafter\def\csname tt@Tolokers@GNNGuard@clean@P1 B=2\endcsname{.808}
\expandafter\def\csname tt@Tolokers@GNNGuard@P1 B=20\endcsname{.459}
\expandafter\def\csname tt@Tolokers@GNNGuard@clean@P1 B=20\endcsname{.808}
\expandafter\def\csname tt@Tolokers@GNNGuard@P2 B=20\endcsname{.636}
\expandafter\def\csname tt@Tolokers@GNNGuard@clean@P2 B=20\endcsname{.808}
\expandafter\def\csname tt@Tolokers@HGB-CARS@P1 B=2\endcsname{.734}
\expandafter\def\csname tt@Tolokers@HGB-CARS@clean@P1 B=2\endcsname{.734}
\expandafter\def\csname tt@Tolokers@HGB-CARS@P1 B=20\endcsname{.734}
\expandafter\def\csname tt@Tolokers@HGB-CARS@clean@P1 B=20\endcsname{.734}
\expandafter\def\csname tt@Tolokers@HGB-CARS@P2 B=20\endcsname{.734}
\expandafter\def\csname tt@Tolokers@HGB-CARS@clean@P2 B=20\endcsname{.734}
\expandafter\def\csname tt@Tolokers@HGB-CARS-D@P1 B=2\endcsname{.734}
\expandafter\def\csname tt@Tolokers@HGB-CARS-D@clean@P1 B=2\endcsname{.734}
\expandafter\def\csname tt@Tolokers@HGB-CARS-D@P1 B=20\endcsname{.734}
\expandafter\def\csname tt@Tolokers@HGB-CARS-D@clean@P1 B=20\endcsname{.734}
\expandafter\def\csname tt@Tolokers@HGB-CARS-D@P2 B=20\endcsname{.734}
\expandafter\def\csname tt@Tolokers@HGB-CARS-D@clean@P2 B=20\endcsname{.734}
\expandafter\def\csname tt@Tolokers@HGB-CARS-Dp@P1 B=2\endcsname{.679}
\expandafter\def\csname tt@Tolokers@HGB-CARS-Dp@clean@P1 B=2\endcsname{.779}
\expandafter\def\csname tt@Tolokers@HGB-CARS-Dp@P1 B=20\endcsname{.734}
\expandafter\def\csname tt@Tolokers@HGB-CARS-Dp@clean@P1 B=20\endcsname{.734}
\expandafter\def\csname tt@Tolokers@HGB-CARS-Dp@P2 B=20\endcsname{.679}
\expandafter\def\csname tt@Tolokers@HGB-CARS-Dp@clean@P2 B=20\endcsname{.779}
\expandafter\def\csname tt@Tolokers@HGB-Graph@P1 B=2\endcsname{.524}
\expandafter\def\csname tt@Tolokers@HGB-Graph@clean@P1 B=2\endcsname{.818}
\expandafter\def\csname tt@Tolokers@HGB-Graph@P1 B=20\endcsname{.380}
\expandafter\def\csname tt@Tolokers@HGB-Graph@clean@P1 B=20\endcsname{.818}
\expandafter\def\csname tt@Tolokers@HGB-Graph@P2 B=20\endcsname{.524}
\expandafter\def\csname tt@Tolokers@HGB-Graph@clean@P2 B=20\endcsname{.818}
\expandafter\def\csname tt@Tolokers@HGB-X@P1 B=2\endcsname{.734}
\expandafter\def\csname tt@Tolokers@HGB-X@clean@P1 B=2\endcsname{.734}
\expandafter\def\csname tt@Tolokers@HGB-X@P1 B=20\endcsname{.734}
\expandafter\def\csname tt@Tolokers@HGB-X@clean@P1 B=20\endcsname{.734}
\expandafter\def\csname tt@Tolokers@HGB-X@P2 B=20\endcsname{.734}
\expandafter\def\csname tt@Tolokers@HGB-X@clean@P2 B=20\endcsname{.734}
\expandafter\def\csname tt@Tolokers@HGB-free@P1 B=2\endcsname{.734}
\expandafter\def\csname tt@Tolokers@HGB-free@clean@P1 B=2\endcsname{.734}
\expandafter\def\csname tt@Tolokers@HGB-free@P1 B=20\endcsname{.734}
\expandafter\def\csname tt@Tolokers@HGB-free@clean@P1 B=20\endcsname{.734}
\expandafter\def\csname tt@Tolokers@HGB-free@P2 B=20\endcsname{.734}
\expandafter\def\csname tt@Tolokers@HGB-free@clean@P2 B=20\endcsname{.734}
\expandafter\def\csname tt@Tolokers@LR@P1 B=2\endcsname{.722}
\expandafter\def\csname tt@Tolokers@LR@clean@P1 B=2\endcsname{.722}
\expandafter\def\csname tt@Tolokers@LR@P1 B=20\endcsname{.722}
\expandafter\def\csname tt@Tolokers@LR@clean@P1 B=20\endcsname{.722}
\expandafter\def\csname tt@Tolokers@LR@P2 B=20\endcsname{.722}
\expandafter\def\csname tt@Tolokers@LR@clean@P2 B=20\endcsname{.722}
\expandafter\def\csname tt@Tolokers@MLP@P1 B=2\endcsname{.740}
\expandafter\def\csname tt@Tolokers@MLP@clean@P1 B=2\endcsname{.740}
\expandafter\def\csname tt@Tolokers@MLP@P1 B=20\endcsname{.740}
\expandafter\def\csname tt@Tolokers@MLP@clean@P1 B=20\endcsname{.740}
\expandafter\def\csname tt@Tolokers@MLP@P2 B=20\endcsname{.740}
\expandafter\def\csname tt@Tolokers@MLP@clean@P2 B=20\endcsname{.740}
\expandafter\def\csname tt@Tolokers@R-GCN@P1 B=2\endcsname{.649}
\expandafter\def\csname tt@Tolokers@R-GCN@clean@P1 B=2\endcsname{.811}
\expandafter\def\csname tt@Tolokers@R-GCN@P1 B=20\endcsname{.416}
\expandafter\def\csname tt@Tolokers@R-GCN@clean@P1 B=20\endcsname{.811}
\expandafter\def\csname tt@Tolokers@R-GCN@P2 B=20\endcsname{.648}
\expandafter\def\csname tt@Tolokers@R-GCN@clean@P2 B=20\endcsname{.811}
\expandafter\def\csname tt@Tolokers@R-GCN-CARS@P1 B=2\endcsname{.740}
\expandafter\def\csname tt@Tolokers@R-GCN-CARS@clean@P1 B=2\endcsname{.740}
\expandafter\def\csname tt@Tolokers@R-GCN-CARS@P1 B=20\endcsname{.740}
\expandafter\def\csname tt@Tolokers@R-GCN-CARS@clean@P1 B=20\endcsname{.740}
\expandafter\def\csname tt@Tolokers@R-GCN-CARS@P2 B=20\endcsname{.740}
\expandafter\def\csname tt@Tolokers@R-GCN-CARS@clean@P2 B=20\endcsname{.740}
\expandafter\def\csname tt@Tolokers@R-GCN-CARS-D@P1 B=2\endcsname{.741}
\expandafter\def\csname tt@Tolokers@R-GCN-CARS-D@clean@P1 B=2\endcsname{.746}
\expandafter\def\csname tt@Tolokers@R-GCN-CARS-D@P1 B=20\endcsname{.740}
\expandafter\def\csname tt@Tolokers@R-GCN-CARS-D@clean@P1 B=20\endcsname{.740}
\expandafter\def\csname tt@Tolokers@R-GCN-CARS-D@P2 B=20\endcsname{.741}
\expandafter\def\csname tt@Tolokers@R-GCN-CARS-D@clean@P2 B=20\endcsname{.746}
\expandafter\def\csname tt@Tolokers@R-GCN-CARS-Dp@P1 B=2\endcsname{.733}
\expandafter\def\csname tt@Tolokers@R-GCN-CARS-Dp@clean@P1 B=2\endcsname{.775}
\expandafter\def\csname tt@Tolokers@R-GCN-CARS-Dp@P1 B=20\endcsname{.740}
\expandafter\def\csname tt@Tolokers@R-GCN-CARS-Dp@clean@P1 B=20\endcsname{.740}
\expandafter\def\csname tt@Tolokers@R-GCN-CARS-Dp@P2 B=20\endcsname{.733}
\expandafter\def\csname tt@Tolokers@R-GCN-CARS-Dp@clean@P2 B=20\endcsname{.775}
\expandafter\def\csname tt@Tolokers@R-GCN-free@P1 B=2\endcsname{.740}
\expandafter\def\csname tt@Tolokers@R-GCN-free@clean@P1 B=2\endcsname{.740}
\expandafter\def\csname tt@Tolokers@R-GCN-free@P1 B=20\endcsname{.740}
\expandafter\def\csname tt@Tolokers@R-GCN-free@clean@P1 B=20\endcsname{.740}
\expandafter\def\csname tt@Tolokers@R-GCN-free@P2 B=20\endcsname{.740}
\expandafter\def\csname tt@Tolokers@R-GCN-free@clean@P2 B=20\endcsname{.740}
\expandafter\def\csname tt@Tolokers@RLR@P1 B=2\endcsname{.529}
\expandafter\def\csname tt@Tolokers@RLR@clean@P1 B=2\endcsname{.768}
\expandafter\def\csname tt@Tolokers@RLR@P1 B=20\endcsname{.244}
\expandafter\def\csname tt@Tolokers@RLR@clean@P1 B=20\endcsname{.768}
\expandafter\def\csname tt@Tolokers@RLR@P2 B=20\endcsname{.529}
\expandafter\def\csname tt@Tolokers@RLR@clean@P2 B=20\endcsname{.768}
\expandafter\def\csname tt@Tolokers@RLR-CARS@P1 B=2\endcsname{.722}
\expandafter\def\csname tt@Tolokers@RLR-CARS@clean@P1 B=2\endcsname{.722}
\expandafter\def\csname tt@Tolokers@RLR-CARS@P1 B=20\endcsname{.722}
\expandafter\def\csname tt@Tolokers@RLR-CARS@clean@P1 B=20\endcsname{.722}
\expandafter\def\csname tt@Tolokers@RLR-CARS@P2 B=20\endcsname{.722}
\expandafter\def\csname tt@Tolokers@RLR-CARS@clean@P2 B=20\endcsname{.722}
\expandafter\def\csname tt@Tolokers@RLR-CARS-D@P1 B=2\endcsname{.722}
\expandafter\def\csname tt@Tolokers@RLR-CARS-D@clean@P1 B=2\endcsname{.722}
\expandafter\def\csname tt@Tolokers@RLR-CARS-D@P1 B=20\endcsname{.722}
\expandafter\def\csname tt@Tolokers@RLR-CARS-D@clean@P1 B=20\endcsname{.722}
\expandafter\def\csname tt@Tolokers@RLR-CARS-D@P2 B=20\endcsname{.722}
\expandafter\def\csname tt@Tolokers@RLR-CARS-D@clean@P2 B=20\endcsname{.722}
\expandafter\def\csname tt@Tolokers@RLR-CARS-Dp@P1 B=2\endcsname{.706}
\expandafter\def\csname tt@Tolokers@RLR-CARS-Dp@clean@P1 B=2\endcsname{.745}
\expandafter\def\csname tt@Tolokers@RLR-CARS-Dp@P1 B=20\endcsname{.722}
\expandafter\def\csname tt@Tolokers@RLR-CARS-Dp@clean@P1 B=20\endcsname{.722}
\expandafter\def\csname tt@Tolokers@RLR-CARS-Dp@P2 B=20\endcsname{.706}
\expandafter\def\csname tt@Tolokers@RLR-CARS-Dp@clean@P2 B=20\endcsname{.745}
\expandafter\def\csname tt@Tolokers@RLR-free@P1 B=2\endcsname{.722}
\expandafter\def\csname tt@Tolokers@RLR-free@clean@P1 B=2\endcsname{.722}
\expandafter\def\csname tt@Tolokers@RLR-free@P1 B=20\endcsname{.722}
\expandafter\def\csname tt@Tolokers@RLR-free@clean@P1 B=20\endcsname{.722}
\expandafter\def\csname tt@Tolokers@RLR-free@P2 B=20\endcsname{.722}
\expandafter\def\csname tt@Tolokers@RLR-free@clean@P2 B=20\endcsname{.722}
\expandafter\def\csname tt@YelpChi@AT-RLR@P1 B=2\endcsname{.366}
\expandafter\def\csname tt@YelpChi@AT-RLR@clean@P1 B=2\endcsname{.773}
\expandafter\def\csname tt@YelpChi@AT-RLR@P1 B=20\endcsname{.172}
\expandafter\def\csname tt@YelpChi@AT-RLR@clean@P1 B=20\endcsname{.765}
\expandafter\def\csname tt@YelpChi@AT-RLR@P2 B=20\endcsname{.597}
\expandafter\def\csname tt@YelpChi@AT-RLR@clean@P2 B=20\endcsname{.784}
\expandafter\def\csname tt@YelpChi@BWGNN@P1 B=2\endcsname{.593}
\expandafter\def\csname tt@YelpChi@BWGNN@clean@P1 B=2\endcsname{.902}
\expandafter\def\csname tt@YelpChi@BWGNN@P1 B=20\endcsname{.425}
\expandafter\def\csname tt@YelpChi@BWGNN@clean@P1 B=20\endcsname{.902}
\expandafter\def\csname tt@YelpChi@BWGNN@P2 B=20\endcsname{.868}
\expandafter\def\csname tt@YelpChi@BWGNN@clean@P2 B=20\endcsname{.902}
\expandafter\def\csname tt@YelpChi@CA-AT@P1 B=2\endcsname{.679}
\expandafter\def\csname tt@YelpChi@CA-AT@clean@P1 B=2\endcsname{.866}
\expandafter\def\csname tt@YelpChi@CA-AT@P1 B=20\endcsname{.591}
\expandafter\def\csname tt@YelpChi@CA-AT@clean@P1 B=20\endcsname{.846}
\expandafter\def\csname tt@YelpChi@CA-AT@P2 B=20\endcsname{.834}
\expandafter\def\csname tt@YelpChi@CA-AT@clean@P2 B=20\endcsname{.922}
\expandafter\def\csname tt@YelpChi@CA-HGB@P1 B=2\endcsname{.919}
\expandafter\def\csname tt@YelpChi@CA-HGB@clean@P1 B=2\endcsname{.962}
\expandafter\def\csname tt@YelpChi@CA-HGB@P1 B=20\endcsname{.710}
\expandafter\def\csname tt@YelpChi@CA-HGB@clean@P1 B=20\endcsname{.956}
\expandafter\def\csname tt@YelpChi@CA-HGB@P2 B=20\endcsname{.944}
\expandafter\def\csname tt@YelpChi@CA-HGB@clean@P2 B=20\endcsname{.966}
\expandafter\def\csname tt@YelpChi@CARE-GNN@P1 B=2\endcsname{.773}
\expandafter\def\csname tt@YelpChi@CARE-GNN@clean@P1 B=2\endcsname{.846}
\expandafter\def\csname tt@YelpChi@CARE-GNN@P1 B=20\endcsname{.720}
\expandafter\def\csname tt@YelpChi@CARE-GNN@clean@P1 B=20\endcsname{.846}
\expandafter\def\csname tt@YelpChi@CARE-GNN@P2 B=20\endcsname{.808}
\expandafter\def\csname tt@YelpChi@CARE-GNN@clean@P2 B=20\endcsname{.846}
\expandafter\def\csname tt@YelpChi@CARS+AT@P1 B=2\endcsname{.837}
\expandafter\def\csname tt@YelpChi@CARS+AT@clean@P1 B=2\endcsname{.837}
\expandafter\def\csname tt@YelpChi@CARS+AT@P1 B=20\endcsname{.837}
\expandafter\def\csname tt@YelpChi@CARS+AT@clean@P1 B=20\endcsname{.837}
\expandafter\def\csname tt@YelpChi@CARS+AT@P2 B=20\endcsname{.896}
\expandafter\def\csname tt@YelpChi@CARS+AT@clean@P2 B=20\endcsname{.896}
\expandafter\def\csname tt@YelpChi@CARS+AT-HGB@P1 B=2\endcsname{.935}
\expandafter\def\csname tt@YelpChi@CARS+AT-HGB@clean@P1 B=2\endcsname{.935}
\expandafter\def\csname tt@YelpChi@CARS+AT-HGB@P1 B=20\endcsname{.935}
\expandafter\def\csname tt@YelpChi@CARS+AT-HGB@clean@P1 B=20\endcsname{.935}
\expandafter\def\csname tt@YelpChi@CARS+AT-HGB@P2 B=20\endcsname{.952}
\expandafter\def\csname tt@YelpChi@CARS+AT-HGB@clean@P2 B=20\endcsname{.952}
\expandafter\def\csname tt@YelpChi@GNNGuard@P1 B=2\endcsname{.169}
\expandafter\def\csname tt@YelpChi@GNNGuard@clean@P1 B=2\endcsname{.917}
\expandafter\def\csname tt@YelpChi@GNNGuard@P1 B=20\endcsname{.080}
\expandafter\def\csname tt@YelpChi@GNNGuard@clean@P1 B=20\endcsname{.917}
\expandafter\def\csname tt@YelpChi@GNNGuard@P2 B=20\endcsname{.850}
\expandafter\def\csname tt@YelpChi@GNNGuard@clean@P2 B=20\endcsname{.917}
\expandafter\def\csname tt@YelpChi@HGB-CARS@P1 B=2\endcsname{.935}
\expandafter\def\csname tt@YelpChi@HGB-CARS@clean@P1 B=2\endcsname{.935}
\expandafter\def\csname tt@YelpChi@HGB-CARS@P1 B=20\endcsname{.935}
\expandafter\def\csname tt@YelpChi@HGB-CARS@clean@P1 B=20\endcsname{.935}
\expandafter\def\csname tt@YelpChi@HGB-CARS@P2 B=20\endcsname{.952}
\expandafter\def\csname tt@YelpChi@HGB-CARS@clean@P2 B=20\endcsname{.952}
\expandafter\def\csname tt@YelpChi@HGB-CARS-D@P1 B=2\endcsname{.935}
\expandafter\def\csname tt@YelpChi@HGB-CARS-D@clean@P1 B=2\endcsname{.935}
\expandafter\def\csname tt@YelpChi@HGB-CARS-D@P1 B=20\endcsname{.935}
\expandafter\def\csname tt@YelpChi@HGB-CARS-D@clean@P1 B=20\endcsname{.935}
\expandafter\def\csname tt@YelpChi@HGB-CARS-D@P2 B=20\endcsname{.952}
\expandafter\def\csname tt@YelpChi@HGB-CARS-D@clean@P2 B=20\endcsname{.952}
\expandafter\def\csname tt@YelpChi@HGB-CARS-Dp@P1 B=2\endcsname{.835}
\expandafter\def\csname tt@YelpChi@HGB-CARS-Dp@clean@P1 B=2\endcsname{.950}
\expandafter\def\csname tt@YelpChi@HGB-CARS-Dp@P1 B=20\endcsname{.935}
\expandafter\def\csname tt@YelpChi@HGB-CARS-Dp@clean@P1 B=20\endcsname{.935}
\expandafter\def\csname tt@YelpChi@HGB-CARS-Dp@P2 B=20\endcsname{.952}
\expandafter\def\csname tt@YelpChi@HGB-CARS-Dp@clean@P2 B=20\endcsname{.952}
\expandafter\def\csname tt@YelpChi@HGB-Graph@P1 B=2\endcsname{.442}
\expandafter\def\csname tt@YelpChi@HGB-Graph@clean@P1 B=2\endcsname{.960}
\expandafter\def\csname tt@YelpChi@HGB-Graph@P1 B=20\endcsname{.200}
\expandafter\def\csname tt@YelpChi@HGB-Graph@clean@P1 B=20\endcsname{.960}
\expandafter\def\csname tt@YelpChi@HGB-Graph@P2 B=20\endcsname{.924}
\expandafter\def\csname tt@YelpChi@HGB-Graph@clean@P2 B=20\endcsname{.960}
\expandafter\def\csname tt@YelpChi@HGB-X@P1 B=2\endcsname{.935}
\expandafter\def\csname tt@YelpChi@HGB-X@clean@P1 B=2\endcsname{.935}
\expandafter\def\csname tt@YelpChi@HGB-X@P1 B=20\endcsname{.935}
\expandafter\def\csname tt@YelpChi@HGB-X@clean@P1 B=20\endcsname{.935}
\expandafter\def\csname tt@YelpChi@HGB-X@P2 B=20\endcsname{.935}
\expandafter\def\csname tt@YelpChi@HGB-X@clean@P2 B=20\endcsname{.935}
\expandafter\def\csname tt@YelpChi@HGB-free@P1 B=2\endcsname{.935}
\expandafter\def\csname tt@YelpChi@HGB-free@clean@P1 B=2\endcsname{.935}
\expandafter\def\csname tt@YelpChi@HGB-free@P1 B=20\endcsname{.935}
\expandafter\def\csname tt@YelpChi@HGB-free@clean@P1 B=20\endcsname{.935}
\expandafter\def\csname tt@YelpChi@HGB-free@P2 B=20\endcsname{.935}
\expandafter\def\csname tt@YelpChi@HGB-free@clean@P2 B=20\endcsname{.935}
\expandafter\def\csname tt@YelpChi@LR@P1 B=2\endcsname{.771}
\expandafter\def\csname tt@YelpChi@LR@clean@P1 B=2\endcsname{.771}
\expandafter\def\csname tt@YelpChi@LR@P1 B=20\endcsname{.771}
\expandafter\def\csname tt@YelpChi@LR@clean@P1 B=20\endcsname{.771}
\expandafter\def\csname tt@YelpChi@LR@P2 B=20\endcsname{.771}
\expandafter\def\csname tt@YelpChi@LR@clean@P2 B=20\endcsname{.771}
\expandafter\def\csname tt@YelpChi@MLP@P1 B=2\endcsname{.836}
\expandafter\def\csname tt@YelpChi@MLP@clean@P1 B=2\endcsname{.836}
\expandafter\def\csname tt@YelpChi@MLP@P1 B=20\endcsname{.836}
\expandafter\def\csname tt@YelpChi@MLP@clean@P1 B=20\endcsname{.836}
\expandafter\def\csname tt@YelpChi@MLP@P2 B=20\endcsname{.836}
\expandafter\def\csname tt@YelpChi@MLP@clean@P2 B=20\endcsname{.836}
\expandafter\def\csname tt@YelpChi@R-GCN@P1 B=2\endcsname{.043}
\expandafter\def\csname tt@YelpChi@R-GCN@clean@P1 B=2\endcsname{.915}
\expandafter\def\csname tt@YelpChi@R-GCN@P1 B=20\endcsname{.009}
\expandafter\def\csname tt@YelpChi@R-GCN@clean@P1 B=20\endcsname{.915}
\expandafter\def\csname tt@YelpChi@R-GCN@P2 B=20\endcsname{.639}
\expandafter\def\csname tt@YelpChi@R-GCN@clean@P2 B=20\endcsname{.915}
\expandafter\def\csname tt@YelpChi@R-GCN-CARS@P1 B=2\endcsname{.836}
\expandafter\def\csname tt@YelpChi@R-GCN-CARS@clean@P1 B=2\endcsname{.836}
\expandafter\def\csname tt@YelpChi@R-GCN-CARS@P1 B=20\endcsname{.836}
\expandafter\def\csname tt@YelpChi@R-GCN-CARS@clean@P1 B=20\endcsname{.836}
\expandafter\def\csname tt@YelpChi@R-GCN-CARS@P2 B=20\endcsname{.897}
\expandafter\def\csname tt@YelpChi@R-GCN-CARS@clean@P2 B=20\endcsname{.897}
\expandafter\def\csname tt@YelpChi@R-GCN-CARS-D@P1 B=2\endcsname{.836}
\expandafter\def\csname tt@YelpChi@R-GCN-CARS-D@clean@P1 B=2\endcsname{.836}
\expandafter\def\csname tt@YelpChi@R-GCN-CARS-D@P1 B=20\endcsname{.836}
\expandafter\def\csname tt@YelpChi@R-GCN-CARS-D@clean@P1 B=20\endcsname{.836}
\expandafter\def\csname tt@YelpChi@R-GCN-CARS-D@P2 B=20\endcsname{.897}
\expandafter\def\csname tt@YelpChi@R-GCN-CARS-D@clean@P2 B=20\endcsname{.897}
\expandafter\def\csname tt@YelpChi@R-GCN-CARS-Dp@P1 B=2\endcsname{.750}
\expandafter\def\csname tt@YelpChi@R-GCN-CARS-Dp@clean@P1 B=2\endcsname{.889}
\expandafter\def\csname tt@YelpChi@R-GCN-CARS-Dp@P1 B=20\endcsname{.836}
\expandafter\def\csname tt@YelpChi@R-GCN-CARS-Dp@clean@P1 B=20\endcsname{.836}
\expandafter\def\csname tt@YelpChi@R-GCN-CARS-Dp@P2 B=20\endcsname{.896}
\expandafter\def\csname tt@YelpChi@R-GCN-CARS-Dp@clean@P2 B=20\endcsname{.896}
\expandafter\def\csname tt@YelpChi@R-GCN-free@P1 B=2\endcsname{.836}
\expandafter\def\csname tt@YelpChi@R-GCN-free@clean@P1 B=2\endcsname{.836}
\expandafter\def\csname tt@YelpChi@R-GCN-free@P1 B=20\endcsname{.836}
\expandafter\def\csname tt@YelpChi@R-GCN-free@clean@P1 B=20\endcsname{.836}
\expandafter\def\csname tt@YelpChi@R-GCN-free@P2 B=20\endcsname{.836}
\expandafter\def\csname tt@YelpChi@R-GCN-free@clean@P2 B=20\endcsname{.836}
\expandafter\def\csname tt@YelpChi@RLR@P1 B=2\endcsname{.181}
\expandafter\def\csname tt@YelpChi@RLR@clean@P1 B=2\endcsname{.798}
\expandafter\def\csname tt@YelpChi@RLR@P1 B=20\endcsname{.033}
\expandafter\def\csname tt@YelpChi@RLR@clean@P1 B=20\endcsname{.798}
\expandafter\def\csname tt@YelpChi@RLR@P2 B=20\endcsname{.546}
\expandafter\def\csname tt@YelpChi@RLR@clean@P2 B=20\endcsname{.798}
\expandafter\def\csname tt@YelpChi@RLR-CARS@P1 B=2\endcsname{.771}
\expandafter\def\csname tt@YelpChi@RLR-CARS@clean@P1 B=2\endcsname{.771}
\expandafter\def\csname tt@YelpChi@RLR-CARS@P1 B=20\endcsname{.771}
\expandafter\def\csname tt@YelpChi@RLR-CARS@clean@P1 B=20\endcsname{.771}
\expandafter\def\csname tt@YelpChi@RLR-CARS@P2 B=20\endcsname{.791}
\expandafter\def\csname tt@YelpChi@RLR-CARS@clean@P2 B=20\endcsname{.791}
\expandafter\def\csname tt@YelpChi@RLR-CARS-D@P1 B=2\endcsname{.771}
\expandafter\def\csname tt@YelpChi@RLR-CARS-D@clean@P1 B=2\endcsname{.771}
\expandafter\def\csname tt@YelpChi@RLR-CARS-D@P1 B=20\endcsname{.771}
\expandafter\def\csname tt@YelpChi@RLR-CARS-D@clean@P1 B=20\endcsname{.771}
\expandafter\def\csname tt@YelpChi@RLR-CARS-D@P2 B=20\endcsname{.792}
\expandafter\def\csname tt@YelpChi@RLR-CARS-D@clean@P2 B=20\endcsname{.792}
\expandafter\def\csname tt@YelpChi@RLR-CARS-Dp@P1 B=2\endcsname{.689}
\expandafter\def\csname tt@YelpChi@RLR-CARS-Dp@clean@P1 B=2\endcsname{.795}
\expandafter\def\csname tt@YelpChi@RLR-CARS-Dp@P1 B=20\endcsname{.771}
\expandafter\def\csname tt@YelpChi@RLR-CARS-Dp@clean@P1 B=20\endcsname{.771}
\expandafter\def\csname tt@YelpChi@RLR-CARS-Dp@P2 B=20\endcsname{.792}
\expandafter\def\csname tt@YelpChi@RLR-CARS-Dp@clean@P2 B=20\endcsname{.792}
\expandafter\def\csname tt@YelpChi@RLR-free@P1 B=2\endcsname{.771}
\expandafter\def\csname tt@YelpChi@RLR-free@clean@P1 B=2\endcsname{.771}
\expandafter\def\csname tt@YelpChi@RLR-free@P1 B=20\endcsname{.771}
\expandafter\def\csname tt@YelpChi@RLR-free@clean@P1 B=20\endcsname{.771}
\expandafter\def\csname tt@YelpChi@RLR-free@P2 B=20\endcsname{.771}
\expandafter\def\csname tt@YelpChi@RLR-free@clean@P2 B=20\endcsname{.771}
\expandafter\def\csname tt@Amazon@BWGNN@clean\endcsname{.978}
\expandafter\def\csname tt@Amazon@CARE-GNN@clean\endcsname{.974}
\expandafter\def\csname tt@Amazon@GNNGuard@clean\endcsname{.976}
\expandafter\def\csname tt@Amazon@HGB-Graph@clean\endcsname{.986}
\expandafter\def\csname tt@Amazon@HGB-X@clean\endcsname{.977}
\expandafter\def\csname tt@Amazon@LR@clean\endcsname{.969}
\expandafter\def\csname tt@Amazon@MLP@clean\endcsname{.975}
\expandafter\def\csname tt@Amazon@R-GCN@clean\endcsname{.976}
\expandafter\def\csname tt@Amazon@RLR@clean\endcsname{.971}
\expandafter\def\csname tt@FDCompCN@BWGNN@clean\endcsname{.688}
\expandafter\def\csname tt@FDCompCN@CARE-GNN@clean\endcsname{.697}
\expandafter\def\csname tt@FDCompCN@GNNGuard@clean\endcsname{.652}
\expandafter\def\csname tt@FDCompCN@HGB-Graph@clean\endcsname{.785}
\expandafter\def\csname tt@FDCompCN@HGB-X@clean\endcsname{.789}
\expandafter\def\csname tt@FDCompCN@LR@clean\endcsname{.680}
\expandafter\def\csname tt@FDCompCN@MLP@clean\endcsname{.715}
\expandafter\def\csname tt@FDCompCN@R-GCN@clean\endcsname{.631}
\expandafter\def\csname tt@FDCompCN@RLR@clean\endcsname{.656}
\expandafter\def\csname tt@S-FFSD@BWGNN@clean\endcsname{.867}
\expandafter\def\csname tt@S-FFSD@CARE-GNN@clean\endcsname{.701}
\expandafter\def\csname tt@S-FFSD@GNNGuard@clean\endcsname{.868}
\expandafter\def\csname tt@S-FFSD@HGB-Graph@clean\endcsname{.869}
\expandafter\def\csname tt@S-FFSD@HGB-X@clean\endcsname{.845}
\expandafter\def\csname tt@S-FFSD@LR@clean\endcsname{.644}
\expandafter\def\csname tt@S-FFSD@MLP@clean\endcsname{.788}
\expandafter\def\csname tt@S-FFSD@R-GCN@clean\endcsname{.878}
\expandafter\def\csname tt@S-FFSD@RLR@clean\endcsname{.660}
\expandafter\def\csname tt@Tolokers@BWGNN@clean\endcsname{.831}
\expandafter\def\csname tt@Tolokers@CARE-GNN@clean\endcsname{.747}
\expandafter\def\csname tt@Tolokers@GNNGuard@clean\endcsname{.808}
\expandafter\def\csname tt@Tolokers@HGB-Graph@clean\endcsname{.818}
\expandafter\def\csname tt@Tolokers@HGB-X@clean\endcsname{.734}
\expandafter\def\csname tt@Tolokers@LR@clean\endcsname{.722}
\expandafter\def\csname tt@Tolokers@MLP@clean\endcsname{.740}
\expandafter\def\csname tt@Tolokers@R-GCN@clean\endcsname{.811}
\expandafter\def\csname tt@Tolokers@RLR@clean\endcsname{.768}
\expandafter\def\csname tt@YelpChi@BWGNN@clean\endcsname{.902}
\expandafter\def\csname tt@YelpChi@CARE-GNN@clean\endcsname{.846}
\expandafter\def\csname tt@YelpChi@GNNGuard@clean\endcsname{.917}
\expandafter\def\csname tt@YelpChi@HGB-Graph@clean\endcsname{.960}
\expandafter\def\csname tt@YelpChi@HGB-X@clean\endcsname{.935}
\expandafter\def\csname tt@YelpChi@LR@clean\endcsname{.771}
\expandafter\def\csname tt@YelpChi@MLP@clean\endcsname{.836}
\expandafter\def\csname tt@YelpChi@R-GCN@clean\endcsname{.915}
\expandafter\def\csname tt@YelpChi@RLR@clean\endcsname{.798}

%% file: tables/numbers2.tex
\newcommand{\BRCleanLoose}{17}
\newcommand{\BRCleanN}{20}
\newcommand{\BRCarsTight}{15}
\newcommand{\BRCarsInformative}{4}

\newcommand{\BRCarsN}{20}
\newcommand{\EGmaxDrop}{.299}
\newcommand{\EGmaxWhere}{HGB-Graph on FDCompCN, P2, $B{=}20$}
\newcommand{\EGmedDropBase}{.108}
\newcommand{\RealMaxDev}{.051}

\newcommand{\RealMeanDev}{$-$.001}
\newcommand{\CIn}{45}

\newcommand{\CIabove}{6}

\newcommand{\CIidentical}{32}
\newcommand{\NIn}{44}

\newcommand{\CIsdMax}{.009}

\newcommand{\SelFree}{82}

\newcommand{\SelN}{225}

\newcommand{\VSn}{50}
\newcommand{\VScarsdGeSub}{50}

\newcommand{\CMcells}{160}
\newcommand{\CMtrap}{132}
\newcommand{\CMcarsdOk}{152}
\newcommand{\CMcarsdGain}{26}

\newcommand{\RGegBetter}{69}
\newcommand{\RGexBetter}{50}

\newcommand{\GKdevMax}{.009}

\newcommand{\FRtwo}{0.83}

\newcommand{\FgapIsoHi}{$-$.253}
\newcommand{\FgapIsoLo}{$-$.074}

\newcommand{\ABgateGain}{$+$.000}

\newcommand{\VLminFrac}{5}
\newcommand{\VLbelowLow}{9}
\newcommand{\VLworstLow}{$-$.099}

\newcommand{\GTnHi}{36}
\newcommand{\GTungDrop}{36}
\newcommand{\GTgain}{36}

\newcommand{\GGwin}{3}

%% file: tables/numbers3.tex
\newcommand{\DCAugCd}{$-$.001}

\newcommand{\DCAugNoValGate}{$-$.203}

\newcommand{\DCRichCd}{$+$.112}

\newcommand{\DCRichBlind}{$+$.076}

\newcommand{\DCRichFA}{$+$.065}

\newcommand{\DCRichRandThree}{$+$.037}

\newcommand{\DCRichRandEight}{$+$.067}

\newcommand{\DCRichPool}{$+$.121}

\newcommand{\DERichCostInfoWin}{20}
\newcommand{\DERichCostInfoLoss}{7}
\newcommand{\DERichCostInfoN}{52}

\newcommand{\DRCdRfour}{$+$.050}

\newcommand{\DRRandEightRfour}{$+$.047}

\newcommand{\DRCdRtwelve}{$+$.172}
\newcommand{\DRBlindRtwelve}{$+$.125}
\newcommand{\DRRandEightRtwelve}{$+$.087}

\newcommand{\DCpoolKmed}{13}
\newcommand{\DCpoolKmax}{24}

\newcommand{\HDn}{16}
\newcommand{\HDexpHit}{15}

\newcommand{\HDbestOther}{7}

\newcommand{\HDzeroShare}{85}
\newcommand{\HDsingleN}{146}
\newcommand{\HHn}{5}
\newcommand{\HHhit}{2}
\newcommand{\HHgainMean}{$+$.089}

\newcommand{\GRwinClosed}{14}

\newcommand{\GRlossMax}{$-$.022}
\newcommand{\GRwinRels}{2}
\newcommand{\GRrelsTotal}{17}
\newcommand{\GRmaxYelp}{$+$.021}
\newcommand{\GRisoYelp}{82}

\newcommand{\GRmaxSffsd}{$+$.061}
\newcommand{\GRisoSffsd}{26}

\newcommand{\CSn}{480}
\newcommand{\CSbelow}{46}
\newcommand{\CSbelowWorst}{$-$.023}
\newcommand{\CSgain}{90}
\newcommand{\CStrap}{95}
\newcommand{\CSnGain}{320}

\newcommand{\CSbelowFD}{34}

\newcommand{\VIknown}{$+$.003}
\newcommand{\VIknownGain}{2}
\newcommand{\VIcells}{10}

\newcommand{\VINaiveZeroBelow}{11}
\newcommand{\VINaiveZeroWorst}{$-$.673}

\newcommand{\VInZero}{40}

\newcommand{\VIMaximinTwo}{.000}
\newcommand{\VIMaximinTwoBelow}{1}

\newcommand{\VIcovTwo}{90}
\newcommand{\VInTwo}{40}

\newcommand{\VIMaximinThreeBelow}{0}

\newcommand{\VIcovThree}{100}

%% file: exp.tex
\paragraph{Setup.} We use the base multi-relational CSBM of \cref{sec:theory} ($N{=}6000$; an identity relation with $(p,d)=(0.8,4)$, a behavior relation $(0.55,10)$, and a dense relation $(0.25,30)$) and five real graphs: YelpChi and Amazon reviews \citep{rayana2015yelp,mcauley2013amazon,dou2020caregnn}; FDCompCN, financial-statement fraud among listed companies linked by investment, supplier, and customer ties \citep{wu2023splitgnn}; S-FFSD, simulated card transactions linked by shared card, merchant, location, and type \citep{xiang2023gtan}; and Tolokers, banned crowd workers linked by common tasks \citep{platonov2023critical}. Profile P1 makes the identity-like relation (R-U-R, U-V-U, C-I-C, T-S-T) cheap to manipulate and P2 makes it expensive (\cref{tab:profiles}); For each test fraud node, the \emph{evaluated worst case} is the minimum score over a query search, a cost-aware projected-gradient attack \citep{geisler2021prbcd}, transfers, the exact attack for linear models (\cref{prop:exact}), and a new \emph{exhaustive-greedy} attack for R-GCN and tree models, which scores every candidate edit exactly at each step, including moves that open or close degree gates (\cref{apd:setup}). Edits of all targets are applied jointly and every node is rescored; certified AUCs follow \cref{prop:cert}. We compare graph-free baselines (MLP, gradient-boosted trees HGB-X, logistic regression LR), clean-trained graph models (RLR, R-GCN, HGB-Graph \citep{tang2023gadbench}, CARE-GNN \citep{dou2020caregnn}, GNNGuard- and BWGNN-style models \citep{zhang2020gnnguard,tang2022bwgnn}), models adversarially trained at the evaluated costs (CA-AT, AT-RLR, CA-HGB), and \cars{}, \cars{}-D, and \cars{}+AT, over five seeds with stratified 40/20/40 splits. The confirmatory analysis is fixed in advance: 45 comparisons of \cars{}-D with its graph-free counterpart (three families, five graphs, three cells), hierarchical bootstrap over seeds and test nodes, and Holm correction; everything else is exploratory. Costs are scenarios, not estimates (\cref{apd:threat}).

\begin{figure*}[t]
\centering
\includegraphics[width=\textwidth]{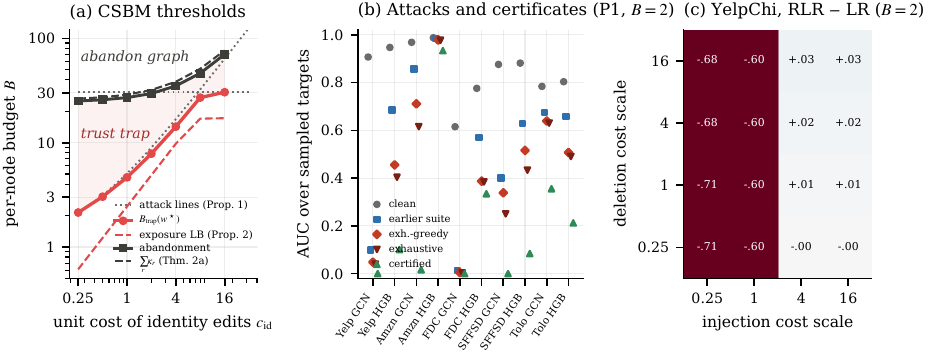}
\caption{(a) CSBM: the numerical trap threshold of the clean-optimal detector, the lower envelope of the trap-attack lines (\cref{prop:structure}), the exposure lower bound (\cref{prop:exposure}), and the abandonment threshold with its bound $\sum_r\kappa_r$ (\cref{thm:abandon}). (b) AUC over 150 sampled test fraud nodes per graph (P1, $B{=}2$, seed 0) for clean-trained R-GCN and HGB-Graph: clean, under the earlier attack suite, under the exhaustive-greedy attack, under exhaustive search within a pool of candidates, and the certified lower bound; the true worst case lies between the last two. (c) YelpChi: exact worst-case AUC of RLR minus LR at $B{=}2$ when injection and deletion costs are scaled relative to P1.}
\label{fig:main}
\end{figure*}

\begin{table*}[ht]
\centering
\scriptsize
\setlength{\tabcolsep}{2.5pt} 

\caption{Clean and evaluated worst-case test AUC (mean over five seeds; on YelpChi, Amazon, and S-FFSD, CARE-GNN, GNNGuard, BWGNN, CA-AT, AT-RLR, and \cars{}+AT (R-GCN) use three). Underlined: trust trap ($\Delta_{\mathrm{clean}}>0.002$ and $\Delta_{\mathrm{rob}}<-0.002$ against the graph-free counterpart: LR for linear models, MLP for neural models, HGB-X for trees). Bold: ours at or above the counterpart (within 0.002). Adversarially trained models and ours are trained or planned at the evaluated cell; their clean column is that of the P1, $B{=}2$ plan. Linear models (RLR, AT-RLR, RLR + \cars{}-D) report exact target-side worst cases; the other graph models report evaluated worst cases, upper bounds on the true worst case, whose certified brackets are labeled in \cref{tab:bracket}. All cells are in \cref{tab:mainfull}.}
\label{tab:main}

\begin{tabular}{l *{15}{r}}
\toprule
& \multicolumn{3}{c}{YelpChi} 
& \multicolumn{3}{c}{Amazon} 
& \multicolumn{3}{c}{FDCompCN} 
& \multicolumn{3}{c}{S-FFSD} 
& \multicolumn{3}{c}{Tolokers} \\
\cmidrule(lr){2-4} \cmidrule(lr){5-7} \cmidrule(lr){8-10} \cmidrule(lr){11-13} \cmidrule(lr){14-16}
Model & clean & P1,2 & P2,20 & clean & P1,2 & P2,20 & clean & P1,2 & P2,20 & clean & P1,2 & P2,20 & clean & P1,2 & P2,20 \\
\midrule
\input{tables/tab_main.tex}
\bottomrule
\end{tabular}
\end{table*}

\paragraph{The theory predicts where the trap opens.} In the CSBM, the numerical trap threshold coincides with the lower envelope of the trap-attack lines to within \TBTtwoLineGap{} (\cref{fig:main}a), and the exposure bound recovers between \TLBratioMin{} and \TLBratioMax{} of it. On real graphs, the exact cost maps of linear models show the same weakest-link structure (\cref{fig:main}c, \cref{fig:costmap}): the clean-trained RLR falls below LR in \CMtrap{} of \CMcells{} (dataset, cost scale, budget) cells, and the trap closes only when every cheap action becomes expensive. On the CSBM, a factorial study over the identity relation's signal, its isolation rate, its cost, and the architecture explains \FRtwo{} of the variance of the trap depth with cost as the dominant factor; isolating 85\% of the nodes on the relation deepens the trap from \FgapIsoLo{} to \FgapIsoHi{} on average (\cref{tab:factorial}), which is the mechanism behind FDCompCN.

\paragraph{The trust trap is the rule.} Clean-trained graph models fall below their graph-free counterparts in \NTrapBase{} of \NCellsBase{} (dataset, model, cell) combinations and lose up to \MaxLoss{} AUC (\cref{tab:main}). FDCompCN is a negative control: no relation carries clean signal, so no model is trapped, yet clean-trained graph models collapse, because 70--91\% of fraudulent firms have no neighbors in a given relation and one injection replaces their empty aggregate (\cref{prop:gnn}). Adversarial training at the evaluated costs does not remove the trap: at P1, $B{=}2$, CA-AT stays below the MLP on every graph, and CA-HGB, the strongest baseline under the earlier attacks, falls below HGB-X on YelpChi, S-FFSD, and Tolokers once attacks are exhaustive.

\paragraph{How close are the attacks to the worst case?} The exhaustive-greedy attack lowers the evaluated worst case of clean-trained nonlinear models by a median of \EGmedDropBase{} AUC over the earlier suite, and by up to \EGmaxDrop{} (\EGmaxWhere{}; \cref{tab:bracket}): it finds a lower score than the suite for \RGegBetter\% of sampled targets (\cref{tab:regret}). Exhaustive search within a pool of candidates still improves on it for \RGexBetter\% of targets on average, and the certified lower bound of \cref{prop:cert} is far below for clean-trained models (\cref{fig:main}b): their reported worst cases are optimistic, and the traps understated. We therefore label every nonlinear bracket: \BRCleanLoose{} of the \BRCleanN{} brackets of clean-trained nonlinear models are loose (width above 0.05), so for them we claim only an upper bound, whereas \BRCarsTight{} of the \BRCarsN{} brackets of \cars{}-D detectors are tight (width at most 0.01) and \BRCarsInformative{} more informative (\cref{tab:bracket}). A single realized joint attack differs from the per-node minimum by \RealMeanDev{} AUC on average and by at most \RealMaxDev{}, so the reported worst cases are achievable up to small cross effects.

\paragraph{Which relation to harden.} A platform can raise the cost of manipulating a relation; which one should it harden? Analyses with equal edit costs cannot pose this question. By \cref{cor:robust}(c), hardening a relation that some cheapest trap policy avoids changes nothing, and to first order the cheapest policy spends on the action of largest exposure per unit cost (\cref{prop:exposure,prop:gnn}). Making one relation at a time unmanipulable and computing the exact threshold of the clean-trained RLR on every graph where it has a clean gain and is trapped (\HDn{} cases: the CSBM, YelpChi, and CSBMs with 4 to 12 relations), \HDzeroShare\% of single-relation hardenings leave the threshold unchanged, and the relation of largest first-order exposure per unit cost, computed from one edit per node without any attack, is the best one to harden in \HDexpHit{} of \HDn{} cases; the most informative relation, the cheapest one, the largest weight, and an exact analysis under equal costs are right in at most \HDbestOther{}; for HGB-Graph, the same pick is best in only \HHhit{} of \HHn{} graphs (\cref{tab:hardening}).

\begin{table}[ht]
\centering\scriptsize
\caption{Selection rules at equal evaluation budget ($K$ validated candidates, median): mean test worst-case AUC difference with the graph-free detector on real graphs, real graphs with six attribute-sharing relations added (Aug.), and CSBMs with 4--12 relations (Rich), the last also with 10\% of the validation labels. Counts and regrets in \cref{tab:decomp}.}
\label{tab:decompmain}
\begin{tabular}{lc *{4}{r}}
\toprule
Rule & $K$ & Real & Aug. & Rich & Rich, 10\% \\
\midrule
\input{tables/tab_decomp_main.tex}
\bottomrule
\end{tabular}
\end{table}

\paragraph{What cost-aware selection adds.} \cars{}-D is at or above its counterpart in \OkDall{} of \CellsDall{} combinations and never more than \GapDall{} below; in the confirmatory analysis, no Holm-adjusted comparison is significantly negative, \CIabove{} are significantly positive, and non-inferiority at a margin of 0.01 AUC holds in \NIn{} of \CIn{} (\cref{tab:ci,tab:cifull}; seed standard deviations up to \CIsdMax{}). On these graphs, which have one to four relations, its value is abstention: it returns the graph-free detector in \SelFree\% of \SelN{} selections, \CIidentical{} comparisons are exact ties, and validation alone would do as well: validating every relation subset matches it in \VScarsdGeSub{} of \VSn{} cases with about twice the attack evaluations, and the gated candidate adds \ABgateGain{} (\cref{tab:valsel,tab:ablation2}). Its gains, where a relation is costly to manipulate, are certified on YelpChi at P2 (\cref{tab:bracket}). Planning matters when relations are many (\cref{tab:decompmain}). On CSBMs with 4 to 12 relations, \cars{}-D gains \DCRichCd{} over the graph-free detector with at most three validated candidates, whereas the same planner without relation-specific costs gains \DCRichBlind{}, and \{graph-free, all relations\} and random searches with three or eight candidates gain \DCRichFA{}, \DCRichRandThree{}, and \DCRichRandEight{}. The advantage grows with the number of relations: with 4 relations all rules are close (\cars{}-D \DRCdRfour{}, random search \DRRandEightRfour{}), with 12 it is \DRCdRtwelve{} against \DRBlindRtwelve{} without costs and \DRRandEightRtwelve{} for random search. Relation-specific costs are what the plug-in contributes: with identical candidates and validation, cost-blind planning is worse in \DERichCostInfoWin{} and better in \DERichCostInfoLoss{} of \DERichCostInfoN{} cases. Validating every configuration evaluated for any rule (VAL-POOL) does better still, \DCRichPool{}, but it validates \DCpoolKmed{} candidates per cell (up to \DCpoolKmax{}), and exhaustive subset search needs $2^R$. On real graphs augmented with six attribute-sharing relations, no rule gains (\cars{}-D \DCAugCd{}), because the added relations carry little signal beyond the features, and the plug-in's gated choice used without validation loses \DCAugNoValGate{} on average: validation is what makes the protocol safe. With \VLminFrac\% of the validation labels, \cars{}-D falls below the graph-free detector in \VLbelowLow\% of draws, by up to \VLworstLow{} (\cref{tab:labels}).

\paragraph{Decisions under cost uncertainty.} With exact attacks over 16 cost scales, \cars{}-D planned at the true costs is at or above LR in \CMcarsdOk{} of \CMcells{} cells and above it in \CMcarsdGain{} (\cref{fig:costmap}). Costs are rarely known: let the defender have an estimate off by a log-normal factor ($\sigma{=}1$) and a box of log half-width $w$ around it (\cref{tab:voi}). Planning at the estimate falls below LR in \VINaiveZeroBelow{} of \VInZero{} draws, by up to \VINaiveZeroWorst{}, because an error that makes a cheap relation look expensive keeps that relation. The maximin plan, at the box's lower corner (\cref{cor:robust}(b)), falls below LR in \VIMaximinTwoBelow{} of \VInTwo{} draws at $w{=}2$ and \VIMaximinThreeBelow{} at $w{=}3$, where the box covers the truth in \VIcovTwo\% and \VIcovThree\% of draws, but its mean gain is \VIMaximinTwo{}, against \VIknown{} with known costs (gains in \VIknownGain{} of \VIcells{} cells): safety is bought by giving up the graph, and the gain a defender with exact costs keeps is the value of cost information; minimax regret behaves the same once the box covers the truth. A defender who knows only a lower bound on costs never falls below LR but never gains either (\cref{tab:costagnostic}). In \CSn{} cases with random cost vectors from an independent and a behavior-structured distribution, the trap persists in \CStrap\% of cases with a clean gain, and \cars{}-D falls below LR in \CSbelow{}, by at most \CSbelowWorst{}, mostly on FDCompCN, whose relations carry no signal (\cref{tab:coststab}). Entity-level actions, gangs, and rolling temporal windows on S-FFSD give the same picture (\cref{tab:entity,tab:gang2,tab:temporal2}).

%% file: tables/tab_main.tex
MLP & .836 & .836 & .836 & .975 & .975 & .975 & .715 & .715 & .715 & .788 & .788 & .788 & .740 & .740 & .740\\
HGB-X & .935 & .935 & .935 & .977 & .977 & .977 & .789 & .789 & .789 & .845 & .845 & .845 & .734 & .734 & .734\\
LR & .771 & .771 & .771 & .969 & .969 & .969 & .680 & .680 & .680 & .644 & .644 & .644 & .722 & .722 & .722\\
\midrule
\multicolumn{16}{l}{\emph{Clean-trained, graph}}\\
RLR & .798 & \underline{.181} & \underline{.546} & .971 & \underline{.865} & \underline{.674} & .656 & .001 & .001 & .660 & \underline{.355} & \underline{.061} & .768 & \underline{.529} & \underline{.529}\\
R-GCN & .915 & \underline{.043} & \underline{.639} & .976 & .739 & .594 & .631 & .003 & .002 & .878 & \underline{.325} & \underline{.143} & .811 & \underline{.649} & \underline{.648}\\
HGB-Graph & .960 & \underline{.442} & \underline{.924} & .986 & \underline{.971} & \underline{.951} & .785 & .473 & .250 & .869 & \underline{.551} & \underline{.385} & .818 & \underline{.524} & \underline{.524}\\
CARE-GNN & .846 & \underline{.773} & \underline{.808} & .974 & .967 & .963 & .697 & .026 & .014 & .701 & .654 & .624 & .747 & \underline{.705} & \underline{.705}\\
GNNGuard-style & .917 & \underline{.169} & .850 & .976 & .912 & .880 & .652 & .014 & .011 & .868 & \underline{.567} & \underline{.529} & .808 & \underline{.637} & \underline{.636}\\
BWGNN-style & .902 & \underline{.593} & .868 & .978 & \underline{.967} & \underline{.946} & .688 & .093 & .082 & .867 & \underline{.715} & \underline{.654} & .831 & \underline{.729} & \underline{.726}\\
\midrule
\multicolumn{16}{l}{\emph{Adversarially trained}}\\
CA-AT (R-GCN) & .866 & \underline{.679} & \underline{.834} & .976 & .892 & .881 & .644 & .229 & .120 & .853 & \underline{.686} & \underline{.628} & .819 & \underline{.734} & \underline{.734}\\
AT-RLR (exact) & .773 & \underline{.366} & \underline{.597} & .967 & .935 & .901 & .619 & .014 & .012 & .645 & .562 & \underline{.529} & .737 & \underline{.698} & \underline{.697}\\
CA-HGB & .962 & \underline{.919} & .944 & .987 & .979 & \underline{.964} & .784 & .748 & .540 & .864 & \underline{.702} & \underline{.639} & .818 & \underline{.680} & \underline{.680}\\
\midrule
\multicolumn{16}{l}{\emph{Ours}}\\
RLR + \cars{} & .771 & \textbf{.771} & \textbf{.791} & .972 & \underline{.951} & \textbf{.969} & .680 & .674 & \textbf{.680} & .644 & .504 & \textbf{.664} & .722 & \textbf{.722} & \textbf{.722}\\
R-GCN + \cars{} & .836 & \textbf{.836} & \textbf{.897} & .976 & .943 & \textbf{.975} & .671 & .669 & \textbf{.715} & .843 & \underline{.328} & \textbf{.858} & .740 & \textbf{.740} & \textbf{.740}\\
HGB + \cars{} & .935 & \textbf{.935} & \textbf{.952} & .982 & \textbf{.976} & \textbf{.977} & .788 & .787 & \textbf{.789} & .852 & \underline{.449} & \textbf{.863} & .734 & \textbf{.734} & \textbf{.734}\\
RLR + \cars{}-D & .771 & \textbf{.771} & \textbf{.792} & .969 & \textbf{.969} & \textbf{.969} & .682 & \textbf{.678} & \textbf{.680} & .644 & \textbf{.644} & \textbf{.664} & .722 & \textbf{.722} & \textbf{.722}\\
R-GCN + \cars{}-D & .836 & \textbf{.836} & \textbf{.897} & .975 & \textbf{.975} & \textbf{.975} & .715 & \textbf{.715} & \textbf{.715} & .788 & \textbf{.788} & \textbf{.858} & .746 & \textbf{.741} & \textbf{.741}\\
HGB + \cars{}-D & .935 & \textbf{.935} & \textbf{.952} & .979 & \textbf{.975} & \textbf{.977} & .790 & \textbf{.789} & \textbf{.789} & .845 & \textbf{.845} & \textbf{.863} & .734 & \textbf{.734} & \textbf{.734}\\
\cars{}+AT (R-GCN) & .837 & \textbf{.837} & \textbf{.896} & .973 & \textbf{.973} & \textbf{.973} & .715 & \textbf{.715} & \textbf{.715} & .782 & .782 & \textbf{.856} & .746 & \textbf{.741} & \textbf{.741}\\
\cars{}+AT (HGB) & .935 & \textbf{.935} & \textbf{.952} & .979 & \textbf{.976} & \textbf{.977} & .790 & \textbf{.789} & \textbf{.789} & .845 & \textbf{.845} & \textbf{.863} & .734 & \textbf{.734} & \textbf{.734}\\

%% file: tables/tab_decomp_main.tex
\cars{}-D & 2 & .000 & $-$.001 & $+$.112 & $+$.110\\
cost-blind planning & 2 & .000 & $-$.003 & $+$.076 & $+$.074\\
no validation (subset) & 1 & $-$.005 & $-$.078 & $+$.101 & $+$.101\\
no validation (gates) & 1 & $-$.044 & $-$.203 & $+$.076 & $+$.076\\
no gated candidate & 2 & .000 & $-$.001 & $+$.112 & $+$.110\\
no graph-free candidate & 2 & $-$.002 & $-$.078 & $+$.103 & $+$.102\\
VAL-FA & 2 & .000 & .000 & $+$.065 & $+$.063\\
VAL-RAND-3 & 3 & $-$.001 & .000 & $+$.037 & $+$.033\\
VAL-RAND-8 & 8 & $-$.002 & .000 & $+$.067 & $+$.061\\
VAL-SUB & 16 & .000 & -- & $+$.060 & $+$.054\\
VAL-POOL & 13 & $-$.002 & $-$.004 & $+$.121 & $+$.116\\

%% file: related.tex
\textbf{Graph fraud detection.} Multi-relational detectors weigh relations through attention or relation-specific weights \citep{schlichtkrull2018rgcn,wang2019han}, select neighbors against camouflage \citep{dou2020caregnn,liu2021pcgnn,zhang2024digin}, or filter spectra under heterophily \citep{tang2022bwgnn,gao2023ghrn,wu2023splitgnn}; benchmarks show that tree ensembles on neighborhood aggregates are strong baselines \citep{tang2023gadbench}. All of them choose relations for their clean signal; we ask how that choice should change once relations have prices.
\textbf{Structural attacks and defenses.} Gradient and combinatorial attacks perturb edges under uniform budgets \citep{zugner2018nettack,geisler2021prbcd}, low-degree nodes are known to be easier targets \citep{zugner2018nettack}, and many defenses are broken by adaptive attacks \citep{mujkanovic2022defenses,tramer2020adaptive}; purification \citep{zhang2020gnnguard,zhang2022rohe} reweights or prunes neighbors, and certificates bound the effect of a number of flips \citep{bojchevski2019certifiable,zugner2019certifiable}; our certificates are instead per-node and cost-aware (interval propagation \citep{gowal2018ibp}, tree bounds \citep{chen2019treeverif}). Fraud gangs coordinate node injections against several targets \citep{choi2025monti}. Cost-aware attacks and defenses price nodes in homogeneous graphs \citep{liao2024riskeeper}, and adversaries with action costs go back to adversarial classification \citep{dalvi2004adversarial,lowd2005adversarial}, with cost-sensitive robustness studied for images \citep{zhang2019costsensitive} and tabular data \citep{kireev2023tabular}. We price relations and actions separately.
\textbf{Strategic classification and robust optimization.} Strategic classification lets agents alter features at a cost \citep{hardt2016strategic,eilat2023strategic}, and adversarial training solves a min-max problem \citep{madry2018pgd}. \Cref{prop:structure} and \cref{thm:abandon} are statements about this min-max problem in a specific structured form: they characterize when an entire input channel, a relation, should be discounted or abandoned, and how the answer depends on the cost vector.
\textbf{CSBM analyses.} Contextual stochastic block models explain when graph convolution helps \citep{deshpande2018csbm,baranwal2021graph,ma2022homophily}; we extend them to several relations, priced manipulations, and a robust rather than a clean criterion.

%% file: conclusion.tex
Taking manipulation prices as given, degree gates strictly beat selection and reweighting of a relation on which part of the population is isolated, within a sharp window that real graphs confirm, and exposure per unit cost says which relation to harden; linear models admit exact worst cases, and nonlinear ones certified brackets. Empirically, the trap is the rule, and validated cost-aware selection is never significantly worse than ignoring the graph, mostly abstains with few relations, and earns its planning step with many.

\textbf{Limitations.} Our guarantees are of four kinds (\cref{tab:claims}): exact results are unconditional within their model; certified bounds assume the target-side model and are loose for clean-trained nonlinear detectors, whose reported worst cases are therefore upper bounds; and the validation guarantee is relative to the evaluated attacks and assumes i.i.d.\ validation nodes. The study is conditional on cost scenarios: no cost was estimated from operational data, errors that make a cheap relation look expensive are harmful unless planning covers them, and without cost information the safe choice is to abstain (\cref{tab:voi}). \cars{}-D needs labeled validation fraud (\cref{tab:labels}); the evidence that planning helps with many relations comes from synthetic graphs, and augmenting real graphs with attribute relations showed no gain (\cref{tab:decomp}); S-FFSD, simulated, is the only dated graph; the phase diagram and the sensitivity study use the earlier attack suite (\cref{tab:rerun}). Eliciting real costs is the next step.

%% file: appendix_proofs.tex
\section{Proofs}
\label{apd:proofs}

\paragraph{CSBM moments.} Let $n^f_r=p_rd_r-k_r$, $n^b_r=(1-p_r)d_r+b_r$, and $n_r=n^f_r+n^b_r$. For a fraudulent node under attack $(b,k)$,
\[
\mu_{1,r}=\frac{m\,n^f_r-m(1-p_r)d_r-(m+\delta)\,b_r}{n_r},\qquad
\Sigma_{1,rr}=\frac{1}{n_r}+\frac{4m^2d_rp_r(1-p_r)}{n_r^2},
\]
where the second term accounts for the binomial composition of the neighborhood. For a benign node, $\mu_{0,r}=(2q_r-1)m$ and $\Sigma_{0,rr}=(1+4m^2q_r(1-q_r))/d^0_r$. For the own feature, $\mu_{y,0}=\pm m$ and $\Sigma_{y,00}=1$. Relations are independent, so the covariances are diagonal, $z_v$ is Gaussian given the class, and $\AUC=\Pr(z_{\text{fraud}}>z_{\text{benign}})=\Phi(\zeta)$ with $\zeta$ as in \eqref{eq:auc}. We write $\sigma_r^2=\Sigma_{1,rr}(0)+\Sigma_{0,rr}$.

\paragraph{Proof of \cref{prop:structure}.} (a) Under per-node budgets, $f$ is trapped at $(B,c)$ if and only if some affordable policy is a trap policy, that is, if and only if $B\ge\max_vc^\top\theta_v(\pi)$ for some $\pi\in\mathcal T(f)$. The set of budgets at which $f$ is trapped is therefore $\{B\ge\inf_{\pi\in\mathcal T(f)}\max_vc^\top\theta_v(\pi)\}$. The infimum is attained: edit counts are nonnegative integers and $c>0$, so only finitely many edit vectors cost at most any given level, and the set of values $\max_vc^\top\theta_v(\pi)$ below any level is finite. The gang case is identical with the mean in place of the maximum, and $\frac1{|T|}\sum_v\le\max_v$ gives $\bar B_{\mathrm{trap}}\le\Btrap$. (b) For fixed $\pi$, $c\mapsto\max_vc^\top\theta_v(\pi)$ and $c\mapsto\frac1{|T|}\sum_vc^\top\theta_v(\pi)$ are nondecreasing and positively homogeneous because $\theta_v\ge0$; both properties survive a pointwise minimum. (c) For fixed $\pi$, $c\mapsto\frac1{|T|}\sum_vc^\top\theta_v(\pi)=c^\top\bar\theta(\pi)$ is linear, and a pointwise minimum of linear functions is concave; it is piecewise linear because, on any bounded set of costs, only finitely many mean edit vectors can attain the minimum. If performance depends on $\pi$ only through a common edit vector $\theta$ ($\theta_v=\theta$ for all $v$), the per-node cost is $c^\top\theta$, linear, and the same argument applies. Without that property, $\max_v$ is convex in $c$; the two-node example in the text, where node 1 can be trapped only by one unit of action 1 and node 2 only by one unit of action 2, gives $\Btrap=\max(c_1,c_2)$, which is convex and not concave. No property of $f$, of the data, or of the performance measure was used. The vertex statement of \cref{cor:robust} follows because a concave function attains its minimum over a polytope at a vertex. \hfill$\square$

\paragraph{Proof of \cref{prop:cert}.} Under the target-side model, the attacked score of each fraudulent test node is at least $\ell_v$ and benign scores do not move, so every fraud--benign pair ordered correctly by $(\ell_v,s_w)$ is ordered correctly under every feasible attack; hence $A_{\mathrm{true}}\ge A_{\mathrm{cert}}$. Any evaluated attack is feasible, so $A_{\mathrm{true}}\le A_{\mathrm{eval}}$. The margin statement takes $\ell_v=s_v-\Delta_v$. For the interval, write $\bar x'_{v,r}=(S'-\sum_{u\in D}x_u+\sum_{u\in I}x_u)/(d-k+b)$ for a deletion set $D$ of $k$ accomplices and an injection set $I$ of $b$ candidates. Coordinate $j$ of $W_r\bar x'_{v,r}$ is $(S_j-\sum_{u\in D}(W_rx_u)_j+\sum_{u\in I}(W_rx_u)_j)/(d-k+b)$, which is smallest when $D$ holds the $k$ accomplices with the largest $(W_rx_u)_j$ and $I$ the $b$ candidates with the smallest, and largest in the opposite case; the bounds are attained coordinatewise, though not jointly. If $d-k+b=0$, the relation is empty and the aggregate is exactly $0$, which the certificate includes as a separate case. Interval arithmetic through ReLU (monotone) and linear maps gives valid bounds for deeper layers; for neighbors and candidates whose own aggregation changes because the target joins or leaves it, we bound their states over every subset of relations in which this happens. For a tree ensemble, every leaf reached by some input in the box of reachable features is a candidate, and the sum over trees of the smallest reachable leaf value lower-bounds the score. \hfill$\square$

\paragraph{Proof of \cref{prop:val}.} Fix a candidate. Its evaluated worst-case validation AUC $\hat A$ is a two-sample U-statistic of i.i.d.\ attacked scores, so $\mathbb E\hat A=A$, the population value. Replacing one fraudulent node changes $\hat A$ by at most $1/n_+$ and one benign node by at most $1/n_-$; by McDiarmid's inequality, $\Pr(|\hat A-A|\ge\varepsilon)\le2\exp(-2\varepsilon^2/(1/n_++1/n_-))$. A union bound over the $K$ candidates gives $|\hat A_k-A_k|<\varepsilon$ for all $k$ with probability at least $1-\delta$ at the stated $\varepsilon$. On that event, the picked candidate $\hat k$ satisfies $A_{\hat k}>\hat A_{\hat k}-\varepsilon\ge\hat A_k-\varepsilon>A_k-2\varepsilon$ for every $k$. \hfill$\square$

\paragraph{Gangs with cross effects.} Let $A_{\mathrm{sep}}$ be the AUC computed from separable attacked fraud scores and clean benign scores, and $A_{\mathrm{real}}$ the AUC after all edits. The two score vectors differ only on $E_+\cup E_-$. A fraudulent node in $E_+$ enters $n_-$ pair indicators and a benign node in $E_-$ enters $n_+$, so $|A_{\mathrm{real}}-A_{\mathrm{sep}}|\le(|E_+|n_-+|E_-|n_+)/(n_+n_-)=|E_+|/n_++|E_-|/n_-$. \hfill$\square$

\paragraph{Proof of \cref{prop:exposure}.} The set $\Theta_B(c)$ is convex and contains $0$, so for $\theta\in\Theta_B(c)$ the segment $t\theta$, $t\in[0,1]$, stays in it. Since $\zeta$ is continuously differentiable on the feasible set (all neighborhood sizes stay positive or the relation empties, where the moments extend continuously),
\[
\zeta(\theta)-\zeta(0)=\int_0^1\nabla\zeta(t\theta)^\top\theta\,dt\ \ge\ -\sum_jL_j(B)\theta_j\ \ge\ -\max_j\frac{L_j(B)}{c_j}\,c^\top\theta\ \ge\ -B\max_j\frac{L_j(B)}{c_j}.
\]
If the right-hand side exceeds $\zeta_0-\zeta(0)$, every feasible attack leaves $\zeta>\zeta_0$, so $\Btrap\ge B$. For the first-order exposure, $\partial\mu_{1,r}/\partial b_r|_0=-(2p_rm+\delta)/d_r$ and $\partial\Sigma_{1,rr}/\partial b_r|_0=-(1+8m^2p_r(1-p_r))/d_r^2$; the variance term enters the derivative of $\zeta$ with the factor $\zeta w_r^2/(2s^2)$, hence the remainder $O(w_r^2d_r^{-2})$. Over small budgets the caps are inactive and the feasible set is the simplex with vertices $0$ and $(B/c_j)e_j$, so the attacker's first-order optimum is a vertex: the whole budget goes to the action of largest exposure. \hfill$\square$

\paragraph{Proof of \cref{prop:gnn}.} Let $d=d_r(v)$, $\bar x=\bar x_{v,r}$ (with $\bar x=0$ if $d=0$), and $S=\sum_{u\in\mathcal N_r(v)}x_u=d\bar x$. After injecting the endpoints $I$ ($|I|=b$) and deleting the accomplices $D$ ($|D|=k$), the relation has $n=d-k+b$ neighbors. \emph{Case $n\ge1$.} The new mean is $\bar x'=(S+\sum_{u\in I}x_u-\sum_{u\in D}x_u)/n$, and since $S=d\bar x$ and $n=d-k+b$,
\[
\bar x'-\bar x=\frac{\sum_{u\in I}(x_u-\bar x)-\sum_{u\in D}(x_u-\bar x)}{n},\qquad \|\bar x'-\bar x\|\le\frac{(b+k)D_{v,r}}{n}.
\]
This identity also holds for $d=0$, where $\bar x=0$ and $D_{v,r}$ bounds the norms of the injected endpoints. \emph{Case $n=0$.} The attack deletes every neighbor ($k=d$, $b=0$), the aggregate becomes $0$ by convention, and $\|\bar x'-\bar x\|=\|\bar x\|$; this change is not controlled by deviations from the mean (two neighbors equal to $\bar x$ give $D_{v,r}=0$), which is why the statement bounds it separately. In both cases the pre-activation changes by $\sum_rW_r(\bar x'_{v,r}-\bar x_{v,r})$, whose norm is at most $\sum_r\|W_r\|_2\Delta_{v,r}$, and $s_v$ is $\Lambda$-Lipschitz in it. For one unit of budget: one injection gives $n=d+1\ge1$ and a change of at most $D_{v,r}/(d+1)$ at cost $\cin_r$; one deletion gives $n=d-1$, a change of at most $D_{v,r}/(d-1)$ if $d\ge2$ and exactly $\|\bar x\|$ if $d=1$, at cost $\cdel_r$. The node-level bounds of \cref{tab:exposure} use these two cases. \hfill$\square$

\paragraph{Proof of \cref{thm:abandon}.} (a) For relation $r$, $\Delta_r(b_r,k_r)\le0$ is equivalent to $m(p_rd_r-k_r)-m(1-p_r)d_r-(m+\delta)b_r\le\mu_{0,r}(d_r-k_r+b_r)$, because the denominator is positive; this is a half-space in $(b_r,k_r)$, so its cheapest point under $\cin_rb_r+\cdel_rk_r$ is a vertex, $(\bar b_r,0)$ or $(0,\bar k_r)$, and $\bar k_r\le p_rd_r$ respects the cap. Hence $\kappa_r$ is the exact cost of neutralizing $r$. Let $w$ be any detector and suppose $B\ge\sum_r\kappa_r$. The attacker neutralizes every relation with $w_r>0$ and leaves the others untouched. Then $w_r\Delta_r(\theta)\le0$ for $w_r>0$, and $w_r\Delta_r(0)\le0$ for $w_r\le0$ because $\Delta_r(0)\ge0$ when $p_r\ge q_r$. The numerator of $\zeta$ is at most $w_0\Delta_0=2mw_0$. If $w_0\le0$, $\zeta\le0<\zeta_0$. If $w_0>0$, the denominator is at least $w_0\sqrt{\Sigma_{1,00}+\Sigma_{0,00}}=\sqrt2w_0$, so $\zeta\le\sqrt2m=\zeta_0$. Thus no detector is above the graph-free one, and $\Bab\le\sum_r\kappa_r$.

(b) Under linear response, action $j$ on relation $r(j)$ lowers $\Delta_{r(j)}$ by $\Delta_{r(j)}/n_j$ per unit, $n_j\in\{\bar b_{r(j)},\bar k_{r(j)}\}$, and variances are fixed, so $\zeta(w;\theta)=(w^\top\Delta-\sum_j\theta_jw_{r(j)}\Delta_{r(j)}/n_j)/\|w\|$ with $\|w\|^2=\sum_{r\ge0}w_r^2\sigma_r^2$. The attacker minimizes a linear function over a simplex, so it spends its budget on one action, and the damage is $B\max_r(w_r)_+\Delta_r/\kappa_r$. For $w^\star_r=\Delta_r/\sigma_r^2$, $w^{\star\top}\Delta=\|w^\star\|^2=\zeta_\star^2$ and the damage is $B\max_r\zeta_r^2/\kappa_r$, which gives $\Btrap(w^\star)$. If $B\ge\sum_r\kappa_r$, part (a) applies. If $B<\sum_r\kappa_r$, the direction $v_r=\kappa_r/\Delta_r$ gives $\zeta(e_0+\varepsilon v)=\zeta_0+\varepsilon(\sum_r\kappa_r-B)/\sigma_0+O(\varepsilon^2)>\zeta_0$ for small $\varepsilon>0$, so $\Bab=\sum_r\kappa_r$. Finally, the robust objective has a concave, positively homogeneous numerator, so maximizing it over $\{\|w\|\le1\}$ is a convex program whose KKT conditions read $\Delta-B\sum_r\mu_r(\Delta_r/\kappa_r)e_r=2\nu\,\mathrm{diag}(\sigma^2)w$, with $\mu$ a probability vector on the relations that attain the maximum $t$. Off its support, $w_r=\lambda\Delta_r/\sigma_r^2$ with $\lambda=1/(2\nu)$, as for $w_0$; on its support, $w_r\Delta_r/\kappa_r=t$ and $w_r\le\lambda\Delta_r/\sigma_r^2$. \hfill$\square$

\paragraph{Coordinated adversaries.}
\begin{proposition}[Gangs convexify the attack curve]
\label{prop:gang}
Let $A(B)$ be the AUC under the per-node worst case at budget $B$, with benign scores fixed and edits acting on their own targets. A gang with average budget $\bar B$ can drive the AUC down to the lower convex envelope $\breve A(\bar B)$ of $A$ (up to rounding of $\phi|T|$). In particular, if $A(0)>A_{\mathrm{free}}>A(B_1)$, the gang traps the detector with average budget $\bar B\le f^\star B_1$, where $f^\star=(A(0)-A_{\mathrm{free}})/(A(0)-A(B_1))$.
\end{proposition}
A detector whose clean advantage is small relative to its per-node damage is trapped by a gang that spends a few units on a few members. When edits also move other nodes' scores (shared endpoints, two-hop aggregation), the realized AUC differs from the envelope by at most $|E_+|/n_++|E_-|/n_-$, where $E_\pm$ are the fraudulent and benign test nodes whose scores are affected by edits other than their own; we measure both.

\paragraph{Proof of \cref{prop:gang}.} Let $F_v(b)$ be the fraction of benign nodes scored above target $v$ when $v$ receives budget $b$ and plays its per-node worst case; with benign scores fixed and edits acting on their own targets, the AUC of an allocation $(b_v)$ is $\frac1{|T|}\sum_vF_v(b_v)$, and $A(b)=\frac1{|T|}\sum_vF_v(b)$. For budgets $b_1,b_2$ and a fraction $\phi\in[0,1]$ with $\phi b_1+(1-\phi)b_2=\bar B$, assign $b_1$ to a uniformly random subset of $\phi|T|$ targets and $b_2$ to the others. The expected AUC is $\phi A(b_1)+(1-\phi)A(b_2)$, and some deterministic subset does at least as well. Minimizing over $(\phi,b_1,b_2)$ gives the lower convex envelope of $A$ on the real line, since two points suffice in one dimension (Carath\'eodory). The second claim takes $b_2=0$, $b_1=B_1$, and $\phi=f^\star$. A gang that targets the most vulnerable members does strictly better whenever the $F_v$ differ. \hfill$\square$

\paragraph{Proof of \cref{prop:exact}.} The score of target $v$ depends on its edits only through the means $\bar x_{v,r}$, and $w_r^\top\bar x'_{v,r}=(\sum_{u\in\mathcal N_r(v)}\pi_r(u)+\sum_{u\in I}\pi_r(u)-\sum_{u\in D}\pi_r(u))/(d-k+b)$ with $\pi_r(u)=w_r^\top x_u$. For fixed counts $(b,k)$ the denominator is fixed, so the minimum takes the $b$ candidates with the smallest $\pi_r$ and the $k$ accomplices with the largest. Prefix sums give the minimum for every $(b,k)$ in $O(b^{\max}_rk^{\max}_r)$; a running minimum over cost units gives the best contribution of relation $r$ for every budget; and the relations are combined by a dynamic program over cost units in $O(R(B/\epsilon)^2)$. The edits of other targets do not enter $v$'s score, which is why the result is exact on the target side; the AUC is then evaluated with all edits applied jointly. \hfill$\square$

\paragraph{Degree gating bounds.}
\begin{proposition}[Degree gating]
\label{prop:gate}
In the setting of \cref{prop:gnn}, let relation $r$ enter $h_v$ only if the current degree satisfies $d_r(v)\ge\tau_r\ge1$. (a) If the gate is open before and after the attack, the attack changes $s_v$ through $r$ by at most $\Lambda\|W_r\|_2(b_r+k_r)D_{v,r}/\tau_r$. (b) Closing an open gate requires $d_r(v)-\tau_r+1$ deletions, hence as many accomplices and a cost of $\cdel_r(d_r(v)-\tau_r+1)$, and changes $s_v$ by at most $\Lambda\|W_r\|_2\|\bar x_{v,r}\|$. (c) Opening a closed gate requires $\tau_r-d_r(v)$ injections, at cost $\cin_r(\tau_r-d_r(v))$. Hence, under a per-node budget $B$, nodes with $d_r(v)>\tau_r-1+B/\cdel_r$ cannot be switched off, nodes with $d_r(v)<\tau_r-B/\cin_r$ cannot be switched on, and the per-unit exposure of every open node is at most $\Lambda\|W_r\|_2D_{v,r}/(\tau_r\min(\cin_r,\cdel_r))$.
\end{proposition}

\paragraph{Proof of \cref{prop:gate}.} (a) Let $n=d_r(v)-k_r+b_r$ be the degree after the attack, with $n\ge\tau_r$ because the gate stays open. The new mean is $\bar x'=(d_r(v)\bar x_{v,r}+\sum_{u\in I}x_u-\sum_{a\in K}x_a)/n$, so $\bar x'-\bar x_{v,r}=(\sum_{u\in I}(x_u-\bar x_{v,r})-\sum_{a\in K}(x_a-\bar x_{v,r}))/n$ and $\|\bar x'-\bar x_{v,r}\|\le(b_r+k_r)D_{v,r}/\tau_r$. The pre-activation moves by at most $\|W_r\|_2$ times this, and $s_v$ by at most $\Lambda$ times that. (b) After the attack the gate is closed, so $d_r(v)-k_r+b_r\le\tau_r-1$ and, since $b_r\ge0$, $k_r\ge d_r(v)-\tau_r+1$; only accomplices can be deleted, at unit cost $\cdel_r$. A closed gate replaces $\bar x_{v,r}$ by $0$, which moves the pre-activation by at most $\|W_r\|_2\|\bar x_{v,r}\|$. (c) Symmetrically, $d_r(v)+b_r-k_r\ge\tau_r$ forces $b_r\ge\tau_r-d_r(v)$. The last statement combines (a)--(c): with budget $B$, at most $B/\cdel_r$ deletions and $B/\cin_r$ injections are affordable, and $b_r+k_r\le B/\min(\cin_r,\cdel_r)$. \hfill$\square$

\paragraph{Proof of \cref{cor:robust}.} If $c'\ge c$, every $\theta\ge0$ with $c'^\top\theta\le B$ satisfies $c^\top\theta\le c'^\top\theta\le B$, and the deletion caps do not depend on costs, so $\Theta_B(c')\subseteq\Theta_B(c)$; a minimum over a smaller set is larger. Finally $c^\top\theta\le\rho B$ if and only if $(c/\rho)^\top\theta\le B$. (b) Every $c\in U$ satisfies $c\ge\underline c$, so $A_B(f;c)\ge A_B(f;\underline c)$ by (a), with equality at $c=\underline c\in U$; the maximin value of each detector is therefore its worst case at $\underline c$. (c) Let $\pi$ be a cheapest trap policy at $c$ with no action on $r$, and let $c'$ differ from $c$ only on the coordinates of $r$, with $c'\ge c$. The cost of $\pi$ is the same under $c'$, so $\Btrap(f;c')\le\Btrap(f;c)$ by \cref{prop:structure}(a), and $\Btrap(f;c')\ge\Btrap(f;c)$ by \cref{prop:structure}(b). \hfill$\square$

\paragraph{Proof of \cref{thm:gatedom}.} Under the Gaussian moments of the CSBM, each (class, degree) component has a Gaussian score, and the AUC is $\sum_{g,h}\pi^f_g\pi^b_h\Phi(\eta_{gh})$ over fraud components $g$ and benign components $h$, with $\eta_{gh}$ the standardized mean difference of the pair. Isolated nodes have an empty aggregate, set to $0$; injected candidates have projection $-(m+\delta)$.
(a) Fix $w>0$ and $b<\tau\le d$. Under the gated detector, an isolated fraud node cannot open the gate ($0+b<\tau$), and closing the gate of a connected fraud node is unaffordable, so the attacker's best response acts on connected fraud nodes only; call it $\pi^\star$. Apply $\pi^\star$ to the connected fraud nodes of the ungated detector, and inject $b$ candidates into every isolated fraud node. Connected nodes then have the same scores under both detectors (their gate is open before and after the attack, since $d\ge\tau$ and injections only raise degrees), benign isolated nodes have the same score $w_0x_v$, and isolated fraud nodes have score $w_0x_v-w(m+\delta)$ instead of $w_0x_v$. Every pair involving an isolated fraud node is therefore ordered correctly with strictly smaller probability and every other pair with the same probability, so the ungated AUC under this attack is strictly below the gated worst case, and the ungated worst case is no larger than this AUC.
(b) At $w=0$, every pair has $\eta_{gh}=2mw_0/(\sqrt2w_0)=\sqrt2m$, so $\partial_w\Phi(\eta_{gh})|_{w=0}=\phi(\sqrt2m)\,\mathbb E[\Delta a_{gh}]/(\sqrt2w_0)$ with $\Delta a_{gh}$ the difference of the relation terms of the pair. Under the attack, the relation term of a connected fraud node has mean $\mu'_d$, that of a connected benign node $\mu_b$, and isolated nodes contribute $0$ (gate closed). Summing over pairs with weights $(1-\pi_0)^2$, $(1-\pi_0)\pi_0$, $\pi_0(1-\pi_0)$, $\pi_0^2$ gives a derivative proportional to $(1-\pi_0)^2(\mu'_d-\mu_b)+(1-\pi_0)\pi_0\mu'_d-\pi_0(1-\pi_0)\mu_b=(1-\pi_0)(\mu'_d-\mu_b)$, which is positive iff $b<\bar b_d$. The attacker's best response may depend on $w$, but at $w=0$ every response gives the same AUC, and the worst case is a minimum of differentiable functions over a finite set of injection counts, so its right derivative at $0$ is the smallest of their derivatives, each equal to the expression above with the corresponding count $b'\le b$, hence positive. Therefore $A_{\mathrm{gated}}(w)>A_{\mathrm{gated}}(0)=A_{\mathrm{drop}}$ for small $w>0$. Combining (a) and (b), $\sup_wA_{\mathrm{gated}}\ge A_{\mathrm{gated}}(w_1)>A_{\mathrm{drop}}$ for some $w_1>0$, and $\sup_wA_{\mathrm{gated}}\ge A_{\mathrm{gated}}(w)>A_{\mathrm{ungated}}(w)$ for every $w>0$; the supremum over ungated $w\ge0$ is attained on a compact range of normalized weights, which gives the strict inequality.  (c) Let $b\ge\tau$. An isolated fraud node can open its gate with $\tau\le b$ injections of the most benign-looking candidates, and its aggregate is then $-(m+\delta)$, exactly the value it reaches in the ungated detector with any number $b'\ge1$ of injections; it can also leave its aggregate at $0$ in both detectors. Its set of reachable relation terms, $\{0,-(m+\delta)\}$, is therefore the same with and without the gate. A connected fraud node has degree $d\ge\tau$, injections only raise its degree, and closing is unaffordable, so its gate stays open and its reachable terms are the same as without the gate. Benign nodes are not attacked and have the same scores in both detectors (isolated ones contribute $0$ either way). Every node thus has the same set of attacked scores, the best responses coincide, and so do the worst-case AUCs. If $\tau>d$, no clean node passes the gate, so the clean scores equal those of the detector without the relation; an attack can only open gates, and with $w\ge0$ it does so only when this lowers a fraud score, so the gated worst case is at most $A_{\mathrm{drop}}$. \hfill$\square$

%% file: appendix_details.tex
\section{Experimental Details}
\label{apd:setup}

\paragraph{Datasets.} \textbf{YelpChi} \citep{rayana2015yelp,dou2020caregnn}: 45{,}954 reviews (14.5\% fraud), 32 features; R-U-R (same user), R-T-R (same product, same month), R-S-R (same product, same rating). Fraud-node mean degrees are 0.29, 25.1, and 142.6, and 82\% of fraudulent reviews have no R-U-R neighbor. \textbf{Amazon} \citep{mcauley2013amazon,dou2020caregnn}: 11{,}944 users (6.9\% fraud), 25 features; U-P-U, U-S-U, U-V-U; following the CARE-GNN release, the first 3{,}305 users stay in the graph but are not used for training or evaluation. Degrees are heavy-tailed: the median U-P-U degree of fraud nodes is 6 and the mean 49. \textbf{FDCompCN} \citep{wu2023splitgnn}: 5{,}317 Chinese listed companies (10.5\% financial-statement fraud), 57 features; C-I-C (investment), C-P-C (supplier), C-S-C (customer). Its relations are sparse (fraud-node degrees 0.79, 0.12, 0.17; 70--91\% of fraud nodes are isolated) and barely homophilous. \textbf{S-FFSD} \citep{xiang2023gtan}: 77{,}881 card transactions, of which 29{,}643 are labeled (17.7\% fraud); features are the log amount and the time stamp. We link each transaction to the three previous transactions with the same source (T-S-T), target merchant (T-M-T), location (T-L-T), and type (T-Y-T); unlabeled transactions stay in the graph but are neither trained on, evaluated, nor used as attack endpoints. The relations are strongly homophilous (mean fraud share of neighbors 0.25--0.48 for fraud against 0.06--0.12 for benign transactions), and S-FFSD is simulated, which we keep in mind when interpreting it. \textbf{Tolokers} \citep{platonov2023critical}: 11{,}758 crowd workers (21.8\% banned), 10 features, one relation (W-T-W: worked on a common task); fraud-node degrees are high (median 38, mean 136). Features are standardized, and we use stratified 40/20/40 splits of the evaluated nodes (five seeds for every real graph, seeds 3 and 4 of YelpChi, Amazon, and S-FFSD without CARE-GNN, GNNGuard, BWGNN, and adversarial training; three for the synthetic graph).

\paragraph{Relation-rich and augmented graphs.} The relation-rich CSBMs have $N{=}6000$ nodes (10\% fraud), $m{=}0.35$, and $R\in\{4,8,12\}$ relations; each relation is informative with probability 0.6 ($p\sim U[0.5,0.85]$, $q\sim U[0.02,0.1]$) and uninformative otherwise ($p{=}q{=}0.1$), has degree $d\in\{2,4,10,30\}$, and with probability 0.3 a fraction 0.5 or 0.8 of the nodes has no edge on it. Injection and deletion costs are drawn independently and log-uniformly in $[0.25,16]$ and rounded to multiples of 0.25, so that a relation's cost carries no information about its signal. Augmented real graphs add to FDCompCN, Tolokers, and Amazon six attribute-sharing relations: for a random feature, nodes are sorted by it and cut into bins of about 50, and each node is linked to three random nodes of its bin (symmetrized); their costs are drawn in the same way, and the original relations keep their profile costs.

\begin{table}[ht]
\centering
\small
\setlength{\tabcolsep}{5pt}

\caption{Cost profiles: unit costs $(\cin,\cdel)$ per relation.}
\label{tab:profiles}

\begin{tabular}{lll}
\toprule
Dataset & Profile & Costs \\
\midrule
Synthetic & $\cid$ & identity $(\cid,\cid)$, behavior $(2,4)$, dense $(1,4)$; P1: $\cid{=}0.5$, P2: $\cid{=}16$ \\
\addlinespace
YelpChi & P1 (cheap identities) & R-U-R $(2,0.5)$, R-T-R $(1,4)$, R-S-R $(1,4)$ \\
 & P2 (costly identities) & R-U-R $(25,25)$, R-T-R $(1,4)$, R-S-R $(1,4)$ \\
 & entity Q1 / Q2 & aged account 2 / 25, fresh account 0.5 / 25, rating 1, timing 1 \\
\addlinespace
Amazon & P1 & U-P-U $(1,4)$, U-S-U $(1,2)$, U-V-U $(1,0.5)$ \\
 & P2 & U-P-U $(1,4)$, U-S-U $(1,2)$, U-V-U $(8,8)$ \\
\addlinespace
FDCompCN & P1 (concealable ownership) & C-I-C $(4,0.5)$, C-P-C $(1,2)$, C-S-C $(1,2)$ \\
 & P2 (transparent registries) & C-I-C $(8,8)$, C-P-C $(1,2)$, C-S-C $(1,2)$ \\
\addlinespace
S-FFSD & P1 (cheap cards) & T-S-T $(2,0.5)$, T-M-T, T-L-T, T-Y-T $(1,4)$ \\
 & P2 (verified cards) & T-S-T $(25,25)$, T-M-T, T-L-T, T-Y-T $(1,4)$ \\
\addlinespace
Tolokers & P1 / P2 & W-T-W $(1,1)$ / $(8,8)$ \\
\bottomrule
\end{tabular}
\end{table}

\paragraph{Models.} MLP and HGB-X (histogram gradient boosting, 300 iterations, learning rate 0.05, balanced class weights) use features only; LR is logistic regression on features (RLR without relations). RLR is a logistic regression on $[x_v,\bar x_{v,1},\dots,\bar x_{v,R}]$. R-GCN has two layers with 32 units, mean aggregation per relation, and a separate ego weight. HGB-Graph is gradient boosting on features and relation means \citep{tang2023gadbench}; a gated relation is a missing value where the gate is closed, which trees handle natively. CARE-GNN is our vectorized re-implementation of the official code (label-aware similarity, top-$p$ neighbor selection with the reinforcement rule, official hyperparameters). GNNGuard-style: per relation and layer, edges are weighted by the cosine similarity of the endpoints' representations, edges below 0.1 are pruned, and weights are renormalized \citep{zhang2020gnnguard}. BWGNN-style: an MLP encoder, the three Beta-wavelet filters of order 2 per relation, a maximum over relations, and an MLP head \citep{tang2022bwgnn}. CA-AT is an R-GCN trained against the stronger of the gradient and query attacks at the evaluated profile and budget \citep{madry2018pgd}, from epoch 40 (one third of training) every 12 epochs, with checkpoints selected on the mean of clean and attacked validation AUC; AT-RLR is RLR trained the same way with the exact attack of \cref{prop:exact} as inner attack, so its training and its evaluation are exact; CA-HGB refits trees on the aggregates of attacked training fraud nodes. Neural models are trained full-batch with Adam (learning rate 0.01, weight decay $5\times10^{-4}$) and class-weighted cross-entropy for 120 epochs on YelpChi and Amazon and 200 elsewhere (300 for RLR), selecting checkpoints on validation AUC.

\begin{algorithm}[ht]
\caption{Cost-aware relation selection (\cars{}, \cars{}-D)}
\label{alg:cars}
\begin{algorithmic}[1]
\REQUIRE graph, training and validation labels, planning costs $(\cin,\cdel)$, budget $B$, robustness factor $\rho$, detector family $\mathcal F$
\STATE fit a relational logistic regression; project ego features and each relation on its direction
\STATE cluster training nodes by log-degree vectors; estimate group moments, $\alpha_r$, $z_r$, $p_r$
\STATE $V(\tau)\leftarrow$ plug-in worst-case AUC at costs $(\cin,\cdel)/\rho$ of the clean-optimal linear detector using relation $r$ on nodes of degree $\ge\tau_r$ (mixture of per-group SLSQP best responses)
\STATE $\tau^{\textsc{cars}}\leftarrow\arg\max V$ over $\tau\in\{0,\infty\}^R$ \hfill(subset selection)
\STATE $\tau^{\textsc{d}}\leftarrow$ coordinate ascent of $V$ over a grid of thresholds from $\tau^{\textsc{cars}}$
\STATE train detectors of $\mathcal F$ for $\tau^{\textsc{cars}}$ and $\tau^{\textsc{d}}$
\STATE \textbf{return} \cars{}: the detector for $\tau^{\textsc{cars}}$; \cars{}-D: among the graph-free counterpart and the two trained detectors, the one with the best evaluated worst-case validation AUC
\end{algorithmic}
\end{algorithm}

\paragraph{\cars{} details.} The plug-in uses $K=8$ fraud and $6$ benign degree groups ($k$-means on $\log(1+d)$ vectors, group degree = median), a candidate pool of the 300 most benign-looking labeled benign nodes, and the threshold grid $\{0,1,2,3,5,8,13,21,34,55,\infty\}$. Per-unit costs of an action that neutralizes a relation with fewer than one edit are inflated accordingly. A relation is informative if its fraud-group mean exceeds its benign-group mean. The per-group best response is a multi-start SLSQP over the edit counts of the relations whose gate is open, enumerating the gate actions (keep, close, open) that are affordable. The validation check of \cars{}-D uses the same attacks as the evaluation (query, gradient for neural models, exact for RLR, exhaustive-greedy for R-GCN and trees, and transfers) at the planning costs; its graph-free candidate is the counterpart itself (LR, the MLP, or HGB-X). A full \cars{}-D planning takes 2--90 seconds on one CPU core.

\paragraph{Attacks.} Every test fraud node is attacked with per-node budget $B$, and we report the per-node minimum score over the applicable attacks, an upper bound on the true worst case. The \emph{query} attack evaluates every allocation of the budget to one action, two actions, or all actions uniformly, with three endpoint rules for injections (the most benign-looking candidates, candidates scored like the target, and typical benign nodes from the interquartile range of benign scores, which defeat detectors that learn to distrust extremely benign neighbors), and keeps per target the most damaging configuration. The \emph{gradient} attack relaxes the candidate edits of each target to $[0,1]$, runs projected descent on the target logits with a per-target projection onto the cost polytope, and rounds randomly under the exact budget \citep{geisler2021prbcd}; degree gates are smoothed during the descent so that it sees activation and deactivation. Two \emph{transfer} attacks apply the gradient edits computed on the clean R-GCN and the exact edits computed on the clean RLR. For RLR models, gated or not, the \emph{exact} attack of \cref{prop:exact} is added; it covers opening and closing gates, since a relation contributes zero whenever the post-attack degree is below the threshold. Hard-edge models (trees, CARE-GNN, GNNGuard, BWGNN) face query and transfer attacks. Fraudsters know their accomplices; with the label oracle removed (\cref{tab:knowledge}), injection endpoints are ranked by the detector's scores among all non-target nodes. The black-box attack transfers gradient edits from a surrogate R-GCN trained on a different labeled sample. Entity-level attacks on YelpChi reassign a review's user (aged or fresh account), rating group, or timing and rebuild every induced edge. Gangs share $\bar B|T|$ and allocate per-node budget levels with a greedy multiple-choice knapsack on the label-free estimate $\sum_v\hat F(s_v)$ of the AUC. The \emph{exhaustive-greedy} attack, used for R-GCN and tree models in every evaluation and in every validation check, scores each candidate move of a target exactly: injecting any node of a candidate pool (3{,}000 nodes for R-GCN, 1{,}000 for trees: a third with the lowest scores, a third with the most extreme features, a third random) into any affordable relation, deleting any accomplice tie, and, for gated relations, injecting the best candidates until a gate opens or deleting accomplices until it closes. For R-GCN, exact single-target scoring recomputes the target's two-layer receptive field, including the change in an injected node's own aggregation when the target joins it; for trees, it recomputes the target's aggregate features. The most damaging move is applied while it lowers the score, after the first step among the 100 (R-GCN) or 50 (trees) best candidates per relation, and a target stops once its score is below the 0.5\% quantile of benign scores. Transfers of its edits computed on the clean R-GCN and HGB-Graph are added for the other detectors. The \emph{realized} joint attack applies, for every target, the edits of the attack that minimized its own score, all together, and rescores every node. \emph{Certificates} follow \cref{prop:cert}: exact per-target worst cases for (gated) RLR; for R-GCN, per-unit exact layer-1 terms, interval bounds on the layer-1 states of neighbors and candidates over every way the target's edits change their neighborhoods, per-unit layer-2 terms, and enumeration of joint budget allocations (or a dynamic program over budget splits when they exceed 4{,}000); for trees, per-relation boxes of reachable neighbor means for every budget share, a covering set of budget splits, and the sum over trees of the smallest reachable leaf. Exact attacks discretize costs on the largest grid of step $0.5/2^j$ that represents them.

\paragraph{Compute.} All experiments run on two CPU cores with 8 GB of memory. The second campaign, with the exhaustive-greedy attack in every evaluation, takes about 100 core-hours, most of it in the main table on YelpChi, Amazon, and S-FFSD (2--4.5 hours per seed); the third campaign (decomposition, hardening, and cost-uncertainty studies) takes about 35 core-hours, most of it in the plug-in planning of the 12-relation CSBMs (one to five hours per graph), whose per-group search enumerates gate actions and grows exponentially with the number of gated relations; the accompanying notebook regenerates every table and figure from cached records in minutes.

%% file: appendix_threat.tex
\section{Threat Model, Protocol, and Scope}
\label{apd:threat}

\paragraph{What an attacker edits.} Relations in these datasets are generated by behavior, not edited directly: an edge appears when two reviews share a user, two firms disclose a supplier tie, or two transactions use the same card. An attacker therefore acts on future events that will generate edges. \Cref{tab:threat} lists, for every relation, the behavior behind an injection and a deletion and the rationale of the two cost profiles. Our edge-level action model abstracts these behaviors as counts with unit costs; on YelpChi, where the entities behind the relations are recoverable, the entity-level attacks of \cref{tab:entity} rebuild every induced edge at once and confirm the edge-level conclusions. Costs are relative difficulties, not prices; only their ratio to the budget matters ($\Theta_{\alpha B}(\alpha c)=\Theta_B(c)$), and \cref{fig:costmap} reports every conclusion over a continuum of cost scales rather than at two profiles.

\begin{table}[ht]
\centering\scriptsize
\caption{Behavior behind each action and rationale of the profiles (P1 / P2).}
\label{tab:threat}
\resizebox{\linewidth}{!}{%
\begin{tabular}{lll p{5.2cm} p{5.2cm}}
\toprule
Dataset & Relation & Edge means & Inject / delete & Profiles\\
\midrule
YelpChi & R-U-R & same user & post from an account with history / use a fresh account per review & P1: fresh accounts are cheap; P2: identity verification makes both costly\\
 & R-T-R & same product, month & review a product in a given month / shift timing & cheap to create, costly to avoid\\
 & R-S-R & same product, rating & choose a rating / change it against the campaign's goal & cheap to create, costly to avoid\\
Amazon & U-P-U & common product & review a product / avoid products reviewed by accomplices & avoiding is costly\\
 & U-S-U & same rating, same week & rate in a week / spread over time & moderate\\
 & U-V-U & similar review text & write similar text / paraphrase & P1: paraphrasing is cheap; P2: text checks make both costly\\
FDCompCN & C-I-C & investment tie & disclose a holding / hold through undisclosed intermediaries & P1: concealable ownership; P2: transparent registries\\
 & C-P-C, C-S-C & supplier, customer tie & disclose a trade tie / route trade through third parties & moderate\\
S-FFSD & T-S-T & same card & reuse a shared card / use a new card & P1: cards are cheap; P2: verified cards\\
 & T-M-T, T-L-T, T-Y-T & merchant, location, type & transact at a merchant / abandon it & cheap to create, costly to avoid\\
Tolokers & W-T-W & common task & join a task / avoid tasks with accomplices & P1: unit costs; P2: costly participation control\\
\bottomrule
\end{tabular}}
\end{table}

\paragraph{Information used by each component.} \Cref{tab:protocol} lists the labels and structure each component sees. The graph structure and features are those of the full graph (transductive setting); no component uses test labels except for computing metrics. The attacker is strong by design: it knows the detector, its accomplices, and which labeled nodes are benign; \cref{tab:knowledge} removes the label oracle and the white-box access.

\begin{table}[ht]
\centering\scriptsize
\caption{Information available to each component (tr, va, te: training, validation, test labels).}
\label{tab:protocol}
\begin{tabular}{lll}
\toprule
Component & Labels & Structure and features\\
\midrule
Detector training & tr & clean full graph\\
\cars{} plug-in moments & tr & degrees and features of the full graph\\
\cars{}-D validation check, \cars{}+AT pick & va (attacked validation fraud) & clean full graph; attacks as below\\
Attack candidates for injections & labeled benign nodes (tr, va, te) & detector scores (white box)\\
Deletions & accomplices (fraud neighbors) of each target & --\\
Test metrics & te & attacked graph\\
Temporal split (S-FFSD) & earliest 40\% / next 20\% / last 40\% & full graph, time feature removed\\
\bottomrule
\end{tabular}
\end{table}

\paragraph{Provenance of costs.} No cost in this paper is estimated from operational data. We distinguish three kinds: \emph{scenario} costs (P1, P2, Q1, Q2), chosen to represent cheap or verified identities and justified only by the behaviors of \cref{tab:threat}; \emph{synthetic} costs (cost maps, random and noisy costs), used to test whether a conclusion survives other cost vectors; and \emph{derived} quantities, such as the break-even cost of each relation (\cref{tab:breakeven}), which are computed from the data and tell a practitioner which cost would make a relation safe, without claiming what the cost is. Every conclusion that depends on costs is therefore a statement about a scenario or a cost region.

\paragraph{Scope of each claim.} \Cref{tab:claims} separates the four kinds of statements the paper makes, and what each does not cover.

\begin{table}[ht]
\centering\scriptsize
\caption{Four kinds of claims and their scope.}
\label{tab:claims}
\resizebox{\linewidth}{!}{%
\begin{tabular}{p{2.2cm} p{4.4cm} p{5.6cm}}
\toprule
Kind & What it guarantees & What it does not cover\\
\midrule
Exact & worst case of linear relational models and CSBM thresholds under the formalized actions and costs (\cref{prop:exact,thm:abandon,thm:gatedom}) & other detectors, other actions, misspecified costs\\
Certified & a lower bound on the true worst-case AUC of R-GCN and tree ensembles under the target-side model (\cref{prop:cert}) & cross effects between targets, actions outside the model; bounds can be loose\\
Statistical & the selected detector is at most $2\varepsilon$ below every candidate under the evaluated attacks, for i.i.d.\ validation nodes and separable evaluation (\cref{prop:val}) & attacks not evaluated, distribution shift, dependence between targets, cost errors\\
Empirical & AUC under the strongest attacks we implemented (exhaustive-greedy, gradient, query, transfers, exact) & the true worst case of nonlinear detectors, which pool search shows can be lower\\
Operational & not claimed & real adversaries, calibrated costs, dated labels\\
\bottomrule
\end{tabular}}
\end{table}

\paragraph{Relation to the workshop version.} \Cref{tab:delta} lists, for each contribution, the prior result, what is new, and what it makes possible; the workshop paper \citep{anon2026workshop} studied the phenomenon on the CSBM, YelpChi, and Amazon. \Cref{tab:rerun} lists the protocol under which every experiment was run.

\begin{table*}[ht]
\centering\scriptsize
\caption{Contributions relative to the workshop version and to the closest prior work: what was known, what is new, and what the new result changes in the conclusions one can draw. The two main results are marked $\star$; the others are tools or supporting results.}
\label{tab:delta}
\resizebox{\textwidth}{!}{%
\begin{tabular}{p{2.6cm} p{3.0cm} p{3.6cm} p{3.9cm} p{4.3cm}}
\toprule
Result & Workshop & Closest prior work & New here & What changes\\
\midrule
Threat model, trust trap & introduced & cost-sensitive adversaries \citep{dalvi2004adversarial,lowd2005adversarial,zhang2019costsensitive}, strategic classification \citep{hardt2016strategic}, priced node attacks \citep{liao2024riskeeper} & unchanged & --\\
Structure (\cref{prop:structure}), robust planning, maximin, hardening (\cref{cor:robust}) & cost monotonicity & edit-count budgets with equal costs \citep{zugner2018nettack,bojchevski2019certifiable} & per-node vs gang thresholds, non-concavity, maximin choice under cost sets, hardening corollary & a budget-only analysis cannot rank relations for hardening; ours does, exactly for linear models (\cref{tab:hardening})\\
$\star$ Exposure (\cref{prop:exposure,prop:gnn}) and abandonment (\cref{thm:abandon}) & CSBM, first order; linear response & degree and robustness of message passing \citep{zugner2019certifiable} & per-node exposure of message passing; abandonment for every detector & exposure per unit cost predicts which relation to harden without any attack (\HDexpHit{} of \HDn{} cases)\\
$\star$ Gating (\cref{prop:gate}, \cref{thm:gatedom}) & -- & neighbor filtering and relation reweighting \citep{dou2020caregnn,zhang2022rohe,zhang2020gnnguard} & strict dominance of degree gates over every selection or reweighting under isolation & in the CSBM with isolated nodes, no selection or reweighting matches a degree gate, within a sharp window; on real graphs gating gains exactly where the conditions hold (\cref{tab:gatereal})\\
Exact attack and certificates (\cref{prop:exact,prop:cert}) & -- & certified graph robustness \citep{bojchevski2019certifiable,zugner2019certifiable}, tree verification \citep{chen2019treeverif} & exact thresholds for linear relational models under heterogeneous costs; brackets for R-GCN and trees & reported worst cases of nonlinear models are labeled tight, informative, or loose (\cref{tab:bracket})\\
Validated selection (\cref{prop:val}) & -- & evaluation against adaptive attacks \citep{tramer2020adaptive} & finite-sample guarantee with radius growing in $\log K$ & few theory-proposed candidates beat cost-blind or random candidates at equal budget on relation-rich graphs (\cref{tab:decomp})\\
Gangs (\cref{prop:gang}) & experiments only & coordinated injections \citep{choi2025monti} & convex envelope with a cross-effect bound & coordinated threat analysis\\
Attacks & query, gradient, transfer & PR-BCD \citep{geisler2021prbcd} & exhaustive-greedy with exact single-target scoring, gate moves, pool search & much stronger evaluation\\
Defense & cost-aware adversarial training & adversarial training \citep{madry2018pgd} & plug-in, \cars{}, \cars{}-D, \cars{}-R, \cars{}+AT & a selection protocol that abstains safely\\
\bottomrule
\end{tabular}}
\end{table*}

\begin{table}[ht]
\centering\scriptsize
\caption{Protocol of every experiment: full suite (with the exhaustive-greedy attack in evaluation and validation) or earlier suite.}
\label{tab:rerun}
\resizebox{\linewidth}{!}{%
\begin{tabular}{p{4.6cm} p{2.2cm} p{4.6cm}}
\toprule
Experiment & Protocol & Seeds / scope\\
\midrule
Main table, brackets, intervals, selection, degree groups, metrics & full & 5 seeds per graph (YelpChi, Amazon, S-FFSD: seeds 3--4 without CARE-GNN, GNNGuard, BWGNN, and adversarial training)\\
Validation-only selection & full & 3 seeds (small graphs), 1 seed (large graphs)\\
Ablations, validation labels & full & 3 seeds per graph\\
Decomposition at equal evaluation budget & full (tree attack pool 300) & real and augmented graphs: seed 0; relation-rich CSBMs: 5 seeds ($R{=}4,8$), 3 seeds ($R{=}12$)\\
Which relation to harden & exact (RLR); full (HGB-Graph) & seed 0; relation-rich CSBMs: 5 seeds\\
Degree gating on real graphs, decisions under cost uncertainty, random cost distributions & exact attack & seed 0\\
Attack regret, certificates, exact-attack scaling & full & seed 0, 150 targets\\
Cost maps, noisy and cost-agnostic planning, misspecification & exact attack & seed 0 (misspecification: YelpChi)\\
Gangs, factorial study, gating check, adversarial training at matched compute & full & seed 0 (factorial: 2 seeds)\\
Temporal split and rolling windows (S-FFSD) & full & 3 seeds; 4 windows\\
Sybil, poisoning, features, label scarcity, attacker knowledge & full & seed 0; YelpChi, Tolokers, FDCompCN\\
Entity-level attacks (YelpChi) & entity attack; \cars{}-D validated under the full suite & 3 seeds\\
Budget sweeps, random cost vectors & full & seed 0\\
Phase diagram, gate-aware attack, stronger AT evaluation, sensitivity & earlier & \cref{apd:earlier}\\
\bottomrule
\end{tabular}}
\end{table}

%% file: appendix_results.tex
\section{Additional Results}
\label{apd:more}

\paragraph{All cells.} \cref{tab:mainfull} adds the P1, $B{=}20$ cell to \cref{tab:main}. At this budget every relation of every real graph can be neutralized, and \cars{}-D falls back to graph-free features everywhere except where the check accepts a gate.

\begin{table*}[ht]
\centering\scriptsize
\caption{Clean and evaluated worst-case test AUC at all cells (mean over seeds), with the conventions of \cref{tab:main}.}
\label{tab:mainfull}
\setlength{\tabcolsep}{3pt}
\begin{tabular}{l cccc cccc cccc cccc cccc}
\toprule
& \multicolumn{4}{c}{YelpChi} & \multicolumn{4}{c}{Amazon} & \multicolumn{4}{c}{FDCompCN} & \multicolumn{4}{c}{S-FFSD} & \multicolumn{4}{c}{Tolokers}\\
\cmidrule(lr){2-5}\cmidrule(lr){6-9}\cmidrule(lr){10-13}\cmidrule(lr){14-17}\cmidrule(lr){18-21}
Model & cl. & P1,2 & P1,20 & P2,20 & cl. & P1,2 & P1,20 & P2,20 & cl. & P1,2 & P1,20 & P2,20 & cl. & P1,2 & P1,20 & P2,20 & cl. & P1,2 & P1,20 & P2,20\\
\input{tables/tab_main_full.tex}
\bottomrule
\end{tabular}
\end{table*}

\paragraph{Attack strength and certified brackets.} \cref{tab:bracket} reports, for clean-trained and \cars{}-D detectors at P1, $B{=}2$ and P2, $B{=}20$, the strongest single attack of the earlier suite (query, gradient, transfers), the exhaustive-greedy attack, the evaluated worst case over all attacks (per-node minimum), the AUC of a single realized joint attack, and the certified lower bound of \cref{prop:cert}. \cref{tab:regret} samples 150 test fraud nodes per graph (P1, $B{=}2$, seed 0) and compares, per node, the earlier suite, the exhaustive-greedy attack, exhaustive search over every edit set within a pool of 24 candidates (a third most benign-looking, a third the best single injections, a third random) and all accomplice deletions, and the certificate; entries are each target's rank among benign test scores, averaged over targets (an AUC over the sampled targets).

\begin{table*}[ht]
\centering\scriptsize
\setlength{\tabcolsep}{3pt}

\caption{Evaluated and certified worst-case AUC (mean over seeds): clean AUC, strongest earlier attack, exhaustive-greedy (EG), all attacks (per-node minimum), single realized joint attack, certified lower bound, and the status of the bracket [certified, all]: tight (width $\le 0.01$, the true worst case is known to within 0.01), informative ($\le 0.05$), or loose (the true worst case is not pinned down and only the evaluated upper bound is reported).}
\label{tab:bracket}

\begin{tabular}{lll *{6}{r} l}
\toprule
Dataset & Cell & Model & clean & earlier & EG & all & realized & certified & bracket \\
\midrule
\input{tables/tab_bracket.tex}
\bottomrule
\end{tabular}
\end{table*}

\begin{table*}[ht]
\centering\scriptsize
\setlength{\tabcolsep}{3.5pt}

\caption{Attack regret (P1, $B{=}2$, seed 0, 150 sampled targets): AUC over the sampled targets when clean, under the earlier suite, the exhaustive-greedy attack, exhaustive search within the pool (combined with the other attacks), and the certified bound; share of targets where the exhaustive-greedy attack beats the suite and where pool search beats the exhaustive-greedy attack.}
\label{tab:regret}

\begin{tabular}{ll c *{5}{r} *{2}{r}}
\toprule
Dataset & Model & $n$ & clean & suite & EG & pool & certified & EG$<$suite & pool$<$EG \\
\midrule
\input{tables/tab_regret.tex}
\bottomrule
\end{tabular}
\end{table*}

\paragraph{Selection statistics and the validation guarantee.} \cref{tab:select} counts the configurations kept by \cars{}-D over seeds, families, and cells, and compares the radius $\varepsilon$ of \cref{prop:val} ($K{=}3$, $\delta{=}0.05$) with the observed change between validation and test worst-case AUC of every candidate. \cref{tab:valsel} compares \cars{}-D with selection by adversarial validation alone: all relation subsets (VAL-SUB), $\{$graph-free, all relations$\}$ (VAL-FA), and $k$ random configurations of subsets and gates including the graph-free one (VAL-RAND-$k$), with the test-optimal configuration among all those evaluated (oracle, small graphs).

\begin{table}[ht]
\centering\scriptsize
\setlength{\tabcolsep}{2.5pt}

\caption{Configurations kept by \cars{}-D (share of runs) and the validation guarantee: mean $\varepsilon$, $2\varepsilon$, largest and mean validation-to-test change over candidates.}
\label{tab:select}

\begin{tabular}{lc *{7}{r}}
\toprule
Dataset & runs & free & \cars{} & gated & $\varepsilon$ & $2\varepsilon$ & max $|\Delta|$ & mean $\Delta$\\
\midrule
\input{tables/tab_select.tex}
\bottomrule
\end{tabular}
\end{table}

\begin{table*}[ht]
\centering\scriptsize
\setlength{\tabcolsep}{3.5pt}

\caption{Test worst-case AUC of the selected detector: plug-in only (\cars{}), \cars{}-D, and validation-only rules; last column: validation evaluations of \cars{}-D / VAL-SUB.}
\label{tab:valsel}

\begin{tabular}{lll *{8}{r}}
\toprule
Dataset & Family & Cell & \cars{} & \cars{}-D & VAL-FA & RAND-3 & RAND-8 & SUB & oracle & evals\\
\midrule
\input{tables/tab_valsel.tex}
\bottomrule
\end{tabular}
\end{table*}

\paragraph{Degree groups and operating metrics.} \cref{tab:groups} splits test fraud nodes by total degree and reports the evaluated worst-case AUC of each group against all benign test nodes (P1, $B{=}2$); \cref{tab:metrics} reports average precision and true-positive rates at 1\% and 5\% false-positive rate under attack.

\begin{table}[ht]
\centering\scriptsize
\setlength{\tabcolsep}{2pt}

\caption{Evaluated worst-case AUC by total degree of the fraud node (P1, $B{=}2$, mean over seeds).}
\label{tab:groups}

\begin{tabular}{ll c *{6}{r}}
\toprule
Dataset & Degree & $n_+$ & MLP & R-GCN & R-GCN+D & HGB-X & HGB-Graph & HGB+D \\
\midrule
\input{tables/tab_groups.tex}
\bottomrule
\end{tabular}
\end{table}

\begin{table}[ht]
\centering\scriptsize
\setlength{\tabcolsep}{4.5pt}

\caption{Operating metrics under the evaluated worst case (P1, $B{=}2$, mean over seeds).}
\label{tab:metrics}

\begin{tabular}{ll *{4}{r}}
\toprule
Dataset & Model & AUC & AP & TPR@1\% & TPR@5\% \\
\midrule
\input{tables/tab_metrics.tex}
\bottomrule
\end{tabular}
\end{table}

\paragraph{Gating against selection and reweighting.} \cref{tab:gatetheory} evaluates the closed-form worst-case AUC of \cref{thm:gatedom} (Gaussian CSBM moments, $m$ and $\delta$ of the base configuration, $p{=}0.8$, $q{=}0.02$): the best gated detector against the best ungated one over $w\ge0$, on a grid of weights. For $\pi_0\ge0.5$, the best ungated weight is $w=0$ in \GTungDrop{} of \GTnHi{} settings and the gated detector beats the relation-free one in \GTgain{}. \cref{tab:gategraph} repeats the comparison on CSBM graphs where a fraction of the nodes is isolated on the identity relation, with RLR and the exact attack ($B{=}2$, deletions on the identity relation made unaffordable). Clean-trained weights rely on the relation too heavily for any configuration to beat the relation-free detector; with the relation's weight scaled by a factor chosen on validation under the exact attack, the sup over $w$ of \cref{thm:gatedom}, the gated detector ($\tau{=}3$) beats both the best ungated one and the relation-free one by more than 0.002 in \GGwin{} of the four settings with $\cid{=}1$, where the attacker affords $b{=}2<\tau$ injections as the theorem requires, and is within 0.002 in the remaining one; with $\cid{=}0.5$, $b{=}4\ge\tau$, the gate can be opened, and gating is no better than dropping the relation.

\begin{table}[ht]
\centering\scriptsize
\setlength{\tabcolsep}{3.5pt}

\caption{Closed-form worst-case AUC, best gated / best ungated linear detector, by degree $d$ of connected nodes, injections $b$, and isolated fraction $\pi_0$; last column: relation-free detector.}
\label{tab:gatetheory}

\begin{tabular}{cc *{6}{r}}
\toprule
$d$ & $b$ & $\pi_0{=}0$ & $0.25$ & $0.5$ & $0.75$ & $0.9$ & free \\
\midrule
\input{tables/tab_gatetheory.tex}
\bottomrule
\end{tabular}
\end{table}

\begin{table}[ht]
\centering\scriptsize
\caption{CSBM graphs with an isolated fraction of nodes on the identity relation: exact worst-case AUC of RLR ungated and gated at $\tau{=}3$, with clean-trained weights or with the identity weight scaled on validation (robust), and without the identity relation ($B{=}2$, seed 0).}
\label{tab:gategraph}
\begin{tabular}{ccc ccccc}
\toprule
$p_{\mathrm{id}}$ & isolated & $\cid$ & ungated & ung.\ robust & dropped & gated & gated robust\\
\midrule
\input{tables/tab_gategraph.tex}
\bottomrule
\end{tabular}
\end{table}

\paragraph{Degree gating on real graphs.} \cref{tab:gatereal,tab:gaterealds} test \cref{thm:gatedom} outside the CSBM. For every relation of every real graph and the CSBM, alone with the ego features, RLR is evaluated exactly for budgets $B\in\{1,2,5,10,20\}$ at P1 and P2 (separable worst-case AUC, the dynamic program of \cref{prop:exact}, which covers opening a gate by injection and closing it by deletion): graph-free LR, the ungated detector with its relation weight block scaled by $\lambda\in\{0,0.1,0.25,0.5,0.75,1\}$ (reweighting, $\lambda{=}0$ is dropping), and the gated detector with $\tau\in\{1,2,3,5,8\}$ and the same scaling; $\lambda$ and $\tau$ are chosen on validation nodes at each budget. A win is a gated detector more than 0.002 above both the best reweighting and LR on test. The gate can be opened when $B\ge\cin_r\tau$.

\begin{table}[ht]
\centering\scriptsize
\caption{Validated degree gating against validated reweighting and dropping, per relation on real graphs and the CSBM (exact attack, five budgets, two profiles): cases, wins, ties ($\le0.002$), losses, and mean and largest gain, by whether the attacker can open the chosen gate.}
\label{tab:gatereal}
\setlength{\tabcolsep}{2.5pt}
\begin{tabular}{l cccccc}
\toprule
Regime & cases & wins & ties & losses & mean & max\\
\midrule
\input{tables/tab_gatereal.tex}
\bottomrule
\end{tabular}
\end{table}

\begin{table}[ht]
\centering\scriptsize
\caption{Degree gating per relation: share of isolated test fraud nodes, wins/losses/cases at P1 and P2, and largest gain.}
\label{tab:gaterealds}
\setlength{\tabcolsep}{2.5pt}
\begin{tabular}{ll c cc c}
\toprule
Graph & Relation & isolated & P1 & P2 & max\\
\midrule
\input{tables/tab_gatereal_ds.tex}
\bottomrule
\end{tabular}
\end{table}

\paragraph{What each part of \cars{}-D contributes.} \cref{tab:ablation2} restricts the candidate set at equal attack budget (each validation check costs one attack evaluation per candidate): the plug-in's choice used without validation, validation between two of the three candidates, and \cars{}-D (five graphs, three seeds, two cells, three families). \cref{tab:selacc} measures whether \cars{}-D keeps a relation when one is worth keeping: a configuration \emph{should} be kept when some evaluated configuration beats the graph-free one on test by more than 0.002 (\cref{tab:valsel}). \cref{tab:labels} repeats the validation check on random fractions of the validation labels (20 draws per fraction).

\begin{table}[ht]
\centering\scriptsize
\setlength{\tabcolsep}{3.5pt}

\caption{Ablation at equal attack budget: number of candidates, mean and worst test difference with the graph-free candidate, runs below it, and mean regret to the test-optimal candidate.}
\label{tab:ablation2}

\begin{tabular}{lc *{4}{r}}
\toprule
Rule & cands & mean $\Delta$ & worst $\Delta$ & below & regret \\
\midrule
\input{tables/tab_ablation2.tex}
\bottomrule
\end{tabular}
\end{table}

\begin{table}[ht]
\centering\scriptsize
\setlength{\tabcolsep}{2.5pt}

\caption{Selection accuracy of \cars{}-D: runs, share picking the test-optimal candidate, runs where a relation should be kept, kept, both, and mean / largest regret.}
\label{tab:selacc}

\begin{tabular}{lc *{6}{r}}
\toprule
Dataset & runs & optimal & should & kept & both & mean reg. & max reg. \\
\midrule
\input{tables/tab_selacc.tex}
\bottomrule
\end{tabular}
\end{table}

\begin{table}[ht]
\centering\scriptsize
\setlength{\tabcolsep}{3pt}

\caption{\cars{}-D with a fraction of the validation labels: median validation fraud nodes, mean, 5\% quantile and worst test difference with the graph-free candidate, share of draws below it, and mean regret.}
\label{tab:labels}

\begin{tabular}{lc *{5}{r}}
\toprule
Labels & $n_+$ & mean $\Delta$ & 5\% & worst & below & regret \\
\midrule
\input{tables/tab_labels.tex}
\bottomrule
\end{tabular}
\end{table}

\paragraph{Decomposition at equal evaluation budget.} \cref{tab:decomp} isolates each ingredient of \cars{}-D. Every candidate configuration that any rule may check is attacked once on all validation nodes and on the test set (full suite; for trees the exhaustive-greedy pool has 300 candidates); each rule then checks its candidates on the validation nodes and keeps the one of highest evaluated worst-case AUC, ties going to the simpler candidate. Rules: \cars{}-D $\{$graph-free, plug-in subset, plug-in gates$\}$; the same planner with every cost set to the geometric mean of the true costs (cost-blind planning); the plug-in's subset or gates without validation; \cars{}-D without the gated or without the graph-free candidate; VAL-FA $\{$graph-free, all relations$\}$; VAL-RAND-$k$, the graph-free configuration and $k{-}1$ random configurations of subsets and gates; VAL-SUB, every relation subset ($R\le4$); and VAL-POOL, every configuration evaluated for any rule. Graphs: the CSBM, FDCompCN, Tolokers, and Amazon (real; YelpChi and S-FFSD are covered by \cref{tab:valsel}), FDCompCN, Tolokers, and Amazon with six attribute-sharing relations added (each node linked to three random nodes in its bin of about 50 nodes along a random feature, as when accounts share a device or an address; costs drawn log-uniformly in $[0.25,16]$; augmented real), and CSBMs with $R\in\{4,8,12\}$ relations of random signal, degree, and isolation whose costs are drawn independently of their signal (relation-rich, five seeds for $R\in\{4,8\}$ and three for $R{=}12$; \cref{apd:setup}). Families RLR (exact attack) and HGB-Graph; two cells per graph. The planner searches subsets exhaustively up to six eligible relations and by local search beyond, so its number of plug-in evaluations grows linearly in $R$ while VAL-SUB needs $2^R$ attack evaluations.

\begin{table*}[ht]
\centering\scriptsize
\caption{Selection rules at equal evaluation budget: number of validated candidates $K$ (median), mean test worst-case AUC difference with the graph-free detector and number of cases more than 0.002 below it, per graph family (full validation labels), mean regret to the best evaluated configuration on relation-rich graphs, and the difference and share of draws below the graph-free detector with 10\% of the validation labels on relation-rich graphs (20 draws).}
\label{tab:decomp}
\setlength{\tabcolsep}{3pt}
\begin{tabular}{l c cc cc ccc cc}
\toprule
& & \multicolumn{2}{c}{Real} & \multicolumn{2}{c}{Augmented real} & \multicolumn{3}{c}{Relation-rich CSBM} & \multicolumn{2}{c}{Rich, 10\% labels}\\
\cmidrule(lr){3-4}\cmidrule(lr){5-6}\cmidrule(lr){7-9}\cmidrule(lr){10-11}
Rule & $K$ & $\Delta$ & below & $\Delta$ & below & $\Delta$ & below & regret & $\Delta$ & below\\
\midrule
\input{tables/tab_decomp.tex}
\bottomrule
\end{tabular}
\end{table*}

\cref{tab:decompds} reports the same comparisons per graph, as paired differences of \cars{}-D with each alternative and 95\% bootstrap intervals (resampling seeds, or cases when there is one seed; cases are cells and families, so intervals over four cases are only indicative).

\begin{table*}[ht]
\centering\scriptsize
\caption{Per-graph paired differences in test worst-case AUC (full validation labels): \cars{}-D minus the graph-free detector, minus cost-blind planning, minus the plug-in's gated choice without validation, minus \cars{}-D without the gated candidate, minus VAL-RAND-8, and minus VAL-POOL; mean with 95\% bootstrap interval; seeds and cases (seeds $\times$ cells $\times$ families).}
\label{tab:decompds}
\setlength{\tabcolsep}{2pt}
\resizebox{\textwidth}{!}{%
\begin{tabular}{l cc cccccc}
\toprule
Graph & seeds & cases & vs graph-free & vs cost-blind & vs no validation & vs no gated candidate & vs VAL-RAND-8 & vs VAL-POOL\\
\midrule
\input{tables/tab_decomp_ds.tex}
\bottomrule
\end{tabular}}
\end{table*}

\paragraph{Which relation to harden.} \cref{tab:hardening} multiplies the injection and deletion costs of one relation at a time by 1000 (unmanipulable within the budget grid) and computes the exact trap threshold of the clean-trained RLR with one dynamic program per relation (\cref{prop:exact}), on every graph, seed, and profile where RLR has a clean gain over LR and is trapped within $B\le64$: the CSBM (P1, P2), YelpChi (P2), and CSBMs with 4, 8, and 12 relations and random costs (five seeds each). A rule is right if it picks a relation whose hardening raises the threshold most. Rules: the first-order exposure per unit cost of \cref{prop:exposure,prop:gnn}, measured as the largest score drop of one injection or one deletion per unit cost, averaged over fraudulent test nodes (no attack is run); the exact analysis under equal costs (the geometric mean), as a budget-only robustness analysis would do; the most informative relation (largest validation AUC loss when dropped); the largest weight norm; and the cheapest relation. Value: threshold gain of the pick relative to the best. \HDzeroShare\% of the \HDsingleN{} single-relation hardenings leave the threshold unchanged, as \cref{cor:robust}(c) predicts for relations off the cheapest trap policy. For HGB-Graph (full suite, $B{=}2$; the CSBM at P2, YelpChi at P2, and three relation-rich CSBMs), making the RLR-exposure pick unmanipulable is also the best single hardening for the trees in \HHhit{} of \HHn{} graphs and raises their evaluated worst-case AUC by \HHgainMean{} on average: the first-order ranking transfers only partly to nonlinear detectors.

\begin{table}[ht]
\centering\scriptsize
\caption{Which relation to harden: share of cases where each rule picks a relation whose hardening raises the exact trap threshold of RLR most, mean share of the best gain obtained, and hit rate on graphs with at least 8 relations.}
\label{tab:hardening}
\setlength{\tabcolsep}{3pt}
\begin{tabular}{l ccc}
\toprule
Rule & right & value & right ($R\ge8$)\\
\midrule
\input{tables/tab_hardening.tex}
\bottomrule
\end{tabular}
\end{table}

\paragraph{Decisions under cost uncertainty.} \cref{tab:voi} gives the defender a cost estimate $\hat c=c\,e^{\epsilon}$ ($\epsilon\sim\mathcal N(0,1)$ per coordinate, four draws) and a box of log half-width $w$ around it, on YelpChi, Amazon, FDCompCN, Tolokers, and the CSBM (P1, $B{=}2$ and P2, $B{=}20$). Naive: RLR + \cars{}-D planned at $\hat c$. Maximin: planned at the lower corner of the box, which is the maximin choice by \cref{cor:robust}(b). Minimax regret: among the relation subsets of RLR (graph-free, singletons, all), the one of smallest maximum regret over the lower corner, the centre, and eight sampled corners of the box. Every choice is evaluated exactly at the true costs. Coverage: share of draws in which the box contains the true costs.

\begin{table}[ht]
\centering\scriptsize
\caption{Decisions under cost uncertainty: box coverage and, for each rule, the mean exact worst-case AUC difference with LR at the true costs and the number of draws more than 0.002 below LR, by log half-width $w$ of the box.}
\label{tab:voi}
\setlength{\tabcolsep}{2.5pt}
\begin{tabular}{c c cc cc cc}
\toprule
& & \multicolumn{2}{c}{Naive} & \multicolumn{2}{c}{Maximin} & \multicolumn{2}{c}{Minimax regret}\\
\cmidrule(lr){3-4}\cmidrule(lr){5-6}\cmidrule(lr){7-8}
$w$ & covered & $\Delta$ & below & $\Delta$ & below & $\Delta$ & below\\
\midrule
\input{tables/tab_voi.tex}
\bottomrule
\end{tabular}
\end{table}

\begin{table*}[ht]
\centering\scriptsize
\caption{\cars{}-D (and \cars{}+AT for trees) minus its graph-free counterpart: mean evaluated worst-case AUC difference with 95\% bootstrap interval. Bold: interval above zero; underlined: interval below zero.}
\label{tab:ci}
\setlength{\tabcolsep}{2.0pt}
\resizebox{\textwidth}{!}{%
\begin{tabular}{l cc cc cc cc}
\toprule
& \multicolumn{2}{c}{RLR + \cars{}-D} & \multicolumn{2}{c}{R-GCN + \cars{}-D} & \multicolumn{2}{c}{HGB + \cars{}-D} & \multicolumn{2}{c}{\cars{}+AT (HGB)}\\
\cmidrule(lr){2-3}\cmidrule(lr){4-5}\cmidrule(lr){6-7}\cmidrule(lr){8-9}
Dataset & P1, $B{=}2$ & P2, $B{=}20$ & P1, $B{=}2$ & P2, $B{=}20$ & P1, $B{=}2$ & P2, $B{=}20$ & P1, $B{=}2$ & P2, $B{=}20$\\
\midrule
\input{tables/tab_ci.tex}
\bottomrule
\end{tabular}}
\end{table*}

\paragraph{Stability under plausible cost distributions.} \cref{tab:coststab} replaces the two profiles by 20 random cost vectors from each of two distributions. \emph{Independent}: every injection and deletion cost log-uniform in $[0.25,16]$. \emph{Structured}, following the behaviors of \cref{apd:threat}: the identity-like relation is expensive to enter (a new or verified account, $[4,32]$) and cheap to leave ($[0.25,1]$), and the other relations are cheap to enter ($[0.25,2]$) and harder to leave, since a shared product, merchant, or task cannot be unshared without changing behavior ($[1,8]$). For each vector and $B\in\{2,20\}$, the clean-trained RLR and RLR + \cars{}-D planned at the true costs are evaluated exactly. The questions are whether the paper's qualitative conclusions survive: is the clean model trapped, does \cars{}-D fall below LR, and when does it gain. The trap persists in \CStrap\% of the \CSnGain{} cases where RLR has a clean gain. \cars{}-D gains in \CSgain{} of \CSn{} cases, most often on YelpChi under the structured distribution, where the identity relation is expensive to enter, and falls below LR in \CSbelow{}, by at most \CSbelowWorst{}; \CSbelowFD{} of these are on FDCompCN, whose relations carry no clean signal, so a validation check that keeps one is wrong by a margin within the radius $2\varepsilon$ of \cref{prop:val} but not zero.

\begin{table*}[ht]
\centering\scriptsize
\caption{Conclusions under random cost vectors (RLR, exact attack, 20 vectors per distribution, $B\in\{2,20\}$): share of trapped cases among those where RLR has a clean gain, number of cases where \cars{}-D is more than 0.002 below LR and its worst difference, number of gains and their mean, median of the smallest cost (injection or deletion) of the relations kept when it gains, and median of the smallest cost over all actions.}
\label{tab:coststab}
\setlength{\tabcolsep}{3pt}
\begin{tabular}{ll c c cc cc cc}
\toprule
Graph & Distribution & cases & trapped & below & worst & gains & mean gain & kept cost & all costs\\
\midrule
\input{tables/tab_coststab.tex}
\bottomrule
\end{tabular}
\end{table*}

\paragraph{Confirmatory comparisons in full.} \cref{tab:cifull} lists the 45 confirmatory comparisons with unrounded mean differences, the standard deviation of the per-seed difference, the hierarchical-bootstrap interval, the Holm-adjusted $p$-value, non-inferiority at a margin of 0.01 AUC (lower end of the interval above $-0.01$), and the smallest margin at which non-inferiority holds. Exact ties (difference 0) are cases in which \cars{}-D returned the graph-free detector in every seed. The margin of 0.01 was not fixed before the analysis; the last column lets a reader apply any margin.

\begin{table*}[ht]
\centering\scriptsize
\caption{Confirmatory comparisons of \cars{}-D with its graph-free counterpart (five seeds): mean difference, seed standard deviation, 95\% hierarchical-bootstrap interval, Holm-adjusted $p$, non-inferiority at margin 0.01, and smallest margin with non-inferiority.}
\label{tab:cifull}
\setlength{\tabcolsep}{3pt}
\begin{tabular}{lll cccccc}
\toprule
Dataset & Family & Cell & difference & SD & 95\% interval & $p_{\mathrm{Holm}}$ & NI(0.01) & margin\\
\midrule
\input{tables/tab_ci_full.tex}
\bottomrule
\end{tabular}
\end{table*}

\begin{figure*}[ht]
\centering
\includegraphics[width=\textwidth]{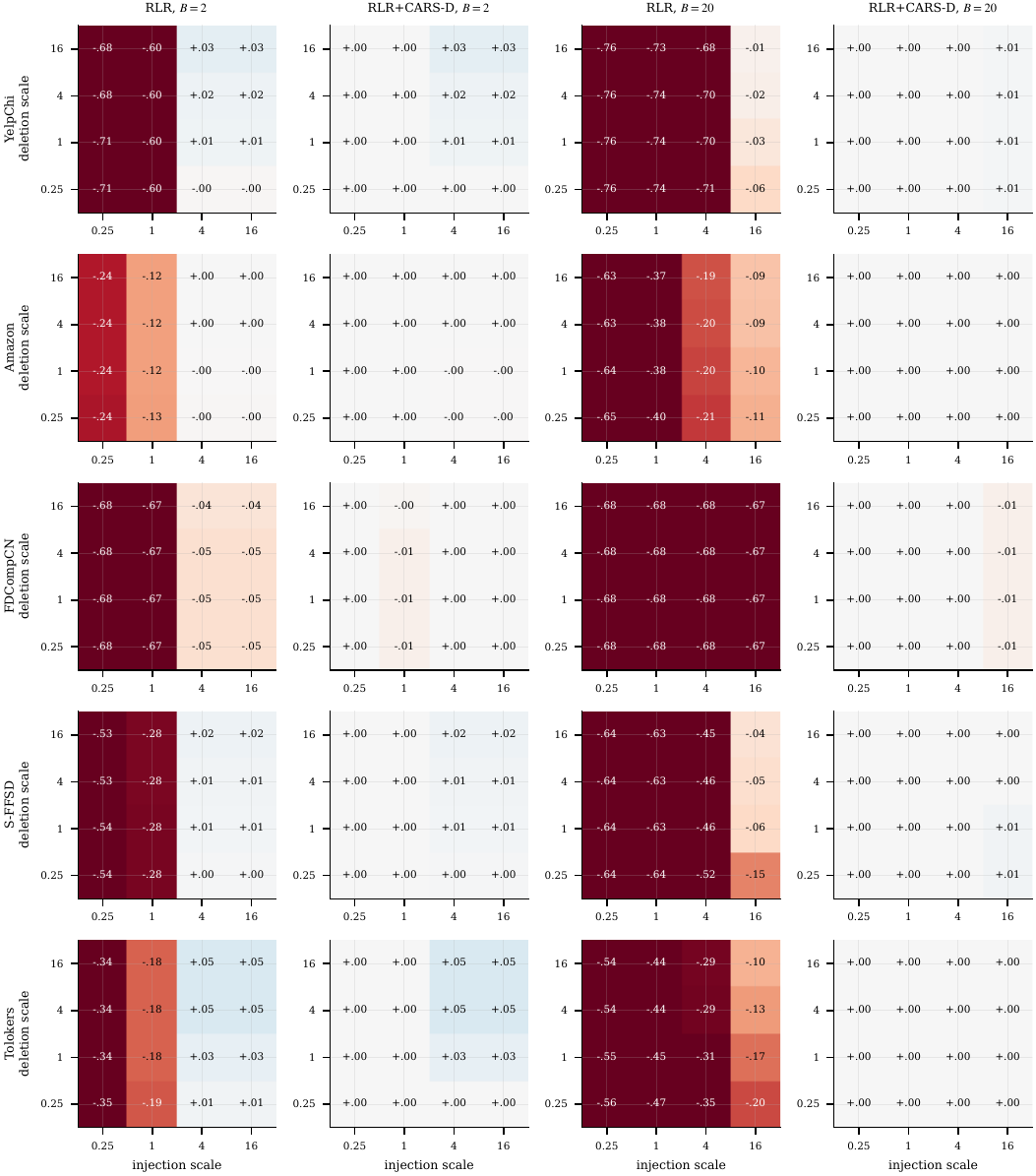}
\caption{Exact cost maps: worst-case AUC minus LR of the clean-trained RLR and of RLR + \cars{}-D planned at the true costs, over injection and deletion cost scales relative to P1, at $B{=}2$ and $B{=}20$ (seed 0).}
\label{fig:costmap}
\end{figure*}

\paragraph{Planning without cost estimates.} \cref{tab:costagnostic} plans RLR + \cars{}-D at the lower corner of the cost grid (every cost a quarter of P1), the plan of a defender who knows only a lower bound on costs, and evaluates it exactly at every cost scale; it compares it with the plan at the true costs of \cref{fig:costmap}. \cref{tab:noiserobust} plans at noisy cost estimates $\hat c=c\,e^{\epsilon}$ and at the lower corner $\hat ce^{-z\sigma}$ of a log box of half-width $z\sigma$, and evaluates at the true costs. \cref{tab:breakeven} reports, for each relation with a clean gain, its exact single-relation threshold at P1 and the factor by which its costs would have to grow for it to resist budgets of 2 and 20 (by homogeneity, \cref{prop:structure}).

\begin{table}[ht]
\centering\scriptsize
\setlength{\tabcolsep}{2.5pt}

\caption{Cost-agnostic planning (RLR, exact, seed 0, 16 cost scales): mean and worst difference with LR of the plan at the lower corner, mean difference of the plan at the true costs, mean and largest price of not knowing the costs, and cells below LR.}
\label{tab:costagnostic}

\begin{tabular}{ll *{6}{r}}
\toprule
Dataset & $B$ & agn.\ mean & agn.\ worst & true mean & price mean & price max & below \\
\midrule
\input{tables/tab_costagnostic.tex}
\bottomrule
\end{tabular}
\end{table}

\begin{table}[ht]
\centering\scriptsize
\setlength{\tabcolsep}{2.2pt}

\caption{Noisy cost estimates (RLR + \cars{}-D, P1, $B{=}2$, ten draws per $\sigma$): plan at the true costs, LR, and for box half-widths $z\in\{0,1,2\}$ the draws below LR and the mean regret to the true plan.}
\label{tab:noiserobust}

\begin{tabular}{ll *{8}{r}}
\toprule
Dataset & $\sigma$ & true & LR & \multicolumn{2}{c}{$z{=}0$} & \multicolumn{2}{c}{$z{=}1$} & \multicolumn{2}{c}{$z{=}2$} \\
\midrule
\input{tables/tab_noiserobust.tex}
\bottomrule
\end{tabular}
\end{table}

\begin{table}[ht]
\centering\scriptsize
\setlength{\tabcolsep}{3.5pt}

\caption{Break-even costs (P1, RLR, seed 0): clean gain over LR of the single-relation model, its exact trap threshold, and the cost multiplier needed to resist $B{=}2$ and $B{=}20$.}
\label{tab:breakeven}

\begin{tabular}{ll *{4}{r}}
\toprule
Dataset & Relation & gain & threshold & $\times$ for 2 & $\times$ for 20 \\
\midrule
\input{tables/tab_breakeven.tex}
\bottomrule
\end{tabular}
\end{table}

\paragraph{Rolling temporal windows.} \cref{tab:temporal2} cuts the time-ordered labeled transactions of S-FFSD into ten blocks and, for windows $w=0,\dots,3$, trains on blocks $w$ to $w{+}3$, validates on block $w{+}4$, and tests on blocks $w{+}5$ and $w{+}6$; it is a case study on simulated data, not evidence of robustness to strategic drift.

\begin{table}[ht]
\centering\scriptsize
\setlength{\tabcolsep}{2.5pt}

\caption{S-FFSD rolling windows: evaluated worst-case AUC by window and cell (seed 0).}
\label{tab:temporal2}

\begin{tabular}{ll *{6}{r}}
\toprule
Window & Cell & MLP & R-GCN & R-GCN+D & HGB-X & HGB-Graph & HGB+D \\
\midrule
\input{tables/tab_temporal2.tex}
\bottomrule
\end{tabular}
\end{table}

\paragraph{Cost maps and noisy costs.} \cref{fig:costmap} scales the P1 injection and deletion costs of every relation by factors in $\{1/4,1,4,16\}$ and reports the exact worst-case AUC (separable and joint attacks with the exact attack) of the clean-trained RLR and of RLR + \cars{}-D planned at the true costs, minus LR. \cref{tab:noise} plans RLR + \cars{}-D at costs multiplied by independent or common log-normal factors and evaluates it exactly at the true P1 costs.

\begin{table}[ht]
\centering\scriptsize
\setlength{\tabcolsep}{2.5pt}

\caption{Noisy planning costs (RLR + \cars{}-D, seed 0, five draws): plan at the true costs, mean and minimum of the noisy plans evaluated exactly at the true costs, LR, and share of draws with the same selection.}
\label{tab:noise}

\begin{tabular}{ll cl *{5}{r}}
\toprule
Dataset & $B$ & $\sigma$ & noise & true plan & noisy (mean) & noisy (min) & LR & same \\
\midrule
\input{tables/tab_noise.tex}
\bottomrule
\end{tabular}
\end{table}

\paragraph{Robust planning.} \cref{tab:misspec} plans HGB + \cars{}-D at either profile with robustness factor $\rho$ and evaluates it under both (seed 0). By \cref{cor:robust}, the guarantee of a plan at costs $c/\rho$ covers every true cost vector above $c/\rho$. On YelpChi at $B{=}20$, the P2 plan collapses when identities are in fact cheap, and $\rho\ge2$ restores graph-free performance under both profiles; at $B{=}2$, even $\rho{=}8$ leaves an R-U-R edit unaffordable in the plan ($25/8>2$) while P1 makes it cheap, so P1 lies outside the covered box and the plan fails, as the corollary allows.

\begin{table}[ht]
\centering\scriptsize
\caption{HGB + \cars{}-D planned at a profile with costs divided by $\rho$ and evaluated under true profiles P1/P2 (evaluated worst-case AUC, seed 0); last column: HGB-X.}
\label{tab:misspec}
\setlength{\tabcolsep}{2.5pt}
\begin{tabular}{llc cccc c}
\toprule
Dataset & $B$ & plan & $\rho{=}1$ & $\rho{=}2$ & $\rho{=}4$ & $\rho{=}8$ & HGB-X\\
\midrule
\input{tables/tab_misspec.tex}
\bottomrule
\end{tabular}
\end{table}

\paragraph{Exact thresholds of relation subsets.} \cref{tab:abandon} computes, with one exact dynamic program per relation subset, the smallest budget at which the separable worst-case AUC of RLR restricted to the subset falls to that of LR (seed 0).

\begin{table}[t]
\centering\scriptsize
\setlength{\tabcolsep}{3.5pt}

\caption{Exact trap thresholds of RLR (seed 0): single relations (in the order of \cref{tab:profiles}), the full model, the sum of single-relation thresholds, and the budget at which every subset is trapped.}
\label{tab:abandon}

\begin{tabular}{lll *{3}{r}}
\toprule
Dataset & Profile & single relations & full & sum & all subsets \\
\midrule
\input{tables/tab_abandon.tex}
\bottomrule
\end{tabular}
\end{table}

\paragraph{Gangs.} \cref{tab:gang2} lets a gang share an average budget $\bar B$ per member; each attacked member also pays a coordination cost $\kappa$, and the gang allocates its budget with the detector's own scores (oracle) or with a surrogate trained on another labeled sample. The realized AUC applies all edits together and rescores every node; its largest deviation from the convex-envelope prediction is \GKdevMax{}.

\begin{table*}[ht]
\centering\scriptsize
\setlength{\tabcolsep}{4pt}

\caption{Gangs (P1, seed 0): realized AUC at average budgets $\bar B\in\{0.5,1,2\}$ for oracle and surrogate allocation, by coordination cost $\kappa$; last column: graph-free trees.}
\label{tab:gang2}

\begin{tabular}{ll c *{7}{r}}
\toprule
Dataset & Model & $\kappa$ & \multicolumn{3}{c}{oracle} & \multicolumn{3}{c}{surrogate} & free \\
\cmidrule(lr){4-6} \cmidrule(lr){7-9}
\midrule
\input{tables/tab_gang2.tex}
\bottomrule
\end{tabular}
\end{table*}

\paragraph{Causal factors.} \cref{tab:factorial} regresses the trap depth (evaluated worst-case AUC minus the graph-free counterpart, $B{=}2$) on the CSBM factors: the identity relation's homophily $p_{\mathrm{id}}\in\{0.55,0.7,0.85\}$, the fraction of nodes isolated on it $\in\{0,0.5,0.85\}$, its cost $\cid\in\{0.5,4\}$, and the architecture (two seeds).

\begin{table}[ht]
\centering\scriptsize
\setlength{\tabcolsep}{6pt}

\caption{OLS coefficients of the trap depth on the CSBM factors (architecture dummies relative to HGB-Graph).}
\label{tab:factorial}

\begin{tabular}{lr}
\toprule
Term & coefficient \\
\midrule
\input{tables/tab_factorial.tex}
\bottomrule
\end{tabular}
\end{table}

\paragraph{Adversarial training at matched compute.} \cref{tab:atcompute} trains CA-AT for a fraction of the standard number of epochs, including the number matched to the wall time of \cars{}-D, and evaluates it with the full suite.

\begin{table*}[ht]
\centering\scriptsize
\setlength{\tabcolsep}{12pt}

\caption{R-GCN + \cars{}-D (wall time) and CA-AT by number of epochs (evaluated worst-case AUC, P1, $B{=}2$, seed 0).}
\label{tab:atcompute}

\begin{tabular}{ll rr}
\toprule
Dataset & Cell & R-GCN + \cars{}-D & CA-AT (epochs: AUC) \\
\midrule
\input{tables/tab_atcompute.tex}
\bottomrule
\end{tabular}
\end{table*}

\begin{table}[ht]
\centering\scriptsize
\setlength{\tabcolsep}{3.5pt}

\caption{Synthetic CSBM, $\cid{=}0.5$: evaluated worst-case AUC (three seeds). Bottom: the attacks on RLR, whose exact worst case is known.}
\label{tab:syn}

\begin{tabular}{l *{6}{r}}
\toprule
Model & clean & $B{=}1$ & $2$ & $4$ & $8$ & $16$ \\
\midrule
\input{tables/tab_syn.tex}
\midrule
Attack on RLR & query & transfer & gradient & exact & & \\
\midrule
\input{tables/tab_gap.tex}
\bottomrule
\end{tabular}
\end{table}

\paragraph{Ablation of \cars{}-D.} \cref{tab:ablation} reports, per detector family and dataset, the mean margin over the graph-free counterpart of the plug-in subset (\cars{}), of the plug-in gated proposal, and of the configuration kept by the validation check (\cars{}-D). The proposals win where the plug-in is accurate (Tolokers, R-GCN; the synthetic graph) and lose where it is not (Amazon and S-FFSD at P1); the check keeps the better configuration in all but a few runs, whose cost is within validation noise.

\begin{table}[ht]
\centering\scriptsize
\caption{Mean margin (AUC) over the graph-free counterpart across the three cells: plug-in subset (S), plug-in gated proposal (G), and configuration kept by the validation check (D).}
\label{tab:ablation}
\setlength{\tabcolsep}{2pt}
\begin{tabular}{l ccc ccc ccc ccc ccc}
\toprule
& \multicolumn{3}{c}{YelpChi} & \multicolumn{3}{c}{Amazon} & \multicolumn{3}{c}{FDCompCN} & \multicolumn{3}{c}{S-FFSD} & \multicolumn{3}{c}{Tolokers}\\
\cmidrule(lr){2-4}\cmidrule(lr){5-7}\cmidrule(lr){8-10}\cmidrule(lr){11-13}\cmidrule(lr){14-16}
Family & S & G & D & S & G & D & S & G & D & S & G & D & S & G & D\\
\midrule
\input{tables/tab_ablation.tex}
\bottomrule
\end{tabular}
\end{table}

\paragraph{Temporal split.} On S-FFSD, models are trained on the earliest 40\% of labeled transactions, validated on the next 20\%, and tested on the last 40\%, without the time stamp as a feature (three seeds, averaged).

\begin{table}[ht]
\centering\small
\caption{S-FFSD, temporal split: clean and evaluated worst-case AUC (mean over three seeds). Models whose training depends on the cost cell (CA-HGB and the \cars{} variants, +C and +D) report the clean AUC of their P1, $B{=}2$ instance, as in \cref{tab:main}.}
\label{tab:temporal}
\begin{tabular}{lcccc}
\toprule
Model & clean & P1, $B{=}2$ & P1, $B{=}20$ & P2, $B{=}20$\\
\midrule
\input{tables/tab_temporal.tex}
\bottomrule
\end{tabular}
\end{table}

\paragraph{Node-level exposure.} For a one-layer relational network, \cref{tab:exposure} compares the node-level bound of \cref{prop:gnn}, computed with each target's degree and its actual candidate endpoints, with the measured score drop of the best single edit per unit cost (seed 0, up to 400 targets). The bound is never violated (\EXPviol{} violations out of \EXPn{} node--relation--action triples); for injections, unlike the dataset-level bound with mean degrees, it ranks nodes by their measured exposure (Spearman up to \EXPrhoMax{}), whereas for deletions, whose damage depends on which accomplices exist, it does not.

\begin{table}[ht]
\centering\small
\caption{Node-level exposure bounds: violations, median ratio of measured damage to bound, and Spearman correlation between bound and measured damage across nodes and across relations.}
\label{tab:exposure}
\setlength{\tabcolsep}{3pt}
\begin{tabular}{ll c c c cc}
\toprule
Dataset & action & $n$ & viol. & ratio & $\rho_{\text{nodes}}$ & $\rho_{\text{rel.}}$\\
\midrule
\input{tables/tab_exposure.tex}
\bottomrule
\end{tabular}
\end{table}

\begin{figure}[ht]
\centering
\includegraphics[width=0.9\columnwidth]{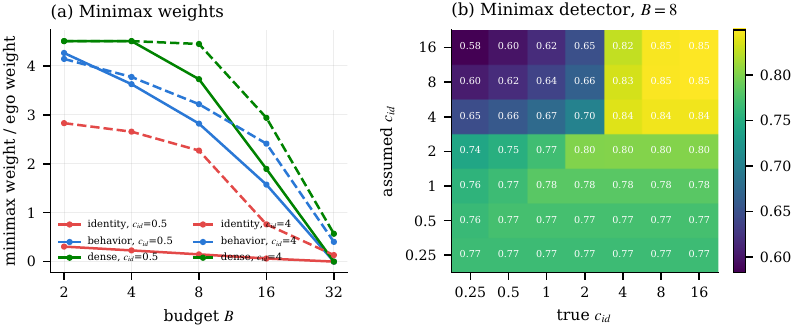}
\caption{(a) Minimax weights relative to the ego weight as the budget grows. (b) Worst-case AUC at $B{=}8$ under the true identity cost (columns) of the minimax detector built for an assumed cost (rows).}
\label{fig:theoryapp}
\end{figure}

\paragraph{Thresholds and weights in the CSBM.} \cref{tab:thresholds} lists, for every identity cost, the numerical trap threshold, the lower envelope of the trap-attack lines, the exposure lower bound, the linear-response threshold, and the abandonment threshold with its bound. The best response is computed by multi-start SLSQP; a dense random search of 20{,}000 feasible attacks per instance followed by local polishing never improves on it by more than $\TGlobalGap$ in AUC across 48 instances. \cref{fig:theoryapp} shows the minimax weights, which are clipped as \cref{thm:abandon}(b) predicts, and the cost misspecification of the minimax detector: rows never decrease for true costs above the assumed one, as \cref{cor:robust} implies.

\begin{table}[ht]
\centering\scriptsize
\setlength{\tabcolsep}{3.5pt}

\caption{CSBM thresholds as a function of the identity cost $\cid$.}
\label{tab:thresholds}

\input{tables/tab_thresholds.tex}
\end{table}

\paragraph{Fidelity of the plug-in.} For every relation subset and single-relation gate, \cref{tab:fidelity} compares the plug-in worst case of the linear detector with the exact worst case of the gated RLR retrained on the configuration (seed 0): rank correlation with the test worst case, rank correlation between validation and test worst cases, the test regret of picking the plug-in's best configuration or the validation's best, and the mean overestimate of the plug-in.

\begin{table}[ht]
\centering\scriptsize
\caption{Fidelity of the plug-in model across configurations.}
\label{tab:fidelity}
\setlength{\tabcolsep}{2.5pt}
\begin{tabular}{llcccccc}
\toprule
Dataset & Cell & configs & $\rho$(plug, test) & $\rho$(val, test) & regret plug & regret val & overest.\\
\midrule
\input{tables/tab_fidelity.tex}
\bottomrule
\end{tabular}
\end{table}

\paragraph{Statistical reliability and operating points.} \cref{tab:significance} reports, per dataset and cell, the mean paired difference over seeds between our detectors and their graph-free counterparts with its 95\% $t$ interval, the number of seeds at or above the counterpart (within 0.002), and the average precision and true-positive rate at 1\% false-positive rate under attack.

\begin{table*}[ht]
\centering\scriptsize
\caption{Paired comparison with the graph-free counterpart over seeds, and operating metrics under attack (ours / counterpart).}
\label{tab:significance}
\begin{tabular}{lllcccccc}
\toprule
Dataset & Cell & Model & seeds & diff. & 95\% CI & at/above & AP & TPR@1\%FPR\\
\midrule
\input{tables/tab_significance.tex}
\bottomrule
\end{tabular}
\end{table*}

\paragraph{Compute.} \cref{tab:cost} compares the median time of one \cars{}-D planning, with the training and validation check of the three detector families, with the training of one CA-AT model at the same cell (one CPU core).

\begin{table}[ht]
\centering\small
\caption{Median wall-clock time (seconds) of \cars{}-D (three families) and of one CA-AT model.}
\label{tab:cost}
\begin{tabular}{lccc}
\toprule
Dataset & \cars{}-D & CA-AT & ratio\\
\midrule
\input{tables/tab_cost.tex}
\bottomrule
\end{tabular}
\end{table}

\begin{figure*}[ht]
\centering
\includegraphics[width=\textwidth]{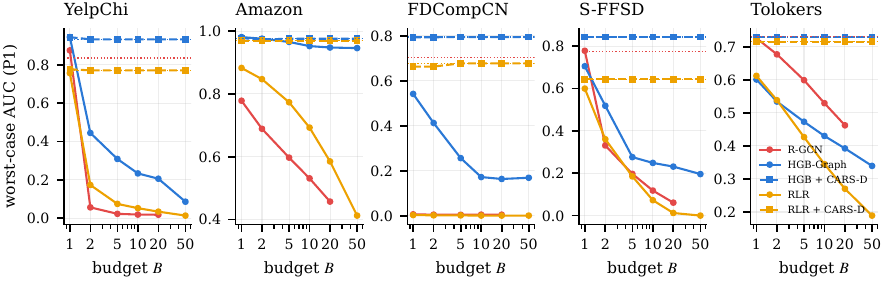}
\caption{Evaluated worst-case AUC at P1 as a function of the per-node budget (seed 0). Solid: clean-trained; dashed: \cars{}-D planned at each budget; dotted: graph-free counterparts (HGB-X, MLP, LR). Gradient attacks are run up to $B{=}20$, so R-GCN is shown up to that budget.}
\label{fig:sweep}
\end{figure*}

\paragraph{Budget sweeps (full suite).} \cref{fig:sweep} traces the evaluated worst case of clean-trained models and of RLR and HGB with \cars{}-D, planned at each budget, from $B{=}1$ to $50$ at P1. With the full suite, the clean-trained curves cross the graph-free lines at budgets of one to two edits on every graph, except trees on Amazon, which cross at $B{=}5$; the \cars{}-D curves stay within 0.002 of them except RLR + \cars{}-D on FDCompCN at $B\le2$ (0.014 below, within the validation radius).

\paragraph{Random cost vectors (full suite).} \cref{tab:randcost} draws 24 cost vectors log-uniformly in $[0.25,16]^{2R}$ per dataset ($B{=}5$, seed 0) and compares the observed trap of RLR (exact attack) and HGB-Graph with the plug-in prediction (the plug-in worst case of the full linear detector below that of the graph-free one).

\begin{table}[ht]
\centering\small
\caption{Random cost vectors: number of traps observed, accuracy of the plug-in trap prediction, and rank correlation between the plug-in margin and the observed margin of RLR.}
\label{tab:randcost}
\setlength{\tabcolsep}{3pt}
\begin{tabular}{l c cc cc c}
\toprule
Dataset & vectors & traps RLR & acc. & traps HGB & acc. & Spearman\\
\midrule
\input{tables/tab_randcost.tex}
\bottomrule
\end{tabular}
\end{table}

\paragraph{Further threat variations under the full suite.} The following checks were rerun with the exhaustive-greedy attack in evaluation and in the validation check of \cars{}-D (seed 0; YelpChi, Tolokers, FDCompCN; entity-level attacks on YelpChi with three seeds).

\paragraph{Attacker knowledge.} \cref{tab:knowledge} compares the white-box attacker with labels (default), the white-box attacker without the label oracle, and a black-box attacker who transfers edits from a surrogate R-GCN trained on its own labeled sample (seed 0).

\begin{table}[ht]
\centering\scriptsize
\caption{Evaluated worst-case AUC under three attacker knowledge levels (seed 0).}
\label{tab:knowledge}
\setlength{\tabcolsep}{2.5pt}
\begin{tabular}{lll ccc}
\toprule
Dataset & Cell & Model & labels & no labels & black box\\
\midrule
\input{tables/tab_knowledge.tex}
\bottomrule
\end{tabular}
\end{table}

\paragraph{Entity-level attacks.} \cref{tab:entity} attacks YelpChi detectors through entities: a review is reassigned to another (aged) account, a fresh account, another rating group, or another posting time, and every induced edge is rebuilt (three seeds). \cars{}-D is planned with the edge-level profile that matches the entity profile (Q1 with P1, Q2 with P2). Under Q1, entity actions trap R-GCN and HGB-Graph at both budgets and RLR at $B{=}5$, as edge actions do under P1; each \cars{}-D variant is within 0.002 of its graph-free counterpart under Q1 (LR, not in the table, scores .771 on YelpChi; MLP .837; HGB-X .935) and above it under Q2.

\begin{table}[ht]
\centering\small
\caption{YelpChi, entity-level attacks (evaluated worst-case AUC, 3 seeds).}
\label{tab:entity}
\begin{tabular}{l c ccc}
\toprule
Model & clean & Q1, $B{=}2$ & Q1, $B{=}5$ & Q2, $B{=}5$\\
\midrule
\input{tables/tab_entity.tex}
\bottomrule
\end{tabular}
\end{table}

\paragraph{Persistent camouflage.} Before training, every fraud node (training, validation, and test) applies the exact attack of a clean RLR surrogate at P1, $B{=}2$, and the edits stay in the graph; detectors and \cars{} are then built on the poisoned graph, and test fraud nodes attack again with the same budget (seed 0). When the camouflage edges concentrate on the surrogate's most benign-looking nodes (c), they leave a signature that graph models detect almost perfectly; when they are diversified over benign nodes (d), the trap persists. In both cases, attacked fraud nodes in the training set act as adversarial augmentation.

\begin{table*}[ht]
\centering\scriptsize
\setlength{\tabcolsep}{3pt}

\caption{Persistent camouflage: evaluated worst-case AUC after poisoning (c: concentrated, d: diversified endpoints).}
\label{tab:poison}

\begin{tabular}{\POISONcols}
\toprule
\input{tables/tab_poison_head.tex}
\midrule
\input{tables/tab_poison.tex}
\bottomrule
\end{tabular}
\end{table*}

\paragraph{Sybil accounts.} The attacker creates 1{,}000 fake accounts that copy the features of random labeled benign nodes with small noise and connects them to targets; each Sybil edge costs $c_s$ on every relation, and deletions keep their P1 costs. Detectors are trained without the Sybils (seed 0).

\begin{table*}[ht]
\centering\scriptsize
\setlength{\tabcolsep}{3pt}

\caption{Sybil attacks: evaluated worst-case AUC for Sybil edge cost $c_s$ and budget $B$.}
\label{tab:sybil}

\begin{tabular}{\SYBILcols}
\toprule
\input{tables/tab_sybil_head.tex}
\midrule
\input{tables/tab_sybil.tex}
\bottomrule
\end{tabular}
\end{table*}

\paragraph{Feature and structure attacks.} Fraudsters may also perturb their own standardized features within an $\ell_\infty$ ball of radius $\epsilon$; graph-free detectors face this feature attack, and graph detectors face it followed by the structural suite at P1, $B{=}2$ on the perturbed graph (seed 0). Feature attacks hurt every detector; they hurt graph-free ones more when the structure is untouched, which shows that relations do buy robustness against feature manipulation as long as they are expensive to edit.

\begin{table}[ht]
\centering\scriptsize
\caption{Evaluated worst-case AUC under feature attacks of radius $\epsilon\in\{0,0.02,0.05,0.1\}$, combined with structural attacks for graph detectors.}
\label{tab:features}
\begin{tabular}{llcccc}
\toprule
Dataset & Model & $0$ & $0.02$ & $0.05$ & $0.1$\\
\midrule
\input{tables/tab_features.tex}
\bottomrule
\end{tabular}
\end{table}

\paragraph{Label scarcity.} \cref{tab:scarcity} trains with 5\%, 10\%, and 20\% of the evaluated nodes (validation 20\%, test the rest; seed 0).

\begin{table*}[ht]
\centering\scriptsize
\caption{Label scarcity: clean AUC of the graph-free baselines and evaluated worst-case AUC at P1, $B{=}2$ / P2, $B{=}20$.}
\label{tab:scarcity}
\setlength{\tabcolsep}{2.5pt}
\begin{tabular}{lc cc cc cc cc cc}
\toprule
& & \multicolumn{2}{c}{graph-free (clean)} & \multicolumn{2}{c}{R-GCN} & \multicolumn{2}{c}{HGB-Graph} & \multicolumn{2}{c}{R-GCN + \cars{}-D} & \multicolumn{2}{c}{HGB + \cars{}-D}\\
\cmidrule(lr){3-4}\cmidrule(lr){5-6}\cmidrule(lr){7-8}\cmidrule(lr){9-10}\cmidrule(lr){11-12}
Dataset & labels & MLP & HGB-X & P1,2 & P2,20 & P1,2 & P2,20 & P1,2 & P2,20 & P1,2 & P2,20\\
\midrule
\input{tables/tab_scarcity.tex}
\bottomrule
\end{tabular}
\end{table*}

%% file: tables/tab_main_full.tex
MLP & .836 & .836 & .836 & .836 & .975 & .975 & .975 & .975 & .715 & .715 & .715 & .715 & .788 & .788 & .788 & .788 & .740 & .740 & .740 & .740\\
HGB-X & .935 & .935 & .935 & .935 & .977 & .977 & .977 & .977 & .789 & .789 & .789 & .789 & .845 & .845 & .845 & .845 & .734 & .734 & .734 & .734\\
LR & .771 & .771 & .771 & .771 & .969 & .969 & .969 & .969 & .680 & .680 & .680 & .680 & .644 & .644 & .644 & .644 & .722 & .722 & .722 & .722\\
\midrule
\multicolumn{21}{l}{\emph{Clean-trained, graph}}\\
RLR & .798 & \underline{.181} & \underline{.033} & \underline{.546} & .971 & \underline{.865} & \underline{.620} & \underline{.674} & .656 & .001 & .001 & .001 & .660 & \underline{.355} & \underline{.005} & \underline{.061} & .768 & \underline{.529} & \underline{.244} & \underline{.529}\\
R-GCN & .915 & \underline{.043} & \underline{.009} & \underline{.639} & .976 & .739 & .588 & .594 & .631 & .003 & .002 & .002 & .878 & \underline{.325} & \underline{.043} & \underline{.143} & .811 & \underline{.649} & \underline{.416} & \underline{.648}\\
HGB-Graph & .960 & \underline{.442} & \underline{.200} & \underline{.924} & .986 & \underline{.971} & \underline{.945} & \underline{.951} & .785 & .473 & .245 & .250 & .869 & \underline{.551} & \underline{.258} & \underline{.385} & .818 & \underline{.524} & \underline{.380} & \underline{.524}\\
CARE-GNN & .846 & \underline{.773} & \underline{.720} & \underline{.808} & .974 & .967 & .961 & .963 & .697 & .026 & .015 & .014 & .701 & .654 & .593 & .624 & .747 & \underline{.705} & \underline{.622} & \underline{.705}\\
GNNGuard-style & .917 & \underline{.169} & \underline{.080} & .850 & .976 & .912 & .867 & .880 & .652 & .014 & .011 & .011 & .868 & \underline{.567} & \underline{.294} & \underline{.529} & .808 & \underline{.637} & \underline{.459} & \underline{.636}\\
BWGNN-style & .902 & \underline{.593} & \underline{.425} & .868 & .978 & \underline{.967} & \underline{.947} & \underline{.946} & .688 & .093 & .082 & .082 & .867 & \underline{.715} & \underline{.402} & \underline{.654} & .831 & \underline{.729} & \underline{.661} & \underline{.726}\\
\midrule
\multicolumn{21}{l}{\emph{Adversarially trained}}\\
CA-AT (R-GCN) & .866 & \underline{.679} & \underline{.591} & \underline{.834} & .976 & .892 & .877 & .881 & .644 & .229 & .166 & .120 & .853 & \underline{.686} & .400 & \underline{.628} & .819 & \underline{.734} & \underline{.603} & \underline{.734}\\
AT-RLR (exact) & .773 & \underline{.366} & .172 & \underline{.597} & .967 & .935 & .887 & .901 & .619 & .014 & .012 & .012 & .645 & .562 & .476 & \underline{.529} & .737 & \underline{.698} & \underline{.614} & \underline{.697}\\
CA-HGB & .962 & \underline{.919} & \underline{.710} & .944 & .987 & .979 & \underline{.960} & \underline{.964} & .784 & .748 & .521 & .540 & .864 & \underline{.702} & \underline{.532} & \underline{.639} & .818 & \underline{.680} & \underline{.469} & \underline{.680}\\
\midrule
\multicolumn{21}{l}{\emph{Ours}}\\
RLR + \cars{} & .771 & \textbf{.771} & \textbf{.771} & \textbf{.791} & .972 & \underline{.951} & \textbf{.969} & \textbf{.969} & .680 & .674 & \textbf{.680} & \textbf{.680} & .644 & .504 & \textbf{.644} & \textbf{.664} & .722 & \textbf{.722} & \textbf{.722} & \textbf{.722}\\
R-GCN + \cars{} & .836 & \textbf{.836} & \textbf{.836} & \textbf{.897} & .976 & .943 & \textbf{.975} & \textbf{.975} & .671 & .669 & \textbf{.715} & \textbf{.715} & .843 & \underline{.328} & \textbf{.788} & \textbf{.858} & .740 & \textbf{.740} & \textbf{.740} & \textbf{.740}\\
HGB + \cars{} & .935 & \textbf{.935} & \textbf{.935} & \textbf{.952} & .982 & \textbf{.976} & \textbf{.977} & \textbf{.977} & .788 & .787 & \textbf{.789} & \textbf{.789} & .852 & \underline{.449} & \textbf{.845} & \textbf{.863} & .734 & \textbf{.734} & \textbf{.734} & \textbf{.734}\\
RLR + \cars{}-D & .771 & \textbf{.771} & \textbf{.771} & \textbf{.792} & .969 & \textbf{.969} & \textbf{.969} & \textbf{.969} & .682 & \textbf{.678} & \textbf{.680} & \textbf{.680} & .644 & \textbf{.644} & \textbf{.644} & \textbf{.664} & .722 & \textbf{.722} & \textbf{.722} & \textbf{.722}\\
R-GCN + \cars{}-D & .836 & \textbf{.836} & \textbf{.836} & \textbf{.897} & .975 & \textbf{.975} & \textbf{.975} & \textbf{.975} & .715 & \textbf{.715} & \textbf{.715} & \textbf{.715} & .788 & \textbf{.788} & \textbf{.788} & \textbf{.858} & .746 & \textbf{.741} & \textbf{.740} & \textbf{.741}\\
HGB + \cars{}-D & .935 & \textbf{.935} & \textbf{.935} & \textbf{.952} & .979 & \textbf{.975} & \textbf{.977} & \textbf{.977} & .790 & \textbf{.789} & \textbf{.789} & \textbf{.789} & .845 & \textbf{.845} & \textbf{.845} & \textbf{.863} & .734 & \textbf{.734} & \textbf{.734} & \textbf{.734}\\
\cars{}+AT (R-GCN) & .837 & \textbf{.837} & \textbf{.837} & \textbf{.896} & .973 & \textbf{.973} & \textbf{.973} & \textbf{.973} & .715 & \textbf{.715} & \textbf{.715} & \textbf{.715} & .782 & .782 & .782 & \textbf{.856} & .746 & \textbf{.741} & \textbf{.740} & \textbf{.741}\\
\cars{}+AT (HGB) & .935 & \textbf{.935} & \textbf{.935} & \textbf{.952} & .979 & \textbf{.976} & \textbf{.977} & \textbf{.977} & .790 & \textbf{.789} & \textbf{.789} & \textbf{.789} & .845 & \textbf{.845} & \textbf{.845} & \textbf{.863} & .734 & \textbf{.734} & \textbf{.734} & \textbf{.734}\\

%% file: tables/tab_bracket.tex
YelpChi & P1, $B{=}2$ & R-GCN & .915 & .112 & .042 & .043 & .043 & .000 & informative\\
YelpChi & P1, $B{=}2$ & HGB-Graph & .960 & .659 & .439 & .442 & .438 & .070 & loose\\
YelpChi & P1, $B{=}2$ & R-GCN + \cars{}-D & .836 & .836 & -- & .836 & .836 & .836 & tight\\
YelpChi & P1, $B{=}2$ & HGB + \cars{}-D & .935 & .935 & .935 & .935 & .935 & .935 & tight\\
YelpChi & P2, $B{=}20$ & R-GCN & .915 & .798 & .632 & .639 & .632 & .000 & loose\\
YelpChi & P2, $B{=}20$ & HGB-Graph & .960 & .944 & .927 & .924 & .922 & .567 & loose\\
YelpChi & P2, $B{=}20$ & R-GCN + \cars{}-D & .897 & .897 & .897 & .897 & .897 & .836 & loose\\
YelpChi & P2, $B{=}20$ & HGB + \cars{}-D & .952 & .952 & .952 & .952 & .952 & .952 & tight\\
\midrule
Amazon & P1, $B{=}2$ & R-GCN & .976 & .879 & .743 & .739 & .742 & .024 & loose\\
Amazon & P1, $B{=}2$ & HGB-Graph & .986 & .980 & .971 & .971 & .971 & .914 & loose\\
Amazon & P1, $B{=}2$ & R-GCN + \cars{}-D & .975 & .975 & -- & .975 & .975 & .975 & tight\\
Amazon & P1, $B{=}2$ & HGB + \cars{}-D & .979 & .977 & .975 & .975 & .975 & .962 & informative\\
Amazon & P2, $B{=}20$ & R-GCN & .976 & .786 & .598 & .594 & .591 & .003 & loose\\
Amazon & P2, $B{=}20$ & HGB-Graph & .986 & .973 & .950 & .951 & .949 & .840 & loose\\
Amazon & P2, $B{=}20$ & R-GCN + \cars{}-D & .975 & .975 & -- & .975 & .975 & .975 & tight\\
Amazon & P2, $B{=}20$ & HGB + \cars{}-D & .977 & .977 & .977 & .977 & .977 & .977 & tight\\
\midrule
FDCompCN & P1, $B{=}2$ & R-GCN & .631 & .006 & .003 & .003 & .002 & .000 & tight\\
FDCompCN & P1, $B{=}2$ & HGB-Graph & .785 & .621 & .469 & .473 & .469 & .387 & loose\\
FDCompCN & P1, $B{=}2$ & R-GCN + \cars{}-D & .715 & .715 & -- & .715 & .715 & .715 & tight\\
FDCompCN & P1, $B{=}2$ & HGB + \cars{}-D & .790 & .789 & .789 & .789 & .789 & .788 & tight\\
FDCompCN & P2, $B{=}20$ & R-GCN & .631 & .002 & .002 & .002 & .001 & .000 & tight\\
FDCompCN & P2, $B{=}20$ & HGB-Graph & .785 & .549 & .239 & .250 & .239 & .142 & loose\\
FDCompCN & P2, $B{=}20$ & R-GCN + \cars{}-D & .715 & .715 & -- & .715 & .715 & .715 & tight\\
FDCompCN & P2, $B{=}20$ & HGB + \cars{}-D & .789 & .789 & .789 & .789 & .789 & .789 & tight\\
\midrule
S-FFSD & P1, $B{=}2$ & R-GCN & .878 & .445 & .361 & .325 & .326 & .000 & loose\\
S-FFSD & P1, $B{=}2$ & HGB-Graph & .869 & .677 & .555 & .551 & .547 & .134 & loose\\
S-FFSD & P1, $B{=}2$ & R-GCN + \cars{}-D & .788 & .788 & -- & .788 & .788 & .788 & tight\\
S-FFSD & P1, $B{=}2$ & HGB + \cars{}-D & .845 & .845 & .845 & .845 & .845 & .845 & tight\\
S-FFSD & P2, $B{=}20$ & R-GCN & .878 & .338 & .141 & .143 & .134 & .000 & loose\\
S-FFSD & P2, $B{=}20$ & HGB-Graph & .869 & .414 & .466 & .385 & .366 & .058 & loose\\
S-FFSD & P2, $B{=}20$ & R-GCN + \cars{}-D & .858 & .858 & .858 & .858 & .858 & .811 & informative\\
S-FFSD & P2, $B{=}20$ & HGB + \cars{}-D & .863 & .863 & .863 & .863 & .863 & .863 & tight\\
\midrule
Tolokers & P1, $B{=}2$ & R-GCN & .811 & .707 & .646 & .649 & .644 & .349 & loose\\
Tolokers & P1, $B{=}2$ & HGB-Graph & .818 & .632 & .529 & .524 & .521 & .222 & loose\\
Tolokers & P1, $B{=}2$ & R-GCN + \cars{}-D & .746 & .744 & .735 & .741 & .741 & .727 & informative\\
Tolokers & P1, $B{=}2$ & HGB + \cars{}-D & .734 & .734 & .734 & .734 & .734 & .734 & tight\\
Tolokers & P2, $B{=}20$ & R-GCN & .811 & .693 & .646 & .648 & .644 & .349 & loose\\
Tolokers & P2, $B{=}20$ & HGB-Graph & .818 & .632 & .529 & .524 & .521 & .222 & loose\\
Tolokers & P2, $B{=}20$ & R-GCN + \cars{}-D & .746 & .744 & .735 & .741 & .741 & .727 & informative\\
Tolokers & P2, $B{=}20$ & HGB + \cars{}-D & .734 & .734 & .734 & .734 & .734 & .734 & tight\\

%% file: tables/tab_regret.tex
Amazon & HGB-\cars{}-D & 150 & .984 & .979 & .975 & .974 & .944 & 71\% & 29\%\\
Amazon & HGB-Graph & 150 & .988 & .984 & .978 & .977 & .934 & 92\% & 68\%\\
Amazon & R-GCN & 150 & .969 & .857 & .711 & .616 & .016 & 99\% & 97\%\\
Amazon & R-GCN-\cars{}-Dp & 150 & .976 & .973 & .970 & .970 & .894 & 78\% & 27\%\\
FDCompCN & HGB-\cars{}-D & 150 & .783 & .783 & .784 & .783 & .777 & 0\% & 3\%\\
FDCompCN & HGB-Graph & 150 & .776 & .571 & .387 & .384 & .335 & 99\% & 45\%\\
FDCompCN & R-GCN & 150 & .615 & .013 & .005 & .003 & .000 & 89\% & 93\%\\
FDCompCN & R-GCN-\cars{}-Dp & 150 & .637 & .635 & .635 & .635 & .630 & 0\% & 11\%\\
S-FFSD & HGB-\cars{}-Dp & 150 & .881 & .657 & .401 & .383 & .124 & 97\% & 27\%\\
S-FFSD & HGB-Graph & 150 & .882 & .629 & .516 & .433 & .084 & 73\% & 42\%\\
S-FFSD & R-GCN & 150 & .876 & .400 & .338 & .250 & .000 & 70\% & 77\%\\
S-FFSD & R-GCN-\cars{}-Dp & 150 & .864 & .684 & .597 & .542 & .000 & 84\% & 87\%\\
Tolokers & HGB-\cars{}-Dp & 150 & .744 & .713 & .680 & .676 & .573 & 39\% & 15\%\\
Tolokers & HGB-Graph & 150 & .804 & .659 & .508 & .491 & .213 & 95\% & 51\%\\
Tolokers & R-GCN & 150 & .784 & .675 & .639 & .631 & .355 & 94\% & 69\%\\
Tolokers & R-GCN-\cars{}-D & 150 & .741 & .725 & .715 & .714 & .650 & 36\% & 33\%\\
YelpChi & HGB-\cars{}-Dp & 150 & .942 & .873 & .837 & .834 & .802 & 37\% & 27\%\\
YelpChi & HGB-Graph & 150 & .947 & .685 & .455 & .404 & .100 & 99\% & 97\%\\
YelpChi & R-GCN & 150 & .907 & .098 & .048 & .040 & .000 & 95\% & 77\%\\
YelpChi & R-GCN-\cars{}-Dp & 150 & .870 & .755 & .730 & .723 & .628 & 38\% & 16\%\\

%% file: tables/tab_select.tex
YelpChi & 45 & 67\% & 29\% & 4\% & .046 & .092 & .019 & $-$.001\\
Amazon & 45 & 96\% & 0\% & 4\% & .127 & .254 & .042 & .002\\
FDCompCN & 45 & 87\% & 7\% & 7\% & .155 & .309 & .081 & $-$.028\\
S-FFSD & 45 & 67\% & 33\% & 0\% & .053 & .105 & .048 & .001\\
Tolokers & 45 & 96\% & 0\% & 4\% & .077 & .155 & .027 & $-$.007\\

%% file: tables/tab_valsel.tex
Amazon & HGB & P1, $B{=}2$ & .973 & .974 & .976 & .976 & .973 & .973 & .977 & 3 / 8\\
Amazon & HGB & P2, $B{=}20$ & .976 & .976 & .976 & .976 & .976 & .976 & .976 & 3 / 8\\
Amazon & R-GCN & P1, $B{=}2$ & .897 & .971 & .971 & .971 & .971 & .971 & .972 & 3 / 8\\
Amazon & R-GCN & P2, $B{=}20$ & .971 & .971 & .971 & .971 & .971 & .971 & .971 & 3 / 8\\
Amazon & RLR & P1, $B{=}2$ & .930 & .969 & .969 & .969 & .967 & .969 & .969 & 3 / 8\\
Amazon & RLR & P2, $B{=}20$ & .969 & .969 & .969 & .969 & .969 & .969 & .969 & 3 / 8\\
FDCompCN & HGB & P1, $B{=}2$ & .781 & .781 & .781 & .781 & .780 & .780 & .784 & 3 / 8\\
FDCompCN & HGB & P2, $B{=}20$ & .781 & .781 & .781 & .781 & .781 & .781 & .781 & 3 / 8\\
FDCompCN & R-GCN & P1, $B{=}2$ & .666 & .718 & .718 & .716 & .716 & .718 & .722 & 3 / 8\\
FDCompCN & R-GCN & P2, $B{=}20$ & .718 & .718 & .718 & .718 & .718 & .718 & .718 & 3 / 8\\
FDCompCN & RLR & P1, $B{=}2$ & .668 & .674 & .679 & .679 & .677 & .674 & .680 & 3 / 8\\
FDCompCN & RLR & P2, $B{=}20$ & .679 & .679 & .679 & .679 & .679 & .679 & .679 & 3 / 8\\
S-FFSD & HGB & P1, $B{=}2$ & .470 & .842 & .842 & .842 & .842 & .842 & -- & 3 / 16\\
S-FFSD & HGB & P2, $B{=}20$ & .860 & .860 & .842 & .842 & .864 & .860 & -- & 3 / 16\\
S-FFSD & RLR & P1, $B{=}2$ & .546 & .643 & .643 & .643 & .643 & .643 & -- & 3 / 16\\
S-FFSD & RLR & P2, $B{=}20$ & .661 & .661 & .643 & .643 & .699 & .661 & -- & 3 / 16\\
Tolokers & HGB & P1, $B{=}2$ & .733 & .733 & .733 & .733 & .733 & .733 & .733 & 3 / 2\\
Tolokers & HGB & P2, $B{=}20$ & .733 & .733 & .733 & .733 & .733 & .733 & .733 & 3 / 2\\
Tolokers & R-GCN & P1, $B{=}2$ & .737 & .739 & .737 & .737 & .745 & .737 & .745 & 3 / 2\\
Tolokers & R-GCN & P2, $B{=}20$ & .737 & .739 & .737 & .741 & .745 & .737 & .745 & 3 / 2\\
Tolokers & RLR & P1, $B{=}2$ & .723 & .723 & .723 & .723 & .723 & .723 & .724 & 3 / 2\\
Tolokers & RLR & P2, $B{=}20$ & .723 & .723 & .723 & .723 & .723 & .723 & .724 & 3 / 2\\
YelpChi & HGB & P1, $B{=}2$ & .933 & .933 & .933 & .933 & .935 & .933 & -- & 3 / 8\\
YelpChi & HGB & P2, $B{=}20$ & .948 & .948 & .933 & .933 & .950 & .948 & -- & 3 / 8\\
YelpChi & RLR & P1, $B{=}2$ & .771 & .771 & .771 & .773 & .792 & .771 & -- & 3 / 8\\
YelpChi & RLR & P2, $B{=}20$ & .794 & .794 & .771 & .771 & .795 & .794 & -- & 3 / 8\\

%% file: tables/tab_groups.tex
YelpChi & 1--2 & 8 & -- & .017 & .717 & -- & .407 & .834\\
YelpChi & 3--9 & 47 & -- & .025 & .816 & -- & .556 & .934\\
YelpChi & 10+ & 2613 & -- & .044 & .837 & -- & .440 & .935\\
Amazon & 10+ & 327 & -- & .742 & .975 & -- & .971 & .975\\
FDCompCN & 0 & 130 & -- & .002 & .763 & -- & .505 & .823\\
FDCompCN & 1--2 & 69 & -- & .003 & .662 & -- & .434 & .749\\
FDCompCN & 3--9 & 21 & -- & .005 & .573 & -- & .416 & .708\\
S-FFSD & 10+ & 2101 & -- & .326 & .788 & -- & .551 & .845\\
Tolokers & 1--2 & 82 & -- & .103 & .734 & -- & .145 & .743\\
Tolokers & 3--9 & 136 & -- & .420 & .739 & -- & .263 & .751\\
Tolokers & 10+ & 807 & -- & .743 & .742 & -- & .607 & .730\\

%% file: tables/tab_metrics.tex
YelpChi & MLP & .836 & .525 & .178 & .397\\
YelpChi & HGB-X & .935 & .783 & .458 & .706\\
YelpChi & R-GCN & .043 & .077 & .001 & .001\\
YelpChi & HGB-Graph & .442 & .120 & .003 & .017\\
YelpChi & R-GCN-\cars{}-D & .836 & .525 & .178 & .397\\
YelpChi & HGB-\cars{}-D & .935 & .783 & .458 & .706\\
YelpChi & CA-HGB & .919 & .723 & .346 & .628\\
YelpChi & \cars{}+AT-HGB & .935 & .783 & .458 & .706\\
Amazon & MLP & .975 & .889 & .818 & .902\\
Amazon & HGB-X & .977 & .906 & .822 & .909\\
Amazon & R-GCN & .739 & .566 & .426 & .550\\
Amazon & HGB-Graph & .971 & .889 & .784 & .886\\
Amazon & R-GCN-\cars{}-D & .975 & .889 & .818 & .902\\
Amazon & HGB-\cars{}-D & .975 & .902 & .810 & .906\\
Amazon & CA-HGB & .979 & .897 & .812 & .908\\
Amazon & \cars{}+AT-HGB & .976 & .901 & .812 & .905\\
FDCompCN & MLP & .715 & .282 & .074 & .250\\
FDCompCN & HGB-X & .789 & .413 & .167 & .416\\
FDCompCN & R-GCN & .003 & .055 & .000 & .000\\
FDCompCN & HGB-Graph & .473 & .121 & .009 & .075\\
FDCompCN & R-GCN-\cars{}-D & .715 & .282 & .074 & .250\\
FDCompCN & HGB-\cars{}-D & .789 & .412 & .169 & .410\\
FDCompCN & CA-HGB & .748 & .346 & .127 & .341\\
FDCompCN & \cars{}+AT-HGB & .789 & .412 & .169 & .410\\
S-FFSD & MLP & .788 & .582 & .314 & .417\\
S-FFSD & HGB-X & .845 & .678 & .430 & .504\\
S-FFSD & R-GCN & .325 & .124 & .001 & .004\\
S-FFSD & HGB-Graph & .551 & .296 & .131 & .195\\
S-FFSD & R-GCN-\cars{}-D & .788 & .582 & .314 & .417\\
S-FFSD & HGB-\cars{}-D & .845 & .678 & .430 & .504\\
S-FFSD & CA-HGB & .702 & .509 & .356 & .398\\
S-FFSD & \cars{}+AT-HGB & .845 & .678 & .430 & .504\\
Tolokers & MLP & .740 & .385 & .032 & .137\\
Tolokers & HGB-X & .734 & .388 & .027 & .154\\
Tolokers & R-GCN & .649 & .312 & .012 & .105\\
Tolokers & HGB-Graph & .524 & .222 & .008 & .051\\
Tolokers & R-GCN-\cars{}-D & .741 & .391 & .035 & .147\\
Tolokers & HGB-\cars{}-D & .734 & .388 & .027 & .154\\
Tolokers & CA-HGB & .680 & .323 & .021 & .107\\
Tolokers & \cars{}+AT-HGB & .734 & .388 & .027 & .154\\

%% file: tables/tab_gatetheory.tex
2.0 & 1.0 & .715 / .715 & .708 / .690 & .702 / .690 & .696 / .690 & .692 / .690 & .690\\
4.0 & 1.0 & .772 / .772 & .752 / .697 & .731 / .690 & .711 / .690 & .698 / .690 & .690\\
8.0 & 2.0 & .824 / .824 & .791 / .701 & .757 / .690 & .724 / .690 & .703 / .690 & .690\\
16.0 & 4.0 & .884 / .884 & .837 / .706 & .789 / .690 & .740 / .690 & .710 / .690 & .690\\

%% file: tables/tab_gategraph.tex
0.7 & 0.5 & 0.5 & .145 & .712 & .717 & .148 & .716\\
0.7 & 0.5 & 1 & .248 & .712 & .717 & .635 & .730\\
0.7 & 0.85 & 0.5 & .028 & .716 & .717 & .029 & .717\\
0.7 & 0.85 & 1 & .060 & .716 & .717 & .692 & .722\\
0.85 & 0.5 & 0.5 & .132 & .712 & .717 & .133 & .716\\
0.85 & 0.5 & 1 & .243 & .712 & .717 & .630 & .733\\
0.85 & 0.85 & 0.5 & .032 & .712 & .717 & .035 & .714\\
0.85 & 0.85 & 1 & .064 & .712 & .717 & .692 & .719\\

%% file: tables/tab_gatereal.tex
gate cannot be opened ($B<\cin_r\tau$) & 80 & 14 & 53 & 13 & $+$.004 & $+$.061\\
gate can be opened & 90 & 0 & 81 & 9 & .000 & $+$.001\\
all & 170 & 14 & 134 & 22 & $+$.002 & $+$.061\\

%% file: tables/tab_gatereal_ds.tex
Amazon & U-P-U & 15\% & 0/0/5 & 0/0/5 & .000\\
Amazon & U-S-U & 1\% & 0/0/5 & 0/0/5 & $+$.001\\
Amazon & U-V-U & 4\% & 0/0/5 & 0/0/5 & $+$.001\\
FDCompCN & C-I-C & 72\% & 0/3/5 & 0/5/5 & $-$.001\\
FDCompCN & C-P-C & 88\% & 0/0/5 & 0/0/5 & .000\\
FDCompCN & C-S-C & 88\% & 0/0/5 & 0/0/5 & $-$.001\\
S-FFSD & T-S-T & 26\% & 1/0/5 & 5/0/5 & $+$.061\\
S-FFSD & T-M-T & 0\% & 0/0/5 & 0/0/5 & .000\\
S-FFSD & T-L-T & 0\% & 0/0/5 & 0/0/5 & .000\\
S-FFSD & T-Y-T & 0\% & 0/0/5 & 0/0/5 & .000\\
Synthetic & identity & 0\% & 0/0/5 & 0/0/5 & .000\\
Synthetic & behavior & 0\% & 0/3/5 & 0/3/5 & .000\\
Synthetic & dense & 0\% & 0/0/5 & 0/0/5 & .000\\
Tolokers & W-T-W & 0\% & 0/5/5 & 0/2/5 & .000\\
YelpChi & R-U-R & 82\% & 3/1/5 & 5/0/5 & $+$.021\\
YelpChi & R-T-R & 1\% & 0/0/5 & 0/0/5 & .000\\
YelpChi & R-S-R & 0\% & 0/0/5 & 0/0/5 & .000\\

%% file: tables/tab_ablation2.tex
plug-in subset (\cars{}) & 1 & $-$.028 & $-$.499 & 23/90 & .036\\
plug-in gates & 1 & $-$.054 & $-$.579 & 56/90 & .061\\
val: free/subset & 2 & .007 & $-$.014 & 2/90 & .001\\
val: free/gates & 2 & .007 & $-$.014 & 3/90 & .000\\
val: subset/gates & 2 & $-$.018 & $-$.372 & 21/90 & .025\\
\cars{}-D & 3 & .007 & $-$.014 & 3/90 & .000\\

%% file: tables/tab_selacc.tex
YelpChi & 45 & 100\% & 15 & 15 & 15 & .000 & .000\\
Amazon & 45 & 93\% & 1 & 2 & 0 & .000 & .007\\
FDCompCN & 45 & 89\% & 4 & 6 & 1 & .001 & .014\\
S-FFSD & 45 & 100\% & 15 & 15 & 15 & .000 & .000\\
Tolokers & 45 & 87\% & 8 & 2 & 2 & .001 & .010\\

%% file: tables/tab_labels.tex
5\% & 26 & .005 & $-$.014 & $-$.099 & 9.4\% & .002\\
10\% & 51 & .005 & $-$.008 & $-$.076 & 8.9\% & .002\\
25\% & 128 & .006 & $-$.002 & $-$.064 & 5.3\% & .001\\
50\% & 256 & .007 & $-$.000 & $-$.052 & 4.1\% & .001\\
100\% & 513 & .007 & .000 & $-$.014 & 3.3\% & .000\\

%% file: tables/tab_decomp.tex
CARS-D & 2 & .000 & 2/16 & $-$.001 & 1/12 & $+$.112 & 2/52 & $+$.011 & $+$.110 & 6\%\\
CARS-D, cost-blind planning & 2 & .000 & 0/16 & $-$.003 & 3/12 & $+$.076 & 1/52 & $+$.047 & $+$.074 & 1\%\\
without validation (plug-in subset) & 1 & $-$.005 & 4/16 & $-$.078 & 4/12 & $+$.101 & 10/52 & $+$.022 & $+$.101 & 19\%\\
without validation (plug-in gates) & 1 & $-$.044 & 8/16 & $-$.203 & 11/12 & $+$.076 & 14/52 & $+$.047 & $+$.076 & 27\%\\
without gates & 2 & .000 & 2/16 & $-$.001 & 2/12 & $+$.112 & 0/52 & $+$.011 & $+$.110 & 3\%\\
without graph-free option & 2 & $-$.002 & 4/16 & $-$.078 & 3/12 & $+$.103 & 11/52 & $+$.020 & $+$.102 & 21\%\\
VAL-FA & 2 & .000 & 0/16 & .000 & 0/12 & $+$.065 & 0/52 & $+$.058 & $+$.063 & 1\%\\
VAL-RAND-3 & 3 & $-$.001 & 2/16 & .000 & 0/12 & $+$.037 & 1/52 & $+$.086 & $+$.033 & 6\%\\
VAL-RAND-8 & 8 & $-$.002 & 4/16 & .000 & 0/12 & $+$.067 & 1/52 & $+$.056 & $+$.061 & 9\%\\
VAL-SUB & 16 & .000 & 2/16 & -- & -- & $+$.060 & 0/20 & $+$.001 & $+$.054 & 4\%\\
VAL-POOL & 13 & $-$.002 & 4/16 & $-$.004 & 4/12 & $+$.121 & 2/52 & $+$.002 & $+$.116 & 8\%\\

%% file: tables/tab_decomp_ds.tex
Synthetic & 1 & 4 & $+$.005 [.000, $+$.016] & $+$.005 [.000, $+$.016] & $+$.010 [.000, $+$.031] & .000 [.000, .000] & $+$.005 [.000, $+$.016] & .000 [.000, .000]\\
FDCompCN & 1 & 4 & $-$.004 [$-$.011, .000] & $-$.004 [$-$.011, .000] & .000 [.000, .000] & $+$.001 [.000, $+$.002] & $+$.006 [$-$.005, $+$.020] & $+$.007 [.000, $+$.019]\\
Tolokers & 1 & 4 & .000 [.000, .000] & .000 [.000, .000] & $+$.017 [$-$.004, $+$.037] & .000 [.000, .000] & .000 [.000, .000] & .000 [.000, .000]\\
Amazon & 1 & 4 & $-$.001 [$-$.002, .000] & .000 [$-$.001, .000] & $+$.151 [$+$.001, $+$.435] & $-$.001 [$-$.002, .000] & .000 [$-$.001, .000] & .000 [.000, .000]\\
FDCompCN+A & 1 & 4 & $-$.004 [$-$.010, .000] & $+$.004 [$-$.011, $+$.020] & $+$.050 [.000, $+$.119] & $+$.001 [.000, $+$.002] & $-$.004 [$-$.010, .000] & $+$.008 [$+$.001, $+$.020]\\
Tolokers+A & 1 & 4 & .000 [.000, .000] & .000 [.000, .000] & $+$.069 [$+$.046, $+$.092] & .000 [.000, .000] & .000 [.000, .000] & .000 [.000, .000]\\
Amazon+A & 1 & 4 & .000 [.000, .000] & .000 [.000, .000] & $+$.487 [$+$.046, $+$.927] & .000 [.000, .000] & .000 [.000, .000] & .000 [.000, .000]\\
CSBM, $R{=}4$ & 5 & 20 & $+$.050 [$+$.006, $+$.126] & $+$.010 [.000, $+$.019] & $+$.008 [.000, $+$.019] & .000 [.000, .000] & $+$.003 [$-$.004, $+$.013] & $-$.010 [$-$.025, .000]\\
CSBM, $R{=}8$ & 5 & 20 & $+$.139 [$+$.012, $+$.267] & $+$.056 [$+$.009, $+$.105] & $+$.029 [$+$.002, $+$.056] & .000 [$-$.006, $+$.007] & $+$.064 [$+$.012, $+$.117] & .000 [.000, .000]\\
CSBM, $R{=}12$ & 3 & 12 & $+$.172 [$+$.122, $+$.212] & $+$.047 [$-$.038, $+$.141] & $+$.097 [$+$.023, $+$.170] & $+$.001 [.000, $+$.004] & $+$.084 [$+$.036, $+$.159] & $-$.022 [$-$.038, .000]\\

%% file: tables/tab_hardening.tex
exposure per unit cost (no attack) & 15/16 & 94\% & 100\%\\
uniform-cost exact analysis & 4/16 & 25\% & 11\%\\
most informative relation & 5/16 & 31\% & 22\%\\
largest weight norm & 4/16 & 25\% & 11\%\\
cheapest relation & 7/16 & 44\% & 33\%\\

%% file: tables/tab_voi.tex
0 & 0\% & $-$.067 & 11/40 & $-$.067 & 11/40 & $-$.028 & 9/40\\
0.5 & 2\% & $-$.067 & 11/40 & $-$.036 & 8/40 & $-$.041 & 7/40\\
1 & 32\% & $-$.067 & 11/40 & $-$.017 & 4/40 & $-$.023 & 4/40\\
2 & 90\% & $-$.067 & 11/40 & .000 & 1/40 & .000 & 0/40\\
3 & 100\% & $-$.067 & 11/40 & .000 & 0/40 & .000 & 0/40\\

%% file: tables/tab_ci.tex
YelpChi & $+$.000 [$+$.000, $+$.000] & \textbf{$+$.021 [$+$.019, $+$.024]} & $+$.000 [$+$.000, $+$.000] & \textbf{$+$.061 [$+$.057, $+$.066]} & $+$.000 [$+$.000, $+$.000] & \textbf{$+$.017 [$+$.015, $+$.020]} & $+$.000 [$+$.000, $+$.000] & \textbf{$+$.017 [$+$.015, $+$.020]}\\
Amazon & $+$.000 [$+$.000, $+$.000] & $+$.000 [$+$.000, $+$.000] & $+$.000 [$+$.000, $+$.000] & $+$.000 [$+$.000, $+$.000] & $-$.002 [$-$.006, $+$.001] & $+$.000 [$+$.000, $+$.000] & $-$.001 [$-$.005, $+$.002] & $+$.000 [$+$.000, $+$.000]\\
FDCompCN & $-$.002 [$-$.012, $+$.009] & $+$.000 [$-$.000, $+$.000] & $+$.000 [$+$.000, $+$.000] & $+$.000 [$+$.000, $+$.000] & $-$.000 [$-$.003, $+$.003] & $+$.000 [$+$.000, $+$.000] & $-$.000 [$-$.003, $+$.003] & $+$.000 [$+$.000, $+$.000]\\
S-FFSD & $+$.000 [$+$.000, $+$.000] & \textbf{$+$.020 [$+$.018, $+$.023]} & $+$.000 [$+$.000, $+$.000] & \textbf{$+$.070 [$+$.062, $+$.079]} & $+$.000 [$+$.000, $+$.000] & \textbf{$+$.018 [$+$.015, $+$.021]} & $+$.000 [$+$.000, $+$.000] & \textbf{$+$.018 [$+$.015, $+$.021]}\\
Tolokers & $+$.000 [$+$.000, $+$.000] & $+$.000 [$+$.000, $+$.000] & $+$.001 [$-$.001, $+$.005] & $+$.001 [$-$.001, $+$.005] & $+$.000 [$+$.000, $+$.000] & $+$.000 [$+$.000, $+$.000] & $+$.000 [$+$.000, $+$.000] & $+$.000 [$+$.000, $+$.000]\\

%% file: tables/tab_coststab.tex
Amazon & independent & 40 & 0\% & 4 & $-$.004 & 2 & $+$.003 & 2.12 & 0.25\\
Amazon & structured & 40 & 0\% & 8 & $-$.002 & 0 & -- & -- & 0.25\\
FDCompCN & independent & 40 & 0\% & 9 & $-$.023 & 0 & -- & -- & 0.25\\
FDCompCN & structured & 40 & 0\% & 25 & $-$.014 & 0 & -- & -- & 0.25\\
S-FFSD & independent & 40 & 100\% & 0 & $-$.001 & 11 & $+$.014 & 2.25 & 0.25\\
S-FFSD & structured & 40 & 100\% & 0 & .000 & 18 & $+$.010 & 0.50 & 0.25\\
Synthetic & independent & 40 & 90\% & 0 & .000 & 10 & $+$.087 & 1.88 & 0.25\\
Synthetic & structured & 40 & 98\% & 0 & .000 & 7 & $+$.051 & 0.50 & 0.25\\
Tolokers & independent & 40 & 75\% & 0 & .000 & 7 & $+$.036 & 1.75 & 1.00\\
Tolokers & structured & 40 & 100\% & 0 & .000 & 0 & -- & -- & 0.75\\
YelpChi & independent & 40 & 100\% & 0 & .000 & 10 & $+$.010 & 0.88 & 0.25\\
YelpChi & structured & 40 & 100\% & 0 & .000 & 25 & $+$.005 & 0.50 & 0.25\\

%% file: tables/tab_ci_full.tex
YelpChi & RLR & P1, $B{=}2$ & 0 & 0.0000 & [0, 0] & 1.00 & yes & 0.0000\\
YelpChi & RLR & P1, $B{=}20$ & 0 & 0.0000 & [0, 0] & 1.00 & yes & 0.0000\\
YelpChi & RLR & P2, $B{=}20$ & $+$.0213 & 0.0026 & [$+$.0187, $+$.0240] & 0.00 & yes & 0.0000\\
YelpChi & R-GCN & P1, $B{=}2$ & 0 & 0.0000 & [0, 0] & 1.00 & yes & 0.0000\\
YelpChi & R-GCN & P1, $B{=}20$ & 0 & 0.0000 & [0, 0] & 1.00 & yes & 0.0000\\
YelpChi & R-GCN & P2, $B{=}20$ & $+$.0612 & 0.0039 & [$+$.0568, $+$.0658] & 0.00 & yes & 0.0000\\
YelpChi & HGB & P1, $B{=}2$ & 0 & 0.0000 & [0, 0] & 1.00 & yes & 0.0000\\
YelpChi & HGB & P1, $B{=}20$ & 0 & 0.0000 & [0, 0] & 1.00 & yes & 0.0000\\
YelpChi & HGB & P2, $B{=}20$ & $+$.0174 & 0.0028 & [$+$.0150, $+$.0203] & 0.00 & yes & 0.0000\\
Amazon & RLR & P1, $B{=}2$ & 0 & 0.0000 & [0, 0] & 1.00 & yes & 0.0000\\
Amazon & RLR & P1, $B{=}20$ & 0 & 0.0000 & [0, 0] & 1.00 & yes & 0.0000\\
Amazon & RLR & P2, $B{=}20$ & 0 & 0.0000 & [0, 0] & 1.00 & yes & 0.0000\\
Amazon & R-GCN & P1, $B{=}2$ & 0 & 0.0000 & [0, 0] & 1.00 & yes & 0.0000\\
Amazon & R-GCN & P1, $B{=}20$ & 0 & 0.0000 & [0, 0] & 1.00 & yes & 0.0000\\
Amazon & R-GCN & P2, $B{=}20$ & 0 & 0.0000 & [0, 0] & 1.00 & yes & 0.0000\\
Amazon & HGB & P1, $B{=}2$ & $-$.0019 & 0.0032 & [$-$.0058, $+$.0005] & 1.00 & yes & 0.0058\\
Amazon & HGB & P1, $B{=}20$ & 0 & 0.0000 & [0, 0] & 1.00 & yes & 0.0000\\
Amazon & HGB & P2, $B{=}20$ & 0 & 0.0000 & [0, 0] & 1.00 & yes & 0.0000\\
FDCompCN & RLR & P1, $B{=}2$ & $-$.0017 & 0.0083 & [$-$.0122, $+$.0091] & 1.00 & no & 0.0122\\
FDCompCN & RLR & P1, $B{=}20$ & $+$.0000 & 0.0000 & [$-$.0000, $+$.0000] & 1.00 & yes & 0.0000\\
FDCompCN & RLR & P2, $B{=}20$ & $+$.0000 & 0.0000 & [$-$.0000, $+$.0000] & 1.00 & yes & 0.0000\\
FDCompCN & R-GCN & P1, $B{=}2$ & 0 & 0.0000 & [0, 0] & 1.00 & yes & 0.0000\\
FDCompCN & R-GCN & P1, $B{=}20$ & 0 & 0.0000 & [0, 0] & 1.00 & yes & 0.0000\\
FDCompCN & R-GCN & P2, $B{=}20$ & 0 & 0.0000 & [0, 0] & 1.00 & yes & 0.0000\\
FDCompCN & HGB & P1, $B{=}2$ & $-$.0001 & 0.0003 & [$-$.0034, $+$.0025] & 1.00 & yes & 0.0034\\
FDCompCN & HGB & P1, $B{=}20$ & 0 & 0.0000 & [0, 0] & 1.00 & yes & 0.0000\\
FDCompCN & HGB & P2, $B{=}20$ & 0 & 0.0000 & [0, 0] & 1.00 & yes & 0.0000\\
S-FFSD & RLR & P1, $B{=}2$ & 0 & 0.0000 & [0, 0] & 1.00 & yes & 0.0000\\
S-FFSD & RLR & P1, $B{=}20$ & 0 & 0.0000 & [0, 0] & 1.00 & yes & 0.0000\\
S-FFSD & RLR & P2, $B{=}20$ & $+$.0203 & 0.0021 & [$+$.0179, $+$.0228] & 0.00 & yes & 0.0000\\
S-FFSD & R-GCN & P1, $B{=}2$ & 0 & 0.0000 & [0, 0] & 1.00 & yes & 0.0000\\
S-FFSD & R-GCN & P1, $B{=}20$ & 0 & 0.0000 & [0, 0] & 1.00 & yes & 0.0000\\
S-FFSD & R-GCN & P2, $B{=}20$ & $+$.0698 & 0.0093 & [$+$.0617, $+$.0789] & 0.00 & yes & 0.0000\\
S-FFSD & HGB & P1, $B{=}2$ & 0 & 0.0000 & [0, 0] & 1.00 & yes & 0.0000\\
S-FFSD & HGB & P1, $B{=}20$ & 0 & 0.0000 & [0, 0] & 1.00 & yes & 0.0000\\
S-FFSD & HGB & P2, $B{=}20$ & $+$.0179 & 0.0013 & [$+$.0152, $+$.0205] & 0.00 & yes & 0.0000\\
Tolokers & RLR & P1, $B{=}2$ & 0 & 0.0000 & [0, 0] & 1.00 & yes & 0.0000\\
Tolokers & RLR & P1, $B{=}20$ & 0 & 0.0000 & [0, 0] & 1.00 & yes & 0.0000\\
Tolokers & RLR & P2, $B{=}20$ & 0 & 0.0000 & [0, 0] & 1.00 & yes & 0.0000\\
Tolokers & R-GCN & P1, $B{=}2$ & $+$.0009 & 0.0020 & [$-$.0012, $+$.0046] & 1.00 & yes & 0.0012\\
Tolokers & R-GCN & P1, $B{=}20$ & 0 & 0.0000 & [0, 0] & 1.00 & yes & 0.0000\\
Tolokers & R-GCN & P2, $B{=}20$ & $+$.0008 & 0.0019 & [$-$.0013, $+$.0045] & 1.00 & yes & 0.0013\\
Tolokers & HGB & P1, $B{=}2$ & 0 & 0.0000 & [0, 0] & 1.00 & yes & 0.0000\\
Tolokers & HGB & P1, $B{=}20$ & 0 & 0.0000 & [0, 0] & 1.00 & yes & 0.0000\\
Tolokers & HGB & P2, $B{=}20$ & 0 & 0.0000 & [0, 0] & 1.00 & yes & 0.0000\\

%% file: tables/tab_costagnostic.tex
Amazon & 2 & .000 & .000 & .000 & .000 & .001 & 0/16\\
Amazon & 20 & .000 & .000 & .000 & .000 & .000 & 0/16\\
FDCompCN & 2 & .000 & .000 & $-$.003 & $-$.003 & .000 & 0/16\\
FDCompCN & 20 & .000 & .000 & $-$.004 & $-$.004 & .000 & 0/16\\
S-FFSD & 2 & .000 & .000 & .005 & .005 & .016 & 0/16\\
S-FFSD & 20 & .000 & .000 & .001 & .001 & .010 & 0/16\\
Tolokers & 2 & .000 & .000 & .015 & .015 & .047 & 0/16\\
Tolokers & 20 & .000 & .000 & .000 & .000 & .000 & 0/16\\
YelpChi & 2 & .000 & .000 & .007 & .007 & .029 & 0/16\\
YelpChi & 20 & .000 & .000 & .001 & .001 & .006 & 0/16\\

%% file: tables/tab_noiserobust.tex
Amazon & 0.5 & .969 & .969 & 0/10 & .000 & 0/10 & .000 & 0/10 & .000\\
Amazon & 1 & .969 & .969 & 4/10 & .027 & 1/10 & .002 & 0/10 & .000\\
FDCompCN & 0.5 & .663 & .677 & 10/10 & .000 & 6/10 & $-$.006 & 1/10 & $-$.013\\
FDCompCN & 1 & .663 & .677 & 5/10 & .125 & 5/10 & .059 & 1/10 & $-$.013\\
S-FFSD & 0.5 & .643 & .643 & 3/10 & .091 & 1/10 & .035 & 0/10 & .000\\
S-FFSD & 1 & .643 & .643 & 4/10 & .103 & 1/10 & .035 & 0/10 & .000\\
Tolokers & 0.5 & .716 & .716 & 1/10 & .018 & 0/10 & .000 & 0/10 & .000\\
Tolokers & 1 & .716 & .716 & 3/10 & .053 & 0/10 & .000 & 0/10 & .000\\
YelpChi & 0.5 & .771 & .771 & 4/10 & .255 & 1/10 & .064 & 0/10 & .000\\
YelpChi & 1 & .771 & .771 & 7/10 & .374 & 1/10 & .060 & 0/10 & .000\\

%% file: tables/tab_breakeven.tex
Amazon & U-P-U & $-$.001 & -- & -- & no clean gain\\
Amazon & U-S-U & $+$.003 & 1 & 2 & 20\\
Amazon & U-V-U & $-$.001 & -- & -- & no clean gain\\
FDCompCN & C-I-C & $-$.005 & -- & -- & no clean gain\\
FDCompCN & C-P-C & $-$.028 & -- & -- & no clean gain\\
FDCompCN & C-S-C & $-$.006 & -- & -- & no clean gain\\
S-FFSD & T-S-T & $+$.018 & 2 & 1 & 10\\
S-FFSD & T-M-T & $-$.001 & -- & -- & no clean gain\\
S-FFSD & T-L-T & $-$.001 & -- & -- & no clean gain\\
S-FFSD & T-Y-T & $-$.000 & -- & -- & no clean gain\\
Tolokers & W-T-W & $+$.047 & 1 & 2 & 20\\
YelpChi & R-U-R & $+$.022 & 2 & 1 & 10\\
YelpChi & R-T-R & $+$.000 & -- & -- & no clean gain\\
YelpChi & R-S-R & $+$.010 & 1 & 2 & 20\\

%% file: tables/tab_temporal2.tex
0 & P1, $B{=}2$ & .531 & .325 & .531 & .619 & .308 & .619\\
0 & P2, $B{=}20$ & .531 & .132 & .758 & .619 & .213 & .717\\
1 & P1, $B{=}2$ & .572 & .284 & .572 & .619 & .272 & .619\\
1 & P2, $B{=}20$ & .572 & .064 & .680 & .619 & .122 & .731\\
2 & P1, $B{=}2$ & .648 & .165 & .648 & .610 & .190 & .610\\
2 & P2, $B{=}20$ & .648 & .050 & .828 & .610 & .119 & .784\\
3 & P1, $B{=}2$ & .461 & .079 & .461 & .686 & .108 & .686\\
3 & P2, $B{=}20$ & .461 & .161 & .862 & .686 & .100 & .812\\

%% file: tables/tab_noise.tex
Amazon & 2 & 0.5 & common & .969 & .969 & .969 & .969 & 0\%\\
Amazon & 2 & 0.5 & independent & .969 & .969 & .969 & .969 & 20\%\\
Amazon & 2 & 1 & common & .969 & .945 & .847 & .969 & 20\%\\
Amazon & 2 & 1 & independent & .969 & .940 & .851 & .969 & 0\%\\
Amazon & 20 & 0.5 & common & .969 & .969 & .969 & .969 & 40\%\\
Amazon & 20 & 0.5 & independent & .969 & .969 & .969 & .969 & 60\%\\
Amazon & 20 & 1 & common & .969 & .969 & .969 & .969 & 20\%\\
Amazon & 20 & 1 & independent & .969 & .969 & .969 & .969 & 0\%\\
FDCompCN & 2 & 0.5 & common & .663 & .663 & .663 & .677 & 100\%\\
FDCompCN & 2 & 0.5 & independent & .663 & .663 & .663 & .677 & 100\%\\
FDCompCN & 2 & 1 & common & .663 & .671 & .663 & .677 & 40\%\\
FDCompCN & 2 & 1 & independent & .663 & .405 & .003 & .677 & 20\%\\
FDCompCN & 20 & 0.5 & common & .677 & .677 & .677 & .677 & 100\%\\
FDCompCN & 20 & 0.5 & independent & .677 & .677 & .677 & .677 & 100\%\\
FDCompCN & 20 & 1 & common & .677 & .677 & .677 & .677 & 100\%\\
FDCompCN & 20 & 1 & independent & .677 & .677 & .677 & .677 & 100\%\\
S-FFSD & 2 & 0.5 & common & .643 & .643 & .643 & .643 & 0\%\\
S-FFSD & 2 & 0.5 & independent & .643 & .460 & .289 & .643 & 0\%\\
S-FFSD & 2 & 1 & common & .643 & .587 & .361 & .643 & 0\%\\
S-FFSD & 2 & 1 & independent & .643 & .494 & .289 & .643 & 0\%\\
S-FFSD & 20 & 0.5 & common & .643 & .643 & .643 & .643 & 100\%\\
S-FFSD & 20 & 0.5 & independent & .643 & .643 & .643 & .643 & 100\%\\
S-FFSD & 20 & 1 & common & .643 & .643 & .643 & .643 & 100\%\\
S-FFSD & 20 & 1 & independent & .643 & .643 & .643 & .643 & 100\%\\
Tolokers & 2 & 0.5 & common & .716 & .716 & .716 & .716 & 60\%\\
Tolokers & 2 & 0.5 & independent & .716 & .680 & .538 & .716 & 0\%\\
Tolokers & 2 & 1 & common & .716 & .680 & .538 & .716 & 40\%\\
Tolokers & 2 & 1 & independent & .716 & .644 & .538 & .716 & 20\%\\
Tolokers & 20 & 0.5 & common & .716 & .716 & .716 & .716 & 100\%\\
Tolokers & 20 & 0.5 & independent & .716 & .716 & .716 & .716 & 100\%\\
Tolokers & 20 & 1 & common & .716 & .716 & .716 & .716 & 100\%\\
Tolokers & 20 & 1 & independent & .716 & .716 & .716 & .716 & 100\%\\
YelpChi & 2 & 0.5 & common & .771 & .517 & .134 & .771 & 60\%\\
YelpChi & 2 & 0.5 & independent & .771 & .517 & .134 & .771 & 40\%\\
YelpChi & 2 & 1 & common & .771 & .524 & .134 & .771 & 20\%\\
YelpChi & 2 & 1 & independent & .771 & .271 & .134 & .771 & 0\%\\
YelpChi & 20 & 0.5 & common & .771 & .771 & .771 & .771 & 80\%\\
YelpChi & 20 & 0.5 & independent & .771 & .771 & .771 & .771 & 60\%\\
YelpChi & 20 & 1 & common & .771 & .771 & .771 & .771 & 60\%\\
YelpChi & 20 & 1 & independent & .771 & .771 & .771 & .771 & 40\%\\

%% file: tables/tab_misspec.tex
YelpChi & 2 & P1 & .933/.933 & .933/.933 & .933/.933 & .933/.933 & .933\\
YelpChi & 2 & P2 & .417/.948 & .417/.948 & .417/.948 & .417/.948 & .933\\
YelpChi & 20 & P1 & .933/.933 & .933/.933 & .933/.933 & .933/.933 & .933\\
YelpChi & 20 & P2 & .269/.948 & .933/.933 & .933/.933 & .933/.933 & .933\\

%% file: tables/tab_abandon.tex
Amazon & P1 & 0, 1, 0 & 1 & 1 & 1\\
Amazon & P2 & 0, 1, 0 & 1 & 1 & 1\\
FDCompCN & P1 & 0, 0, 0 & 0 & 0 & 0\\
FDCompCN & P2 & 0, 0, 0 & 0 & 0 & 0\\
S-FFSD & P1 & 2, 0, 0, 0 & 1 & 2 & 2\\
S-FFSD & P2 & 25, 0, 0, 0 & 1 & 25 & 25\\
Tolokers & P1 & 1 & 1 & 1 & 1\\
Tolokers & P2 & 8 & 8 & 8 & 8\\
YelpChi & P1 & 2, 1, 1 & 1 & 4 & 2\\
YelpChi & P2 & 25, 1, 1 & 1 & 27 & 25\\

%% file: tables/tab_gang2.tex
Tolokers & HGB-\cars{}-D & 0 & .729 & .729 & .729 & -- & -- & -- & .729\\
Tolokers & HGB-\cars{}-D & 0.5 & .729 & .729 & .729 & -- & -- & -- & .729\\
Tolokers & HGB-\cars{}-D & 1 & .729 & .729 & .729 & -- & -- & -- & .729\\
Tolokers & HGB-Graph & 0 & .625 & .561 & .509 & .652 & .596 & .540 & .729\\
Tolokers & HGB-Graph & 0.5 & .659 & .590 & .524 & .681 & .622 & .559 & .729\\
Tolokers & HGB-Graph & 1 & .682 & .614 & .540 & .702 & .643 & .576 & .729\\
Tolokers & R-GCN & 0 & .700 & .661 & .615 & .710 & .674 & .632 & .729\\
Tolokers & R-GCN & 0.5 & .711 & .671 & .622 & .719 & .683 & .638 & .729\\
Tolokers & R-GCN & 1 & .720 & .680 & .630 & .727 & .690 & .644 & .729\\
YelpChi & HGB-\cars{}-D & 0 & .933 & .933 & .933 & -- & -- & -- & .933\\
YelpChi & HGB-\cars{}-D & 0.5 & .933 & .933 & .933 & -- & -- & -- & .933\\
YelpChi & HGB-\cars{}-D & 1 & .933 & .933 & .933 & -- & -- & -- & .933\\
YelpChi & HGB-Graph & 0 & .766 & .610 & .421 & .784 & .632 & .438 & .933\\
YelpChi & HGB-Graph & 0.5 & .801 & .668 & .473 & .816 & .691 & .489 & .933\\
YelpChi & HGB-Graph & 1 & .825 & .710 & .526 & .839 & .731 & .546 & .933\\
YelpChi & R-GCN & 0 & .673 & .433 & .048 & .681 & .455 & .088 & .933\\
YelpChi & R-GCN & 0.5 & .723 & .529 & .170 & .728 & .545 & .198 & .933\\
YelpChi & R-GCN & 1 & .756 & .594 & .283 & .760 & .604 & .309 & .933\\

%% file: tables/tab_factorial.tex
R-GCN (vs HGB-Graph) & -0.028\\
RLR (vs HGB-Graph) & -0.157\\
$p_{\mathrm{id}}$ & -0.190\\
isolated fraction & -0.334\\
$\log\cid$ & +0.254\\
$p_{\mathrm{id}}\times$ isolated & +0.173\\
intercept & +0.028\\
$R^2$ & 0.83\\

%% file: tables/tab_atcompute.tex
FDCompCN & P1, $B{=}2$ & .702 (16s) & 37: .453, 50: .490, 100: .266, 200: .531, 400: .472\\
Synthetic & P1, $B{=}2$ & .814 (21s) & 36: .867, 50: .885, 100: .902, 200: .894, 400: .889\\
Tolokers & P1, $B{=}2$ & .736 (50s) & 28: .693, 50: .739, 100: .737, 200: .724, 400: .700\\

%% file: tables/tab_syn.tex
MLP (graph-free) & .681 & .681 & .681 & .681 & .681 & .681\\
LR (graph-free) & .702 & .702 & .702 & .702 & .702 & .702\\
HGB-X (graph-free) & .616 & .616 & .616 & .616 & .616 & .616\\
RLR & .893 & .518 & .170 & .028 & .004 & .000\\
R-GCN & .993 & .756 & .280 & .067 & .012 & .002\\
HGB-Graph & .888 & .508 & .389 & .167 & .042 & .011\\
CA-AT (R-GCN) & .991 & .952 & .919 & .880 & .755 & .486\\
AT-RLR (exact) & .855 & .775 & .725 & .571 & .394 & .455\\
RLR + \cars{} & .843 & .802 & .702 & .702 & .702 & .702\\
R-GCN + \cars{} & .951 & .922 & .828 & .681 & .681 & .681\\
HGB + \cars{} & .820 & .717 & .530 & .616 & .616 & .616\\
R-GCN + \cars{}-D & .951 & .922 & .828 & .681 & .681 & .681\\
\cars{}+AT & .991 & .952 & .919 & .880 & .755 & .681\\

%% file: tables/tab_gap.tex
RLR, $B{=}1$ & .727 & .753 & .659 & .518\\
RLR, $B{=}2$ & .610 & .608 & .415 & .170\\
RLR, $B{=}4$ & .462 & .486 & .231 & .028\\
RLR, $B{=}8$ & .422 & .359 & .108 & .004\\
RLR, $B{=}16$ & .318 & .211 & .023 & .000\\

%% file: tables/tab_ablation.tex
RLR & .007 & -0.020 & .007 & -0.006 & -0.205 & .000 & -0.002 & -0.048 & -0.001 & -0.040 & -0.074 & .007 & .000 & -0.011 & .000\\
R-GCN & .020 & -0.009 & .020 & -0.010 & -0.058 & .000 & -0.015 & -0.064 & .000 & -0.130 & -0.069 & .023 & .000 & -0.004 & .001\\
HGB & .006 & -0.028 & .006 & -0.000 & -0.014 & -0.001 & -0.001 & -0.006 & -0.000 & -0.126 & -0.091 & .006 & .000 & -0.036 & .000\\

%% file: tables/tab_temporal.tex
LR & .425 & .425 & .425 & .425\\
MLP & .560 & .560 & .560 & .560\\
HGB-X & .593 & .593 & .593 & .593\\
RLR & .361 & .202 & .035 & .091\\
R-GCN & .811 & .145 & .105 & .178\\
HGB & .820 & .186 & .130 & .178\\
CA-HGB & .791 & .334 & .230 & .513\\
RLR+C & .361 & .202 & .425 & .423\\
R-GCN+C & .811 & .145 & .560 & .765\\
HGB+C & .820 & .186 & .593 & .793\\
RLR+D & .425 & .425 & .425 & .431\\
R-GCN+D & .560 & .560 & .560 & .809\\
HGB+D & .593 & .593 & .593 & .766\\

%% file: tables/tab_exposure.tex
Amazon & delete & 1211 & 0 & 0.030 & 0.05 & .500\\
Amazon & inject & 1974 & 0 & 0.008 & 0.89 & .500\\
FDCompCN & delete & 35 & 0 & 0.000 & -0.10 & 1.000\\
FDCompCN & inject & 1344 & 0 & 0.032 & 0.68 & .500\\
S-FFSD & delete & 2225 & 0 & 0.018 & 0.42 & 1.000\\
S-FFSD & inject & 3200 & 0 & 0.030 & 0.52 & .800\\
Tolokers & delete & 767 & 0 & nan & 0.59 & --\\
Tolokers & inject & 800 & 0 & 0.040 & 0.92 & --\\
YelpChi & delete & 1634 & 0 & 0.041 & 0.32 & 1.000\\
YelpChi & inject & 2400 & 0 & 0.059 & 0.95 & 1.000\\

%% file: tables/tab_thresholds.tex
\begin{tabular}{lrrrrrr}
\toprule
$\cid$ & $\Btrap$ (num.) & envelope & exposure LB & linear resp. & abandon (num.) & $\sum_r\kappa_r$\\
\midrule
0.25 & 2.1 & 2.1 & 0.6 & 1.3 & 25.1 & 26.4\\
0.5 & 3.0 & 3.0 & 1.2 & 2.6 & 25.8 & 27.2\\
1 & 4.6 & 4.6 & 2.4 & 5.2 & 27.0 & 28.8\\
2 & 7.8 & 7.8 & 4.9 & 10.3 & 29.6 & 32.0\\
4 & 14.2 & 14.2 & 9.8 & 20.6 & 35.0 & 38.4\\
8 & 27.0 & 27.0 & 17.0 & 29.6 & 46.3 & 51.1\\
16 & 30.4 & 30.5 & 17.2 & 29.6 & 70.3 & 76.6\\
\bottomrule
\end{tabular}

%% file: tables/tab_fidelity.tex
Amazon & P1, $B{=}2$ & 20 & 0.23 & 0.98 & 0.004 & 0.002 & +0.035\\
Amazon & P1, $B{=}20$ & 20 & 0.11 & 1.00 & 0.039 & 0.000 & +0.122\\
Amazon & P2, $B{=}20$ & 20 & 0.43 & 1.00 & 0.004 & 0.000 & +0.116\\
FDCompCN & P1, $B{=}2$ & 20 & 0.52 & 0.71 & 0.017 & 0.017 & +0.147\\
FDCompCN & P1, $B{=}20$ & 20 & 0.69 & 0.97 & 0.004 & 0.000 & +0.217\\
FDCompCN & P2, $B{=}20$ & 20 & 0.72 & 0.88 & 0.000 & 0.026 & +0.196\\
S-FFSD & P1, $B{=}2$ & 32 & 0.68 & 1.00 & 0.097 & 0.000 & +0.101\\
S-FFSD & P1, $B{=}20$ & 32 & 0.68 & 1.00 & 0.000 & 0.000 & +0.138\\
S-FFSD & P2, $B{=}20$ & 32 & 0.81 & 1.00 & 0.041 & 0.003 & +0.071\\
Synthetic & P1, $B{=}2$ & 20 & 0.82 & 1.00 & 0.000 & 0.000 & +0.103\\
Synthetic & P1, $B{=}20$ & 20 & 0.86 & 0.99 & 0.000 & 0.000 & +0.016\\
Synthetic & P2, $B{=}20$ & 20 & 0.87 & 1.00 & 0.000 & 0.000 & +0.037\\
Tolokers & P1, $B{=}2$ & 6 & 0.77 & 1.00 & 0.009 & 0.000 & +0.086\\
Tolokers & P1, $B{=}20$ & 6 & 0.64 & 0.94 & 0.000 & 0.000 & +0.058\\
Tolokers & P2, $B{=}20$ & 6 & 0.83 & 1.00 & 0.009 & 0.000 & +0.082\\
YelpChi & P1, $B{=}2$ & 20 & 0.66 & 0.99 & 0.092 & 0.000 & +0.001\\
YelpChi & P1, $B{=}20$ & 20 & 0.80 & 1.00 & 0.001 & 0.000 & +0.079\\
YelpChi & P2, $B{=}20$ & 20 & 0.71 & 1.00 & 0.002 & 0.000 & +0.093\\

%% file: tables/tab_significance.tex
Amazon & P1, $B{=}2$ & RLR + CARS-D & 5 & +0.000 & [+0.000, +0.000] & 5/5 & .872 / .872 & .769 / .769\\
Amazon & P1, $B{=}20$ & RLR + CARS-D & 5 & +0.000 & [+0.000, +0.000] & 5/5 & .872 / .872 & .769 / .769\\
Amazon & P2, $B{=}20$ & RLR + CARS-D & 5 & +0.000 & [+0.000, +0.000] & 5/5 & .872 / .872 & .769 / .769\\
FDCompCN & P1, $B{=}2$ & RLR + CARS-D & 5 & -0.002 & [-0.012, +0.009] & 3/5 & .207 / .224 & .036 / .041\\
FDCompCN & P1, $B{=}20$ & RLR + CARS-D & 5 & +0.000 & [-0.000, +0.000] & 5/5 & .224 / .224 & .041 / .041\\
FDCompCN & P2, $B{=}20$ & RLR + CARS-D & 5 & +0.000 & [-0.000, +0.000] & 5/5 & .224 / .224 & .041 / .041\\
S-FFSD & P1, $B{=}2$ & RLR + CARS-D & 5 & +0.000 & [+0.000, +0.000] & 5/5 & .261 / .261 & .002 / .002\\
S-FFSD & P1, $B{=}20$ & RLR + CARS-D & 5 & +0.000 & [+0.000, +0.000] & 5/5 & .261 / .261 & .002 / .002\\
S-FFSD & P2, $B{=}20$ & RLR + CARS-D & 5 & +0.020 & [+0.018, +0.023] & 5/5 & .382 / .261 & .084 / .002\\
Tolokers & P1, $B{=}2$ & RLR + CARS-D & 5 & +0.000 & [+0.000, +0.000] & 5/5 & .372 / .372 & .042 / .042\\
Tolokers & P1, $B{=}20$ & RLR + CARS-D & 5 & +0.000 & [+0.000, +0.000] & 5/5 & .372 / .372 & .042 / .042\\
Tolokers & P2, $B{=}20$ & RLR + CARS-D & 5 & +0.000 & [+0.000, +0.000] & 5/5 & .372 / .372 & .042 / .042\\
YelpChi & P1, $B{=}2$ & RLR + CARS-D & 5 & +0.000 & [+0.000, +0.000] & 5/5 & .390 / .390 & .099 / .099\\
YelpChi & P1, $B{=}20$ & RLR + CARS-D & 5 & +0.000 & [+0.000, +0.000] & 5/5 & .390 / .390 & .099 / .099\\
YelpChi & P2, $B{=}20$ & RLR + CARS-D & 5 & +0.021 & [+0.018, +0.025] & 5/5 & .400 / .390 & .090 / .099\\
Amazon & P1, $B{=}2$ & R-GCN + CARS-D & 5 & +0.000 & [+0.000, +0.000] & 5/5 & .889 / .889 & .818 / .818\\
Amazon & P1, $B{=}20$ & R-GCN + CARS-D & 5 & +0.000 & [+0.000, +0.000] & 5/5 & .889 / .889 & .818 / .818\\
Amazon & P2, $B{=}20$ & R-GCN + CARS-D & 5 & +0.000 & [+0.000, +0.000] & 5/5 & .889 / .889 & .818 / .818\\
FDCompCN & P1, $B{=}2$ & R-GCN + CARS-D & 5 & +0.000 & [+0.000, +0.000] & 5/5 & .282 / .282 & .074 / .074\\
FDCompCN & P1, $B{=}20$ & R-GCN + CARS-D & 5 & +0.000 & [+0.000, +0.000] & 5/5 & .282 / .282 & .074 / .074\\
FDCompCN & P2, $B{=}20$ & R-GCN + CARS-D & 5 & +0.000 & [+0.000, +0.000] & 5/5 & .282 / .282 & .074 / .074\\
S-FFSD & P1, $B{=}2$ & R-GCN + CARS-D & 5 & +0.000 & [+0.000, +0.000] & 5/5 & .582 / .582 & .314 / .314\\
S-FFSD & P1, $B{=}20$ & R-GCN + CARS-D & 5 & +0.000 & [+0.000, +0.000] & 5/5 & .582 / .582 & .314 / .314\\
S-FFSD & P2, $B{=}20$ & R-GCN + CARS-D & 5 & +0.070 & [+0.058, +0.081] & 5/5 & .686 / .582 & .420 / .314\\
Tolokers & P1, $B{=}2$ & R-GCN + CARS-D & 5 & +0.001 & [-0.002, +0.003] & 5/5 & .391 / .385 & .035 / .032\\
Tolokers & P1, $B{=}20$ & R-GCN + CARS-D & 5 & +0.000 & [+0.000, +0.000] & 5/5 & .385 / .385 & .032 / .032\\
Tolokers & P2, $B{=}20$ & R-GCN + CARS-D & 5 & +0.001 & [-0.002, +0.003] & 5/5 & .391 / .385 & .035 / .032\\
YelpChi & P1, $B{=}2$ & R-GCN + CARS-D & 5 & +0.000 & [+0.000, +0.000] & 5/5 & .525 / .525 & .178 / .178\\
YelpChi & P1, $B{=}20$ & R-GCN + CARS-D & 5 & +0.000 & [+0.000, +0.000] & 5/5 & .525 / .525 & .178 / .178\\
YelpChi & P2, $B{=}20$ & R-GCN + CARS-D & 5 & +0.061 & [+0.056, +0.066] & 5/5 & .637 / .525 & .238 / .178\\
Amazon & P1, $B{=}2$ & HGB + CARS-D & 5 & -0.002 & [-0.006, +0.002] & 3/5 & .902 / .906 & .810 / .822\\
Amazon & P1, $B{=}20$ & HGB + CARS-D & 5 & +0.000 & [+0.000, +0.000] & 5/5 & .906 / .906 & .822 / .822\\
Amazon & P2, $B{=}20$ & HGB + CARS-D & 5 & +0.000 & [+0.000, +0.000] & 5/5 & .906 / .906 & .822 / .822\\
FDCompCN & P1, $B{=}2$ & HGB + CARS-D & 5 & -0.000 & [-0.001, +0.000] & 5/5 & .412 / .413 & .169 / .167\\
FDCompCN & P1, $B{=}20$ & HGB + CARS-D & 5 & +0.000 & [+0.000, +0.000] & 5/5 & .413 / .413 & .167 / .167\\
FDCompCN & P2, $B{=}20$ & HGB + CARS-D & 5 & +0.000 & [+0.000, +0.000] & 5/5 & .413 / .413 & .167 / .167\\
S-FFSD & P1, $B{=}2$ & HGB + CARS-D & 5 & +0.000 & [+0.000, +0.000] & 5/5 & .678 / .678 & .430 / .430\\
S-FFSD & P1, $B{=}20$ & HGB + CARS-D & 5 & +0.000 & [+0.000, +0.000] & 5/5 & .678 / .678 & .430 / .430\\
S-FFSD & P2, $B{=}20$ & HGB + CARS-D & 5 & +0.018 & [+0.016, +0.020] & 5/5 & .710 / .678 & .448 / .430\\
Tolokers & P1, $B{=}2$ & HGB + CARS-D & 5 & +0.000 & [+0.000, +0.000] & 5/5 & .388 / .388 & .027 / .027\\
Tolokers & P1, $B{=}20$ & HGB + CARS-D & 5 & +0.000 & [+0.000, +0.000] & 5/5 & .388 / .388 & .027 / .027\\
Tolokers & P2, $B{=}20$ & HGB + CARS-D & 5 & +0.000 & [+0.000, +0.000] & 5/5 & .388 / .388 & .027 / .027\\
YelpChi & P1, $B{=}2$ & HGB + CARS-D & 5 & +0.000 & [+0.000, +0.000] & 5/5 & .783 / .783 & .458 / .458\\
YelpChi & P1, $B{=}20$ & HGB + CARS-D & 5 & +0.000 & [+0.000, +0.000] & 5/5 & .783 / .783 & .458 / .458\\
YelpChi & P2, $B{=}20$ & HGB + CARS-D & 5 & +0.017 & [+0.014, +0.021] & 5/5 & .822 / .783 & .489 / .458\\
Amazon & P1, $B{=}2$ & CARS+AT & 3 & +0.000 & [+0.000, +0.000] & 3/3 & .891 / .891 & .818 / .818\\
Amazon & P1, $B{=}20$ & CARS+AT & 3 & +0.000 & [+0.000, +0.000] & 3/3 & .891 / .891 & .818 / .818\\
Amazon & P2, $B{=}20$ & CARS+AT & 3 & +0.000 & [+0.000, +0.000] & 3/3 & .891 / .891 & .818 / .818\\
FDCompCN & P1, $B{=}2$ & CARS+AT & 5 & +0.000 & [+0.000, +0.000] & 5/5 & .282 / .282 & .074 / .074\\
FDCompCN & P1, $B{=}20$ & CARS+AT & 5 & +0.000 & [+0.000, +0.000] & 5/5 & .282 / .282 & .074 / .074\\
FDCompCN & P2, $B{=}20$ & CARS+AT & 5 & +0.000 & [+0.000, +0.000] & 5/5 & .282 / .282 & .074 / .074\\
S-FFSD & P1, $B{=}2$ & CARS+AT & 3 & +0.000 & [+0.000, +0.000] & 3/3 & .567 / .567 & .291 / .291\\
S-FFSD & P1, $B{=}20$ & CARS+AT & 3 & +0.000 & [+0.000, +0.000] & 3/3 & .567 / .567 & .291 / .291\\
S-FFSD & P2, $B{=}20$ & CARS+AT & 3 & +0.074 & [+0.047, +0.100] & 3/3 & .682 / .567 & .418 / .291\\
Tolokers & P1, $B{=}2$ & CARS+AT & 5 & +0.001 & [-0.002, +0.003] & 5/5 & .391 / .385 & .035 / .032\\
Tolokers & P1, $B{=}20$ & CARS+AT & 5 & +0.000 & [+0.000, +0.000] & 5/5 & .385 / .385 & .032 / .032\\
Tolokers & P2, $B{=}20$ & CARS+AT & 5 & +0.001 & [-0.002, +0.003] & 5/5 & .391 / .385 & .035 / .032\\
YelpChi & P1, $B{=}2$ & CARS+AT & 3 & +0.000 & [+0.000, +0.000] & 3/3 & .528 / .528 & .180 / .180\\
YelpChi & P1, $B{=}20$ & CARS+AT & 3 & +0.000 & [+0.000, +0.000] & 3/3 & .528 / .528 & .180 / .180\\
YelpChi & P2, $B{=}20$ & CARS+AT & 3 & +0.059 & [+0.052, +0.067] & 3/3 & .633 / .528 & .240 / .180\\
Amazon & P1, $B{=}2$ & CARS+AT-HGB & 5 & -0.001 & [-0.006, +0.003] & 4/5 & .901 / .906 & .812 / .822\\
Amazon & P1, $B{=}20$ & CARS+AT-HGB & 5 & +0.000 & [+0.000, +0.000] & 5/5 & .906 / .906 & .822 / .822\\
Amazon & P2, $B{=}20$ & CARS+AT-HGB & 5 & +0.000 & [+0.000, +0.000] & 5/5 & .906 / .906 & .822 / .822\\
FDCompCN & P1, $B{=}2$ & CARS+AT-HGB & 5 & -0.000 & [-0.001, +0.000] & 5/5 & .412 / .413 & .169 / .167\\
FDCompCN & P1, $B{=}20$ & CARS+AT-HGB & 5 & +0.000 & [+0.000, +0.000] & 5/5 & .413 / .413 & .167 / .167\\
FDCompCN & P2, $B{=}20$ & CARS+AT-HGB & 5 & +0.000 & [+0.000, +0.000] & 5/5 & .413 / .413 & .167 / .167\\
S-FFSD & P1, $B{=}2$ & CARS+AT-HGB & 5 & +0.000 & [+0.000, +0.000] & 5/5 & .678 / .678 & .430 / .430\\
S-FFSD & P1, $B{=}20$ & CARS+AT-HGB & 5 & +0.000 & [+0.000, +0.000] & 5/5 & .678 / .678 & .430 / .430\\
S-FFSD & P2, $B{=}20$ & CARS+AT-HGB & 5 & +0.018 & [+0.016, +0.020] & 5/5 & .710 / .678 & .448 / .430\\
Tolokers & P1, $B{=}2$ & CARS+AT-HGB & 5 & +0.000 & [+0.000, +0.000] & 5/5 & .388 / .388 & .027 / .027\\
Tolokers & P1, $B{=}20$ & CARS+AT-HGB & 5 & +0.000 & [+0.000, +0.000] & 5/5 & .388 / .388 & .027 / .027\\
Tolokers & P2, $B{=}20$ & CARS+AT-HGB & 5 & +0.000 & [+0.000, +0.000] & 5/5 & .388 / .388 & .027 / .027\\
YelpChi & P1, $B{=}2$ & CARS+AT-HGB & 5 & +0.000 & [+0.000, +0.000] & 5/5 & .783 / .783 & .458 / .458\\
YelpChi & P1, $B{=}20$ & CARS+AT-HGB & 5 & +0.000 & [+0.000, +0.000] & 5/5 & .783 / .783 & .458 / .458\\
YelpChi & P2, $B{=}20$ & CARS+AT-HGB & 5 & +0.017 & [+0.014, +0.021] & 5/5 & .822 / .783 & .489 / .458\\

%% file: tables/tab_cost.tex
YelpChi & 187 & 1427 & 7.6\\
Amazon & 221 & 1148 & 5.2\\
FDCompCN & 45 & 58 & 1.3\\
S-FFSD & 180 & 749 & 4.2\\
Tolokers & 42 & 173 & 4.1\\

%% file: tables/tab_randcost.tex
Amazon & 24 & 0 & 0.58 & 0 & 0.58 & 0.71\\
FDCompCN & 24 & 0 & 0.00 & 0 & 0.00 & 0.79\\
S-FFSD & 24 & 24 & 1.00 & 24 & 1.00 & 0.72\\
Tolokers & 24 & 20 & 0.79 & 12 & 0.88 & 1.00\\
YelpChi & 24 & 24 & 1.00 & 22 & 0.92 & 0.76\\

%% file: tables/tab_knowledge.tex
FDCompCN & P1, $B{=}2$ & HGB + CARS-D & .794 & .794 & .794\\
FDCompCN & P1, $B{=}2$ & HGB-Graph & .411 & .420 & .764\\
FDCompCN & P1, $B{=}2$ & R-GCN & .005 & .005 & .264\\
FDCompCN & P1, $B{=}2$ & R-GCN + CARS-D & .702 & .702 & .702\\
FDCompCN & P1, $B{=}2$ & RLR & .002 & .002 & .370\\
FDCompCN & P1, $B{=}2$ & RLR + CARS-D & .663 & .663 & .672\\
FDCompCN & P1, $B{=}20$ & HGB + CARS-D & .794 & .794 & .794\\
FDCompCN & P1, $B{=}20$ & HGB-Graph & .159 & .162 & .653\\
FDCompCN & P1, $B{=}20$ & R-GCN & .005 & .005 & .186\\
FDCompCN & P1, $B{=}20$ & R-GCN + CARS-D & .702 & .702 & .702\\
FDCompCN & P1, $B{=}20$ & RLR & .001 & .001 & .321\\
FDCompCN & P1, $B{=}20$ & RLR + CARS-D & .677 & .677 & .677\\
Tolokers & P1, $B{=}2$ & HGB + CARS-D & .729 & .729 & .729\\
Tolokers & P1, $B{=}2$ & HGB-Graph & .538 & .539 & .737\\
Tolokers & P1, $B{=}2$ & R-GCN & .673 & .674 & .755\\
Tolokers & P1, $B{=}2$ & R-GCN + CARS-D & .735 & .735 & .758\\
Tolokers & P1, $B{=}2$ & RLR & .538 & .538 & .700\\
Tolokers & P1, $B{=}2$ & RLR + CARS-D & .716 & .716 & .716\\
Tolokers & P1, $B{=}20$ & HGB + CARS-D & .729 & .729 & .729\\
Tolokers & P1, $B{=}20$ & HGB-Graph & .484 & .483 & .686\\
Tolokers & P1, $B{=}20$ & R-GCN & .484 & .490 & .698\\
Tolokers & P1, $B{=}20$ & R-GCN + CARS-D & .731 & .731 & .731\\
Tolokers & P1, $B{=}20$ & RLR & .270 & .269 & .633\\
Tolokers & P1, $B{=}20$ & RLR + CARS-D & .716 & .716 & .716\\
YelpChi & P1, $B{=}2$ & HGB + CARS-D & .933 & .933 & .933\\
YelpChi & P1, $B{=}2$ & HGB-Graph & .444 & .434 & .742\\
YelpChi & P1, $B{=}2$ & R-GCN & .055 & .055 & .237\\
YelpChi & P1, $B{=}2$ & R-GCN + CARS-D & .835 & .835 & .835\\
YelpChi & P1, $B{=}2$ & RLR & .173 & .172 & .667\\
YelpChi & P1, $B{=}2$ & RLR + CARS-D & .771 & .771 & .771\\
YelpChi & P1, $B{=}20$ & HGB + CARS-D & .933 & .933 & .933\\
YelpChi & P1, $B{=}20$ & HGB-Graph & .204 & .200 & .610\\
YelpChi & P1, $B{=}20$ & R-GCN & .016 & .016 & .120\\
YelpChi & P1, $B{=}20$ & R-GCN + CARS-D & .835 & .835 & .835\\
YelpChi & P1, $B{=}20$ & RLR & .032 & .032 & .680\\
YelpChi & P1, $B{=}20$ & RLR + CARS-D & .771 & .771 & .771\\

%% file: tables/tab_entity.tex
MLP & .837 & -- & -- & --\\
HGB-X & .935 & -- & -- & --\\
RLR & .796 & .781 & .762 & .746\\
R-GCN & .916 & .520 & .436 & .850\\
HGB-Graph & .959 & .777 & .749 & .935\\
RLR + CARS-D & .776 & .769 & .769 & .790\\
R-GCN + CARS-D & .857 & .837 & .837 & .896\\
HGB + CARS-D & .940 & .935 & .935 & .951\\

%% file: tables/tab_poison_head.tex
Model & FDCo.c & FDCo.d & Tolo.c & Tolo.d & Yelp.c & Yelp.d\\

%% file: tables/tab_poison.tex
LR & .677 & .677 & .716 & .716 & .771 & .771\\
MLP & .702 & .702 & .731 & .731 & .835 & .835\\
HGB-X & .794 & .794 & .729 & .729 & .933 & .933\\
RLR & .012 & .001 & .870 & .586 & .872 & .580\\
R-GCN & .777 & .006 & .995 & .722 & .976 & .601\\
HGB & .991 & .859 & .984 & .667 & .968 & .866\\
CA-HGB & .000 & .615 & .989 & .731 & .991 & .950\\
RLR+C & .000 & .663 & .870 & .716 & .889 & .771\\
R-GCN+C & .008 & .662 & .995 & .731 & .921 & .835\\
HGB+C & .991 & .790 & .984 & .729 & .971 & .933\\
RLR+D & .677 & .663 & .870 & .716 & .889 & .771\\
R-GCN+D & .702 & .702 & .995 & .755 & .931 & .835\\
HGB+D & .991 & .790 & .984 & .729 & .971 & .933\\

%% file: tables/tab_sybil_head.tex
Model & FDCo S1,2 & FDCo S1,5 & FDCo S4,20 & Tolo S1,2 & Tolo S1,5 & Tolo S4,20 & Yelp S1,2 & Yelp S1,5 & Yelp S4,20\\

%% file: tables/tab_sybil.tex
LR & .677 & .677 & .677 & .716 & .716 & .716 & .771 & .771 & .771\\
MLP & .702 & .702 & .702 & .731 & .731 & .731 & .835 & .835 & .835\\
HGB-X & .794 & .794 & .794 & .729 & .729 & .729 & .933 & .933 & .933\\
RLR & .003 & .002 & .002 & .571 & .483 & .456 & .166 & .131 & .129\\
R-GCN & .005 & .004 & .004 & .693 & .621 & .596 & .040 & .016 & .009\\
HGB & .270 & .178 & .178 & .541 & .483 & .473 & .058 & .034 & .033\\
GNNGuard & .023 & .011 & .011 & .679 & .611 & .607 & .400 & .391 & .381\\
BWGNN & .204 & .197 & .200 & .773 & .768 & .756 & .659 & .642 & .640\\
RLR+D & .677 & .677 & .677 & .716 & .716 & .716 & .771 & .771 & .769\\
R-GCN+D & .702 & .702 & .702 & .737 & .731 & .731 & .835 & .835 & .833\\
HGB+D & .794 & .794 & .794 & .729 & .729 & .729 & .933 & .933 & .933\\

%% file: tables/tab_features.tex
YelpChi & LR & .771 & .752 & .721 & .665\\
YelpChi & MLP & .835 & .809 & .767 & .686\\
YelpChi & HGB-X & .933 & .841 & .749 & .661\\
YelpChi & RLR & .173 & .161 & .146 & .121\\
YelpChi & R-GCN & .056 & .055 & .050 & .044\\
YelpChi & HGB & .444 & .353 & .315 & .292\\
YelpChi & RLR+D & .771 & .752 & .721 & .665\\
YelpChi & R-GCN+D & .835 & .809 & .767 & .686\\
YelpChi & HGB+D & .933 & .841 & .749 & .661\\
FDCompCN & LR & .677 & .376 & .137 & .056\\
FDCompCN & MLP & .702 & .401 & .170 & .104\\
FDCompCN & HGB-X & .794 & .107 & .051 & .018\\
FDCompCN & RLR & .002 & .002 & .002 & .002\\
FDCompCN & R-GCN & .005 & .005 & .005 & .005\\
FDCompCN & HGB & .412 & .058 & .026 & .007\\
FDCompCN & RLR+D & .663 & .408 & .176 & .076\\
FDCompCN & R-GCN+D & .702 & .401 & .170 & .104\\
FDCompCN & HGB+D & .793 & .130 & .062 & .022\\
Tolokers & LR & .716 & .697 & .669 & .625\\
Tolokers & MLP & .731 & .688 & .630 & .560\\
Tolokers & HGB-X & .729 & .605 & .476 & .442\\
Tolokers & RLR & .538 & .528 & .512 & .486\\
Tolokers & R-GCN & .677 & .661 & .638 & .599\\
Tolokers & HGB & .534 & .485 & .464 & .462\\
Tolokers & RLR+D & .716 & .697 & .669 & .625\\
Tolokers & R-GCN+D & .736 & .705 & .658 & .593\\
Tolokers & HGB+D & .729 & .605 & .476 & .442\\

%% file: tables/tab_scarcity.tex
YelpChi & 0.05 & .770 & .869 & .069 & .565 & .416 & .782 & .770 & .856 & .869 & .908\\
YelpChi & 0.1 & .800 & .893 & .097 & .662 & .427 & .857 & .800 & .873 & .893 & .926\\
YelpChi & 0.2 & .818 & .920 & .046 & .651 & .440 & .902 & .818 & .885 & .920 & .944\\
FDCompCN & 0.05 & .602 & .699 & .001 & .000 & .695 & .499 & .602 & .602 & .699 & .699\\
FDCompCN & 0.1 & .679 & .760 & .001 & .000 & .667 & .433 & .679 & .679 & .765 & .760\\
FDCompCN & 0.2 & .694 & .781 & .001 & .001 & .628 & .415 & .694 & .694 & .782 & .781\\
Tolokers & 0.05 & .718 & .697 & .579 & .578 & .475 & .475 & .718 & .718 & .697 & .697\\
Tolokers & 0.1 & .719 & .706 & .586 & .583 & .437 & .437 & .719 & .719 & .706 & .706\\
Tolokers & 0.2 & .731 & .721 & .621 & .619 & .506 & .505 & .731 & .731 & .721 & .721\\

%% file: appendix_earlier.tex
\section{Earlier Robustness Checks}
\label{apd:earlier}

The checks in this section (gate-aware attacks, stronger evaluation of adversarial training, sensitivity, and the phase diagram) were run before the exhaustive-greedy attack was added, with the earlier suite (query, gradient, and transfer attacks, and the exact attack for linear models); \cref{tab:rerun} lists every experiment and its protocol. The exhaustive-greedy attack includes gate moves, which supersedes the gate-aware attack, and the stronger evaluation of adversarial training is superseded by the main table and \cref{tab:atcompute}. For clean-trained nonlinear models, the stronger attack can only lower the reported worst cases, so the traps shown here are, if anything, understated; rows of \cars{}-D and adversarially trained models were not revalidated under the exhaustive-greedy attack and should be read as evidence about the earlier threat model only.

\paragraph{Gate-aware attacks.} For every gated configuration that \cars{}-D proposed or kept, \cref{tab:adaptive} adds an attack that targets the gates: for every combination of keep, open (inject the missing edges), and close (delete accomplices until the degree falls below the threshold) over the gated relations, the remaining budget is spent on injections into the opened relation or the cheapest one, with most-benign-looking or typical endpoints (seed 0).

\begin{table}[ht]
\centering\scriptsize
\setlength{\tabcolsep}{4.5pt}

\caption{Gate-aware attacks against gated detectors: evaluated worst case of the standard suite, of the gate attack alone, and of both.}
\label{tab:adaptive}

\begin{tabular}{lll *{3}{r}}
\toprule
Dataset & Cell & Detector & suite & gate & both \\
\midrule
\input{tables/tab_adaptive.tex}
\bottomrule
\end{tabular}
\end{table}

\paragraph{Stronger evaluation of adversarial training.} \cref{tab:strongat} re-attacks CA-AT with three random restarts of a 200-step gradient attack, an AUC-targeted gradient loss, and 40 extra random allocations in the query attack; the last column is R-GCN + \cars{}-D in the same cell.

\begin{table}[ht]
\centering\scriptsize
\setlength{\tabcolsep}{2.5pt}

\caption{CA-AT under the standard suite and under stronger attacks (evaluated worst-case AUC; synthetic: 3 seeds).}
\label{tab:strongat}

\begin{tabular}{ll *{6}{r}}
\toprule
Dataset & Cell & suite & AUC loss & restarts & random alloc. & all & \cars{}-D \\
\midrule
\input{tables/tab_strongat.tex}
\bottomrule
\end{tabular}
\end{table}

\paragraph{Sensitivity of \cars{}-D.} \cref{tab:sensitivity} varies the number of degree groups ($K_f\in\{4,8,16\}$), the candidate pool (100, 300, 1000), and the threshold grid (default or coarse $\{0,2,8,34,\infty\}$), and reports the range of the evaluated worst case of the kept detector, the number of distinct configurations kept, and the median planning time (seed 0).

\begin{table}[ht]
\centering\scriptsize
\caption{Sensitivity of \cars{}-D to its hyperparameters: minimum and maximum worst-case AUC of the kept detector, number of distinct choices, median planning and checking time (seconds).}
\label{tab:sensitivity}
\begin{tabular}{lllcccc}
\toprule
Dataset & Cell & Family & min & max & choices & s\\
\midrule
\input{tables/tab_sensitivity.tex}
\bottomrule
\end{tabular}
\end{table}

\begin{figure*}[ht]
\centering
\includegraphics[width=\textwidth]{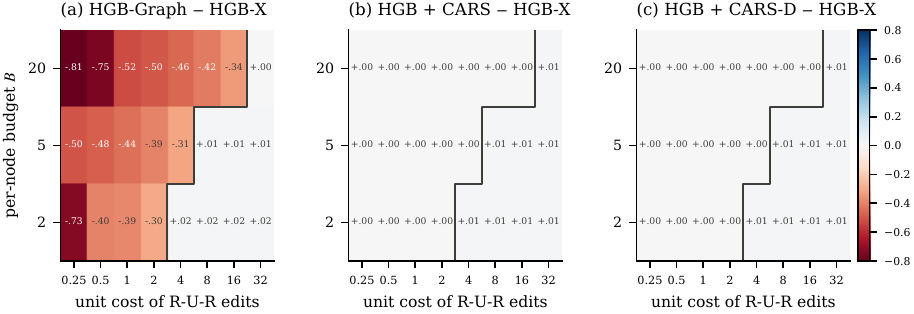}
\caption{YelpChi: evaluated worst-case AUC minus graph-free trees for HGB-Graph, HGB + \cars{}, and HGB + \cars{}-D, each planned per cell; the line marks $B=c$.}
\label{fig:phaseapp}
\end{figure*}

%% file: tables/tab_adaptive.tex
Amazon & P1, $B{=}2$ & HGB+D (kept) & .979 & .983 & .979\\
Amazon & P1, $B{=}2$ & HGB+D (prop.) & .979 & .983 & .979\\
Amazon & P1, $B{=}2$ & R-GCN+D (prop.) & .975 & .977 & .975\\
Amazon & P1, $B{=}2$ & RLR+D (prop.) & .967 & .971 & .967\\
Amazon & P1, $B{=}20$ & HGB+D (prop.) & .959 & .968 & .959\\
Amazon & P1, $B{=}20$ & R-GCN+D (prop.) & .808 & .879 & .803\\
Amazon & P1, $B{=}20$ & RLR+D (prop.) & .405 & .935 & .406\\
Amazon & P2, $B{=}20$ & HGB+D (prop.) & .963 & .969 & .962\\
Amazon & P2, $B{=}20$ & R-GCN+D (prop.) & .830 & .905 & .828\\
Amazon & P2, $B{=}20$ & RLR+D (prop.) & .390 & .933 & .390\\
FDCompCN & P1, $B{=}2$ & HGB+D (kept) & .793 & .794 & .793\\
FDCompCN & P1, $B{=}2$ & HGB+D (prop.) & .793 & .794 & .793\\
FDCompCN & P1, $B{=}2$ & R-GCN+D (prop.) & .662 & .664 & .662\\
FDCompCN & P1, $B{=}2$ & RLR+D (prop.) & .663 & .668 & .663\\
FDCompCN & P1, $B{=}20$ & HGB+D (prop.) & .794 & .794 & .794\\
FDCompCN & P1, $B{=}20$ & R-GCN+D (prop.) & .707 & .707 & .707\\
FDCompCN & P1, $B{=}20$ & RLR+D (prop.) & .673 & .677 & .673\\
FDCompCN & P2, $B{=}20$ & HGB+D (prop.) & .794 & .794 & .794\\
FDCompCN & P2, $B{=}20$ & R-GCN+D (prop.) & .707 & .707 & .707\\
FDCompCN & P2, $B{=}20$ & RLR+D (prop.) & .677 & .677 & .677\\
S-FFSD & P1, $B{=}2$ & HGB+D (prop.) & .651 & .714 & .633\\
S-FFSD & P1, $B{=}2$ & R-GCN+D (prop.) & .663 & .785 & .662\\
S-FFSD & P1, $B{=}2$ & RLR+D (prop.) & .554 & .621 & .554\\
Tolokers & P1, $B{=}2$ & HGB+D (prop.) & .723 & .742 & .721\\
Tolokers & P1, $B{=}2$ & R-GCN+D (kept) & .744 & .756 & .744\\
Tolokers & P1, $B{=}2$ & R-GCN+D (prop.) & .744 & .756 & .744\\
Tolokers & P1, $B{=}2$ & RLR+D (prop.) & .719 & .728 & .719\\
Tolokers & P2, $B{=}20$ & HGB+D (prop.) & .722 & .742 & .720\\
Tolokers & P2, $B{=}20$ & R-GCN+D (kept) & .744 & .756 & .744\\
Tolokers & P2, $B{=}20$ & R-GCN+D (prop.) & .744 & .756 & .744\\
Tolokers & P2, $B{=}20$ & RLR+D (prop.) & .719 & .728 & .719\\
YelpChi & P1, $B{=}2$ & HGB+D (prop.) & .867 & .886 & .863\\
YelpChi & P1, $B{=}2$ & R-GCN+D (prop.) & .775 & .799 & .772\\
YelpChi & P1, $B{=}2$ & RLR+D (prop.) & .683 & .742 & .683\\

%% file: tables/tab_strongat.tex
Synthetic & P1, $B{=}4$ & .887 & .853 & .874 & .925 & .842 & .681\\
Tolokers & P1, $B{=}2$ & .762 & .762 & .760 & .770 & .746 & .741\\
Tolokers & P2, $B{=}20$ & .764 & .764 & .763 & .769 & .752 & .741\\
YelpChi & P1, $B{=}2$ & .752 & .759 & .757 & .808 & .731 & .836\\
YelpChi & P2, $B{=}20$ & .901 & .896 & .895 & .915 & .892 & .897\\

%% file: tables/tab_sensitivity.tex
Amazon & P1, $B{=}2$ & HGB & .977 & .980 & 2 & 123\\
Amazon & P1, $B{=}2$ & RLR & .969 & .969 & 1 & 123\\
Amazon & P2, $B{=}20$ & HGB & .966 & .976 & 3 & 98\\
Amazon & P2, $B{=}20$ & RLR & .969 & .969 & 1 & 98\\
S-FFSD & P1, $B{=}2$ & HGB & .842 & .842 & 1 & 180\\
S-FFSD & P1, $B{=}2$ & RLR & .643 & .643 & 1 & 180\\
S-FFSD & P2, $B{=}20$ & HGB & .860 & .860 & 1 & 108\\
S-FFSD & P2, $B{=}20$ & RLR & .661 & .661 & 1 & 108\\
Tolokers & P1, $B{=}2$ & HGB & .729 & .729 & 1 & 12\\
Tolokers & P1, $B{=}2$ & RLR & .716 & .716 & 1 & 12\\
Tolokers & P2, $B{=}20$ & HGB & .729 & .729 & 1 & 10\\
Tolokers & P2, $B{=}20$ & RLR & .716 & .716 & 1 & 10\\
YelpChi & P1, $B{=}2$ & HGB & .933 & .933 & 1 & 131\\
YelpChi & P1, $B{=}2$ & RLR & .771 & .771 & 1 & 131\\
YelpChi & P2, $B{=}20$ & HGB & .948 & .948 & 1 & 90\\
YelpChi & P2, $B{=}20$ & RLR & .794 & .794 & 1 & 90\\

%% file: main.bbl
\begin{thebibliography}{36}
\providecommand{\natexlab}[1]{#1}
\providecommand{\url}[1]{\texttt{#1}}
\expandafter\ifx\csname urlstyle\endcsname\relax
  \providecommand{\doi}[1]{doi: #1}\else
  \providecommand{\doi}{doi: \begingroup \urlstyle{rm}\Url}\fi

\bibitem[Baranwal et~al.(2021)Baranwal, Fountoulakis, and Jagannath]{baranwal2021graph}
Baranwal, A., Fountoulakis, K., and Jagannath, A.
\newblock Graph convolution for semi-supervised classification: Improved linear separability and out-of-distribution generalization.
\newblock In \emph{Proceedings of the 38th International Conference on Machine Learning (ICML)}, pp.\  684--693, 2021.

\bibitem[Bojchevski \& G{\"u}nnemann(2019)Bojchevski and G{\"u}nnemann]{bojchevski2019certifiable}
Bojchevski, A. and G{\"u}nnemann, S.
\newblock Certifiable robustness to graph perturbations.
\newblock In \emph{Advances in Neural Information Processing Systems (NeurIPS)}, 2019.

\bibitem[Chen et~al.(2019)Chen, Zhang, Si, Li, Boning, and Hsieh]{chen2019treeverif}
Chen, H., Zhang, H., Si, S., Li, Y., Boning, D., and Hsieh, C.-J.
\newblock Robustness verification of tree-based models.
\newblock In \emph{Advances in Neural Information Processing Systems}, 2019.

\bibitem[Choi et~al.(2025)Choi, Kim, and Whang]{choi2025monti}
Choi, J., Kim, H., and Whang, J.~J.
\newblock Unveiling the threat of fraud gangs to graph neural networks: Multi-target graph injection attacks against {GNN}-based fraud detectors.
\newblock In \emph{Proceedings of the AAAI Conference on Artificial Intelligence}, volume~39, pp.\  16028--16036, 2025.

\bibitem[Dalvi et~al.(2004)Dalvi, Domingos, Mausam, Sanghai, and Verma]{dalvi2004adversarial}
Dalvi, N., Domingos, P., Mausam, Sanghai, S., and Verma, D.
\newblock Adversarial classification.
\newblock In \emph{Proceedings of the ACM SIGKDD International Conference on Knowledge Discovery and Data Mining}, pp.\  99--108, 2004.

\bibitem[Deshpande et~al.(2018)Deshpande, Sen, Montanari, and Mossel]{deshpande2018csbm}
Deshpande, Y., Sen, S., Montanari, A., and Mossel, E.
\newblock Contextual stochastic block models.
\newblock In \emph{Advances in Neural Information Processing Systems (NeurIPS)}, 2018.

\bibitem[Dou et~al.(2020)Dou, Liu, Sun, Deng, Peng, and Yu]{dou2020caregnn}
Dou, Y., Liu, Z., Sun, L., Deng, Y., Peng, H., and Yu, P.~S.
\newblock Enhancing graph neural network-based fraud detectors against camouflaged fraudsters.
\newblock In \emph{Proceedings of the 29th ACM International Conference on Information and Knowledge Management (CIKM)}, pp.\  315--324, 2020.

\bibitem[Eilat et~al.(2023)Eilat, Finkelshtein, Baskin, and Rosenfeld]{eilat2023strategic}
Eilat, I., Finkelshtein, B., Baskin, C., and Rosenfeld, N.
\newblock Strategic classification with graph neural networks.
\newblock In \emph{International Conference on Learning Representations (ICLR)}, 2023.

\bibitem[Gao et~al.(2023)Gao, Wang, He, Liu, Feng, and Zhang]{gao2023ghrn}
Gao, Y., Wang, X., He, X., Liu, Z., Feng, H., and Zhang, Y.
\newblock Addressing heterophily in graph anomaly detection: A perspective of graph spectrum.
\newblock In \emph{Proceedings of the ACM Web Conference (WWW)}, pp.\  1528--1538, 2023.

\bibitem[Geisler et~al.(2021)Geisler, Schmidt, {\c{S}}irin, Z{\"u}gner, Bojchevski, and G{\"u}nnemann]{geisler2021prbcd}
Geisler, S., Schmidt, T., {\c{S}}irin, H., Z{\"u}gner, D., Bojchevski, A., and G{\"u}nnemann, S.
\newblock Robustness of graph neural networks at scale.
\newblock In \emph{Advances in Neural Information Processing Systems (NeurIPS)}, 2021.

\bibitem[Gowal et~al.(2018)Gowal, Dvijotham, Stanforth, Bunel, Qin, Uesato, Arandjelovic, Mann, and Kohli]{gowal2018ibp}
Gowal, S., Dvijotham, K., Stanforth, R., Bunel, R., Qin, C., Uesato, J., Arandjelovic, R., Mann, T., and Kohli, P.
\newblock On the effectiveness of interval bound propagation for training verifiably robust models.
\newblock \emph{arXiv preprint arXiv:1810.12715}, 2018.

\bibitem[Hardt et~al.(2016)Hardt, Megiddo, Papadimitriou, and Wootters]{hardt2016strategic}
Hardt, M., Megiddo, N., Papadimitriou, C., and Wootters, M.
\newblock Strategic classification.
\newblock In \emph{Proceedings of the 2016 ACM Conference on Innovations in Theoretical Computer Science (ITCS)}, pp.\  111--122, 2016.

\bibitem[Kireev et~al.(2023)Kireev, Kulynych, and Troncoso]{kireev2023tabular}
Kireev, K., Kulynych, B., and Troncoso, C.
\newblock Adversarial robustness for tabular data through cost and utility awareness.
\newblock In \emph{Network and Distributed System Security Symposium (NDSS)}, 2023.

\bibitem[Liao et~al.(2024)Liao, Fu, Wang, Wei, and Xu]{liao2024riskeeper}
Liao, J., Fu, W., Wang, C., Wei, Z., and Xu, J.
\newblock Value at adversarial risk: A graph defense strategy against cost-aware attacks.
\newblock In \emph{Proceedings of the AAAI Conference on Artificial Intelligence}, volume~38, pp.\  13763--13771, 2024.

\bibitem[Liu et~al.(2021)Liu, Ao, Qin, Chi, Feng, Yang, and He]{liu2021pcgnn}
Liu, Y., Ao, X., Qin, Z., Chi, J., Feng, J., Yang, H., and He, Q.
\newblock Pick and choose: A {GNN}-based imbalanced learning approach for fraud detection.
\newblock In \emph{Proceedings of the Web Conference (WWW)}, pp.\  3168--3177, 2021.

\bibitem[Lowd \& Meek(2005)Lowd and Meek]{lowd2005adversarial}
Lowd, D. and Meek, C.
\newblock Adversarial learning.
\newblock In \emph{Proceedings of the ACM SIGKDD International Conference on Knowledge Discovery and Data Mining}, pp.\  641--647, 2005.

\bibitem[Ma et~al.(2022)Ma, Liu, Shah, and Tang]{ma2022homophily}
Ma, Y., Liu, X., Shah, N., and Tang, J.
\newblock Is homophily a necessity for graph neural networks?
\newblock In \emph{International Conference on Learning Representations (ICLR)}, 2022.

\bibitem[Madry et~al.(2018)Madry, Makelov, Schmidt, Tsipras, and Vladu]{madry2018pgd}
Madry, A., Makelov, A., Schmidt, L., Tsipras, D., and Vladu, A.
\newblock Towards deep learning models resistant to adversarial attacks.
\newblock In \emph{International Conference on Learning Representations (ICLR)}, 2018.

\bibitem[McAuley \& Leskovec(2013)McAuley and Leskovec]{mcauley2013amazon}
McAuley, J.~J. and Leskovec, J.
\newblock From amateurs to connoisseurs: Modeling the evolution of user expertise through online reviews.
\newblock In \emph{Proceedings of the 22nd International Conference on World Wide Web (WWW)}, pp.\  897--908, 2013.

\bibitem[Mujkanovic et~al.(2022)Mujkanovic, Geisler, G{\"u}nnemann, and Bojchevski]{mujkanovic2022defenses}
Mujkanovic, F., Geisler, S., G{\"u}nnemann, S., and Bojchevski, A.
\newblock Are defenses for graph neural networks robust?
\newblock In \emph{Advances in Neural Information Processing Systems (NeurIPS)}, 2022.

\bibitem[Platonov et~al.(2023)Platonov, Kuznedelev, Diskin, Babenko, and Prokhorenkova]{platonov2023critical}
Platonov, O., Kuznedelev, D., Diskin, M., Babenko, A., and Prokhorenkova, L.
\newblock A critical look at the evaluation of {GNNs} under heterophily: Are we really making progress?
\newblock In \emph{International Conference on Learning Representations (ICLR)}, 2023.

\bibitem[Rayana \& Akoglu(2015)Rayana and Akoglu]{rayana2015yelp}
Rayana, S. and Akoglu, L.
\newblock Collective opinion spam detection: Bridging review networks and metadata.
\newblock In \emph{Proceedings of the 21st ACM SIGKDD International Conference on Knowledge Discovery and Data Mining}, pp.\  985--994, 2015.

\bibitem[Schlichtkrull et~al.(2018)Schlichtkrull, Kipf, Bloem, van~den Berg, Titov, and Welling]{schlichtkrull2018rgcn}
Schlichtkrull, M., Kipf, T.~N., Bloem, P., van~den Berg, R., Titov, I., and Welling, M.
\newblock Modeling relational data with graph convolutional networks.
\newblock In \emph{The Semantic Web: 15th International Conference (ESWC)}, pp.\  593--607, 2018.

\bibitem[Tang et~al.(2022)Tang, Li, Gao, and Li]{tang2022bwgnn}
Tang, J., Li, J., Gao, Z., and Li, J.
\newblock Rethinking graph neural networks for anomaly detection.
\newblock In \emph{Proceedings of the 39th International Conference on Machine Learning (ICML)}, pp.\  21076--21089, 2022.

\bibitem[Tang et~al.(2023)Tang, Hua, Gao, Zhao, and Li]{tang2023gadbench}
Tang, J., Hua, F., Gao, Z., Zhao, P., and Li, J.
\newblock {GADBench}: Revisiting and benchmarking supervised graph anomaly detection.
\newblock In \emph{Advances in Neural Information Processing Systems (NeurIPS), Datasets and Benchmarks Track}, 2023.

\bibitem[Tram{\`e}r et~al.(2020)Tram{\`e}r, Carlini, Brendel, and Madry]{tramer2020adaptive}
Tram{\`e}r, F., Carlini, N., Brendel, W., and Madry, A.
\newblock On adaptive attacks to adversarial example defenses.
\newblock In \emph{Advances in Neural Information Processing Systems (NeurIPS)}, 2020.

\bibitem[Vincess \& Melvin(2026)Vincess and Melvin]{anon2026workshop}
Vincess, D. and Melvin, F.
\newblock Not all edges cost the same: Relation-specific manipulation costs and the trust trap in fraud graphs.
\newblock In \emph{GraphML Workshop at the Asian Conference on Machine Learning (ACML)}, 2026.

\bibitem[Wang et~al.(2019)Wang, Ji, Shi, Wang, Ye, Cui, and Yu]{wang2019han}
Wang, X., Ji, H., Shi, C., Wang, B., Ye, Y., Cui, P., and Yu, P.~S.
\newblock Heterogeneous graph attention network.
\newblock In \emph{Proceedings of the World Wide Web Conference (WWW)}, pp.\  2022--2032, 2019.

\bibitem[Wu et~al.(2023)Wu, Yao, Zhang, Chao, and Li]{wu2023splitgnn}
Wu, B., Yao, X., Zhang, B., Chao, K.-M., and Li, Y.
\newblock {SplitGNN}: Spectral graph neural network for fraud detection against heterophily.
\newblock In \emph{Proceedings of the 32nd ACM International Conference on Information and Knowledge Management (CIKM)}, 2023.

\bibitem[Xiang et~al.(2023)Xiang, Zhu, Cheng, Li, Zhao, Ouyang, Chen, and Zheng]{xiang2023gtan}
Xiang, S., Zhu, M., Cheng, D., Li, E., Zhao, R., Ouyang, Y., Chen, L., and Zheng, Y.
\newblock Semi-supervised credit card fraud detection via attribute-driven graph representation.
\newblock In \emph{Proceedings of the AAAI Conference on Artificial Intelligence}, volume~37, pp.\  14557--14565, 2023.

\bibitem[Zhang et~al.(2024)Zhang, Xu, Lv, Shi, Shen, Jin, and Dong]{zhang2024digin}
Zhang, J., Xu, Z., Lv, D., Shi, Z., Shen, D., Jin, J., and Dong, F.
\newblock {DiG-In-GNN}: Discriminative feature guided {GNN}-based fraud detector against inconsistencies in multi-relation fraud graph.
\newblock In \emph{Proceedings of the AAAI Conference on Artificial Intelligence}, volume~38, pp.\  9323--9331, 2024.

\bibitem[Zhang et~al.(2022)Zhang, Wang, Zhu, Shi, Zhang, and Zhou]{zhang2022rohe}
Zhang, M., Wang, X., Zhu, M., Shi, C., Zhang, Z., and Zhou, J.
\newblock Robust heterogeneous graph neural networks against adversarial attacks.
\newblock In \emph{Proceedings of the AAAI Conference on Artificial Intelligence}, volume~36, pp.\  4363--4370, 2022.

\bibitem[Zhang \& Evans(2019)Zhang and Evans]{zhang2019costsensitive}
Zhang, X. and Evans, D.
\newblock Cost-sensitive robustness against adversarial examples.
\newblock In \emph{International Conference on Learning Representations}, 2019.

\bibitem[Zhang \& Zitnik(2020)Zhang and Zitnik]{zhang2020gnnguard}
Zhang, X. and Zitnik, M.
\newblock {GNNGuard}: Defending graph neural networks against adversarial attacks.
\newblock In \emph{Advances in Neural Information Processing Systems (NeurIPS)}, 2020.

\bibitem[Z{\"u}gner \& G{\"u}nnemann(2019)Z{\"u}gner and G{\"u}nnemann]{zugner2019certifiable}
Z{\"u}gner, D. and G{\"u}nnemann, S.
\newblock Certifiable robustness and robust training for graph convolutional networks.
\newblock In \emph{Proceedings of the 25th ACM SIGKDD International Conference on Knowledge Discovery and Data Mining (KDD)}, 2019.

\bibitem[Z{\"u}gner et~al.(2018)Z{\"u}gner, Akbarnejad, and G{\"u}nnemann]{zugner2018nettack}
Z{\"u}gner, D., Akbarnejad, A., and G{\"u}nnemann, S.
\newblock Adversarial attacks on neural networks for graph data.
\newblock In \emph{Proceedings of the 24th ACM SIGKDD International Conference on Knowledge Discovery \& Data Mining}, pp.\  2847--2856, 2018.

\end{thebibliography}
